\documentclass[journal=jacsat,manuscript=article]{achemso}
\usepackage[version=3]{mhchem}
    \PassOptionsToPackage{numbers, compress}{natbib}
\usepackage{subcaption}
\usepackage[utf8]{inputenc} % allow utf-8 input
\usepackage[T1]{fontenc}    % use 8-bit T1 fonts
\usepackage{hyperref}       % hyperlinks
\usepackage{url}            % simple URL typesetting
\usepackage{booktabs}       % professional-quality tables
\usepackage{amsfonts}       % blackboard math symbols
\usepackage{nicefrac}       % compact symbols for 1/2, etc.
\usepackage{microtype}      % microtypography
\usepackage{xcolor}         % colors
\usepackage{graphicx} 
\usepackage{amsmath}
\usepackage{amssymb}
\usepackage{mathtools}
\usepackage{amsthm}
\usepackage{array}
\usepackage{bbm}
\usepackage{thm-restate}
\usepackage{float}

\usepackage{graphicx}%
\usepackage{multirow}%
\usepackage{amsmath,amssymb,amsfonts}%
\usepackage{amsthm}%
\usepackage{mathrsfs}%
\usepackage[title]{appendix}%
\usepackage{xcolor}%
\usepackage{textcomp}%
\usepackage{manyfoot}%
\usepackage{booktabs}%
\usepackage{listings}%
\usepackage{color,soul}
\usepackage{tablefootnote}
\usepackage{academicons}
\usepackage{orcidlink}

\title{Advancing Drug Discovery with Enhanced Chemical Understanding via Asymmetric Contrastive Multimodal Learning}

\author{Yifei Wang\,\orcidlink{0000-0002-8295-5534}}
\affiliation[Brandeis]
{Department of Computer Science, Brandeis University, MA}
\altaffiliation{Equal contribution}
\author{Yunrui Li\,\orcidlink{0009-0006-6052-2304}}
\affiliation[Brandeis]
{Department of Computer Science, Brandeis University, MA}
\altaffiliation{Equal contribution}
\author{Lin Liu\,\orcidlink{0000-0002-3948-5004}}
\affiliation[Stanford]
{Department of Chemistry, Stanford University, CA}
\author{Pengyu Hong\,\orcidlink{0000-0002-3177-2754}}
\affiliation[Brandeis]
{Department of Computer Science, Brandeis University, MA}

\author{Hao Xu\,\orcidlink{0000-0002-9795-5633}}
\altaffiliation{Equal contribution}
\affiliation[Northeastern University]
{Department of Physics, Northeastern University, MA}
\alsoaffiliation[Brigham and Woman's Hospital]
{Department of Medicine, Brigham and Woman's Hospital, Harvard Medical School, MA}

\email{h.xu@northeastern.edu}
\begin{document}
\maketitle
\begin{abstract}
 The versatility of multimodal deep learning holds tremendous promise for advancing scientific research and practical applications. As this field continues to evolve, the collective power of cross-modal analysis promises to drive transformative innovations, opening new frontiers in chemical understanding and drug discovery. Hence, we introduce \textbf{A}symmetric \textbf{C}ontrastive \textbf{M}ultimodal \textbf{L}earning (ACML), a specifically designed approach to enhance molecular understanding and accelerate advancements in drug discovery. ACML harnesses the power of effective asymmetric contrastive learning to seamlessly transfer information from various chemical modalities to molecular graph representations. By combining pre-trained chemical unimodal encoders and a shallow-designed graph encoder with 5 layers, ACML facilitates the assimilation of coordinated chemical semantics from different modalities, leading to comprehensive representation learning with efficient training. We demonstrate the effectiveness of this framework through large-scale cross-modality retrieval and isomer discrimination tasks. Additionally, ACML enhances interpretability by revealing chemical semantics in graph presentations and bolsters the expressive power of graph neural networks, as evidenced by improved performance in molecular property prediction tasks from MoleculeNet and Therapeutics Data Commons (TDC). Ultimately, ACML exemplifies its potential to revolutionize molecular representational learning, offering deeper insights into the chemical semantics of diverse modalities and paving the way for groundbreaking advancements in chemical research and drug discovery.
\end{abstract}

\section{Introduction}
\label{sec1}
%The Introduction section I (multi-modality and its applications):
Multimodal deep learning (MMDL) is a thriving and interdisciplinary research field that is dedicated to enhancing artificial intelligence (AI) capabilities in comprehending, reasoning, and deducing valuable information across various communicative modalities \cite{bib1}. It serves as an effective approach for the communication and integration of diverse data sources (text, images, audio, video, sensor data, etc), inspiring the innovative creation of more accurate and powerful AI models.  For example, DALL-E2 \cite{bib4} can showcase substantial progress in image generation and modification given a short text prompt.  Witnessing the recent surge of research in image and video comprehension \cite{bib2, bib3}, text-to-image generation \cite{bib4, bib5}, and embodied autonomous agents \cite{bib6, bib7}, multidisciplinary research, including biology, chemistry, and physics, also starts to embrace multimodal deep learning to unlock profound insights into complex systems, tackle challenging scientific problems, and push the boundaries of knowledge \cite{bib8, bib9, bib10, ektefaie2023multimodal, wen2023multimodal}. 

% \begin{figure}[t]
%     \centering\includegraphics[width=0.8\linewidth]{figures/chem_modalities.png}
%     \caption{Representative Chemical Modalities}
%     \label{fig:chem_modalities}
% \end{figure}

\begin{figure}[H]
    \centering
    \includegraphics[width=1.0\linewidth]{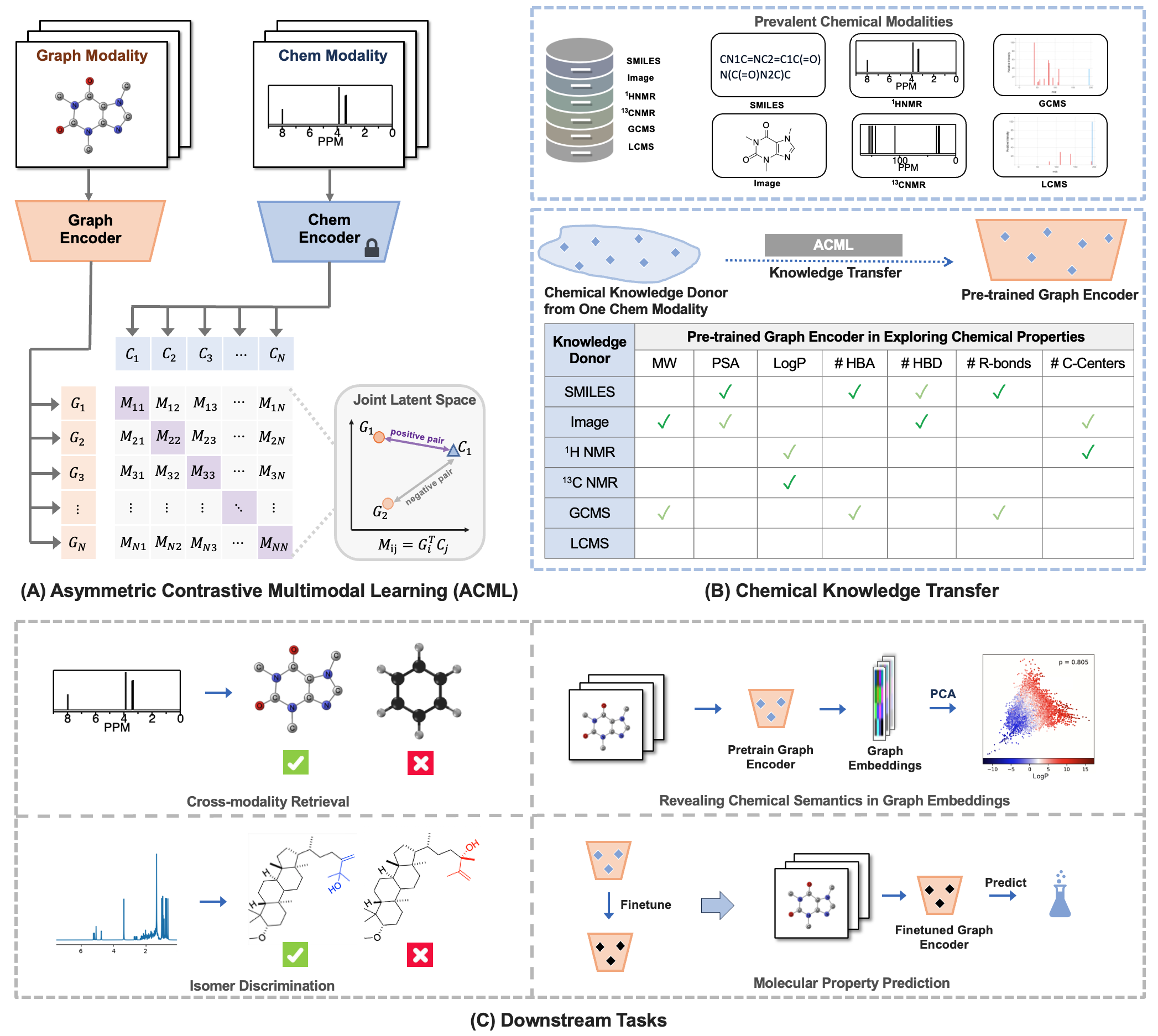}
    \caption{The framework of ACML (\textbf{A}symmetric \textbf{C}ontrastive \textbf{M}ultimodal \textbf{L}earning). \textbf{a}. \textbf{The conceptual view of ACML}. Multimodal learning on multiple pairs of graph-chemical modalities (SMILES, Images, NMR, or Mass spectrometry) is established through asymmetric CLIP architecture \cite{radford2021learning}. The graph encoder is trained through ACML, while the encoder of chemical modality is pretrained and exempted from training. Embeddings from two modalities are projected into a joint latent space, aiming to align multiple views of the same molecule while distancing the embeddings from different molecules.  \textbf{b}. \textbf{Chemical knowledge transfer in ACML}. For each chemical modality, an individual from-scratch graph encoder is paired and trained through the ACML framework, enabling it to express the learned chemical semantics from the corresponding chemical modality through latent embeddings. \textbf{c}. \textbf{Comprehensive downstream tasks} to demonstrate the effectiveness, interpretability, and generalization ability of graph learning in ACML.}
    \label{fig:chem_framework}
\end{figure}
% multi-modality in chemistry
In the realm of chemistry, MMDL has been extensively adapted to facilitate the representation learning across several chemical modalities, including chemical language, chemical notations, and molecular graphs \cite{bib11, bib12, bib13, bib14, bib15, bib16, bib17}. It is essential to note that the potential of MMDL extends far beyond the above modalities \cite{bib18, zhang2020transfer,  gao20202d, li2023deep, zou2023deep, hirst2023ml, lai2023artificial, tao2024machine}. For instance, the molecular graphical depiction (referred to as image in the rest of the paper) \cite{clevert2021img2mol} can offer powerful visual illustrations of molecules, providing unique insights into their spatial arrangements. Nuclear magnetic resonance (NMR) spectra, including \textsuperscript{1}H NMR and \textsuperscript{13}C NMR, offer a view of functional groups and characteristic features, enabling a deeper understanding of molecular behaviors and functionalities. Additionally, mass spectra from techniques such as gas chromatography–mass spectrometry (GCMS) and liquid chromatography–mass spectrometry (LCMS) provide a clue of molecular composition and fragmentation, aiding the identification and characterization of compounds. However, representing molecular information comprehensively using a single modality is a challenging task \cite{wang2022molecular}. For example, string-based representations like SMILES \cite{weininger1988smiles} lack important topology information, whereas molecular depictions fail to include electronic information. MMDL could provide a solution by allowing the communication and integration of various heterogeneous modalities. However, there is limited work exploring such advantages in chemical systems, transcending the limitations associated with each modality.

%The Introduction section II(contrastive learning), its connection with chemical application, and the challenges:
Among the implementations of MMDL, contrastive learning has emerged as one of the most prevalent self-supervised approaches for information communication across multiple modalities \cite{zong2023self}. Its objective is to align multiple views of the same instance across modalities while distancing the representations from different instances. One typical approach is to adopt a coordinated mechanism, like CLIP (Contrastive Language-Image Pretraining) framework \cite{radford2021learning}, to enhance multimodal communication. The coordinated model establishes correspondences between the learned representations by first projecting these representations onto a joint hidden space and then contrasting positive pairs (similar instances) against negative pairs (dissimilar instances). In the domain of chemistry, great efforts have been directed towards using contrastive learning for tasks such as property predictions \cite{bib13}, unimodal pattern mining \cite{bib21}, and zero-shot classifications \cite{bib18}. However, there remains a notable gap in research investigating the potential for integrating information across different modalities and elucidating the interpretability of learned representations within the context of multimodal coordination. 
Many recent works have demonstrated strong advantages of molecular images \cite{zeng2022accurate}, SMILES \cite{li2022novel}, GCMS \cite{guo2024gcmsformer}, and molecular graphs \cite{jiang2021could} to produce molecular representations for drug discovery. But what contributes to such superiority? It's important to recognize that the drug discovery process operates on multiple layers of informational hierarchy. These layers range from atomic-level properties such as hydrogen-bonding acceptors (HBA) to motif-level characteristics like hydrogen-bonding donors (HBD) and the number of stereogenic (chiral) centers, and finally to molecular-level attributes such as molecular weight, topological polar surface area (TPSA), and the logarithmic partition coefficient (LogP). To answer the question, we may need to investigate how diverse modalities impact on these hierarchical properties. 
%The Introduction section IV(our proposal):

%The inheritance of molecular graph-like structure allows graph to convey extensive information about molecules so that graph encoders like graph neural networks (GNNs) can capture atom-level (node-level), functional group-level (motif-level), and molecule-level (graph-level) information \cite{bib22, gilmer2017neural, wang2023motif, guo2022graph}. Graph, as a flexible data structure, possesses the capability to represent complex relationships and interactions between nodes and accommodate changes in data by adding or removing node/edge features or modifying the topology. This inherent flexibility make graph a valuable acceptor to assimilate information from other modalities. A promising approach is to transfer the information from other modalities into the graph. Comparison of the resulting graph embedding space could enable comprehensive evaluation and a more holistic understanding of information richness of different input chemical modalities. 

In light of these opportunities, we propose a novel approach called Asymmetric Contrastive Multimodal Learning (ACML) specifically designed for molecules. It leverages contrastive learning between the molecular graph and other prevalent chemical modalities, including G-SMILES, G-Image, G-\textsuperscript{1}H NMR, G-\textsuperscript{13}C NMR, G-GCMS, and G-LCMS pairs (G for the abbreviation of Graph), to transfer the information from other chemical modalities into the graph representations in an asymmetric way. As graph representation can express hierarchical molecular information via carefully crafted graph neural networks (GNNs) \cite{bib22, gilmer2017neural, alsentzer2020subgraph, sun2021sugar, guo2022graph, sun2022prediction, wang2023motif, mucllari2023novel}, we regard it as a valuable receptor containing basic topology information of a given molecule to assimilate information from other modalities. Unlike other initiatives in molecular properties prediction \cite{wu2018topp, wu2018quantitative}, our emphasis is primarily on direct data mining utilizing graph latent representation. ACML enables graph representation learning to capture knowledge across various chemical modalities, promoting a more holistic understanding of the hierarchical molecular information within diverse input chemical modalities. This multimodal learning framework not only enhances the interpretability of learned representations but also holds the potential to significantly improve the expressive power of graph neural networks in tasks related to drug discovery.  Figure~\ref{fig:chem_framework} illustrates the conceptual view of the ACML framework.

%Further extension is required!
In our proposed framework, each ACML model involves the graph encoder and one chemical unimodal encoder, for example, G-SMILES represents the multimodal learning between graph and SMILES modalities. We utlize a dedicated pre-trained unimodal encoder for each chemical modalitiy (SMILES, Image, \textsuperscript{1}H NMR, \textsuperscript{13}C NMR, GCMS, and LCMS), keeping its parameters fixed in the ACML training stage. Concurrently, we train a simple and shallow graph encoder with 5 convolutional layers to effectively capture different representations of each modality. This approach allows the fixed chemical encoder to effectively transfer knowledge to graph encoders through asymmetric contrastive learning. Finally, we analyze and compare the embeddings generated by the trained graph encoders to discover their explanatory power. In summary, the advantages of the ACML framework can be viewed as follows:
\begin{enumerate}
    \item Leverage effective asymmetric contrastive learning between graph and chemical modalities.
    \item Achieve an efficient training scheme using the shallow graph encoder with 5 convolutional layers and pre-trained chemical encoders.
    \item Empower knowledge transfer from chemical modalities into graph encoder during cross-modal chemical semantics learning.
    \item Demonstrate the effectiveness and interpretability of graph learning in ACML, including three key tasks: (1) cross-modality retrieval (2) isomer discrimination, (3) revealing chemical semantics in graph embeddings, and (4) molecular property prediction.
\end{enumerate}
\section{Methodology}

The proposed ACML framework is built upon the contrastive learning framework inspired by Radford et al.~\cite{radford2021learning}. It effectively promotes correspondences between modalities of the same molecule (positive pairs) while contrasting them with different molecules (negative pairs). The entire framework contains four components: (1) a frozen unimodal encoder of chemical modality, (2) a trainable graph encoder, (3) projection modules, and (4) contrastive loss. The goal is to map the hidden representations of the molecular graph and the chemical modality into the joint latent space. Details of each component and training settings are provided in \textit{\textbf{Supporting Information Material S1: Model Architecture}}.

% For each chemical modality, we make use of effective pre-trained encoders from publicly available sources, which have been demonstrated to be effective in the corresponding downstream tasks (See Table~\ref{tab:pretrain-encoder}).
The inputs of each ACML framework consist of a molecular graph modality and one chemical modality. A frozen pretrained encoder is applied for each chemical modality while a trainable graph encoder with random initialization is employed for the graph modality. These inputs are then fed into their respective encoders to produce embeddings. Mathematically, let $g$ and $c$ denote a molecule's graph representation and its chemical modality representation, their embeddings after passing to their corresponding encoders are:
\begin{equation}
    \mathbf{h}_g = \text{ENCODER}_{\text{graph}}(g), \quad \mathbf{h}_c = \text{ENCODER}_{\text{chem}}(c)
\end{equation}

It's worth noting that $\mathbf{h}_g \in \mathbb{R}^{d_{1}}$ and $\mathbf{h}_c \in \mathbb{R}^{d_{2}}$ may have different dimensions, i.e., $d_{1} \neq d_{2}$. To ensure that the information from two modalities could be coordinated in the same dimensional latent space, we apply two projection modules after both encoders. These modules assist in facilitating and coordinating between two modalities, resulting in the same dimensional hidden representations as projection outputs. In this study, the projection modules are designed as multiple-layer perceptrons (MLPs), which could be formulated as:
\begin{equation}
\label{equ:projection}
    \mathbf{h}_{g}^{\text{proj}} = \text{MLP}_{\text{graph}}(\mathbf{h}_g), \quad \mathbf{h}_{c}^{\text{proj}} = \text{MLP}_{\text{chem}}(\mathbf{h}_c)
\end{equation}
where $\mathbf{h}_{g}^{\text{proj}} \in \mathbb{R}^{d}$ and $\mathbf{h}_{g}^{\text{proj}} \in \mathbb{R}^{d}$ have the same dimension.

Figure~\ref{fig:chem_framework} provides the conceptual view of ACML. The asymmetric contrastive learning is established on multiple pairs of graph-chemical modalities. Positive pairs are formed by matching the molecular graph modality and its corresponding unimodal chemical modality (e.g., SMILES, Image, NMR, or Mass spectrometry) of the same molecule, while negative pairs are formed by pairing the graph of a molecule with the modality representation of a different molecule. During training, given a minibatch of molecules, we compute a pairwise similarity score matrix of the embeddings between the graph and another chemical modality. In this matrix, the diagonal elements correspond to positive pairs (i.e., matching representations of the same molecule), while the off-diagonal elements correspond to negative pairs (i.e., mismatched molecule representations). We chose binary  contrastive loss due to its simplicity, interpretability, and lower computational overhead. Advanced options such as triplet margin loss \cite{schroff2015facenet} and Supervised Contrastive Hard Loss \cite{yu2023enzyme, jiang2024supervised} may offer improved instance-level discrimination and enhance modality alignment, especially in ambiguous or low-information scenarios, which serves as future direction.

The ACML framework exhibits adaptability to diverse encoder designs, facilitating information extraction and compression. In our study, the encoders for each modality encompass convolutional neural networks, trained to capture features for subsequent reconstruction tasks. Due to the inherent characteristics of each modality, the particulars of the encoder designs exhibit subtle variations. For instance, while \textsuperscript{13}C NMR and \textsuperscript{1}H NMR spectra both pertain to nuclear magnetic resonance, \textsuperscript{1}H NMR presents a higher degree of noise in comparison to \textsuperscript{13}C NMR, along with a greater prevalence of peak overlap. In contrast, \textsuperscript{13}C NMR peaks demonstrate enhanced distinctiveness. To address these disparities, \textsuperscript{1}H NMR data may require some preprocessing to extract peak-related, condensed information which leads to a lightweight encoder that strategically aims at extracting pivotal features from the peaks. For \textsuperscript{13}C NMR, the unaltered experimental data can be employed directly. The encoder architecture for \textsuperscript{13}C NMR has more convolutional layers, tailored to capture intricate details.

\section{Results and Discussion}
\label{sec2}
\subsection{Cross-modality Retrieval}
Molecules can be represented by a variety of modalities, such as depiction images, SMILES notations, and traditional chemical characterization techniques. However, each modality possesses distinct capabilities and constraints when it comes to deciphering molecular structures. In the context of structural attributes, molecular depiction image and isomeric SMILES notation offer a detailed overview of the whole molecule from atom level to molecule level, facilitating the elucidation of their three-dimensional atomic configurations. NMR spectra provide distinct windows into the carbon and hydrogen atomic landscapes of the molecule, offering partial structural information about the molecule. Mass spectra provide crucial information about molecular mass and patterns of fragmentation, encompassing both molecular-level and motif-level information. To evaluate each ACML model, we conduct a cross-modality retrieval task: determine whether the model can accurately match a chemical modality to its corresponding graph modality from a large database of molecular graphs. Higher retrieval accuracy indicates better multi-modal coordination between graph and chemical representations. Note that the sample in the graph database never appears in training.

\begin{figure}[H]
    \centering
    \includegraphics[width=0.98\linewidth]{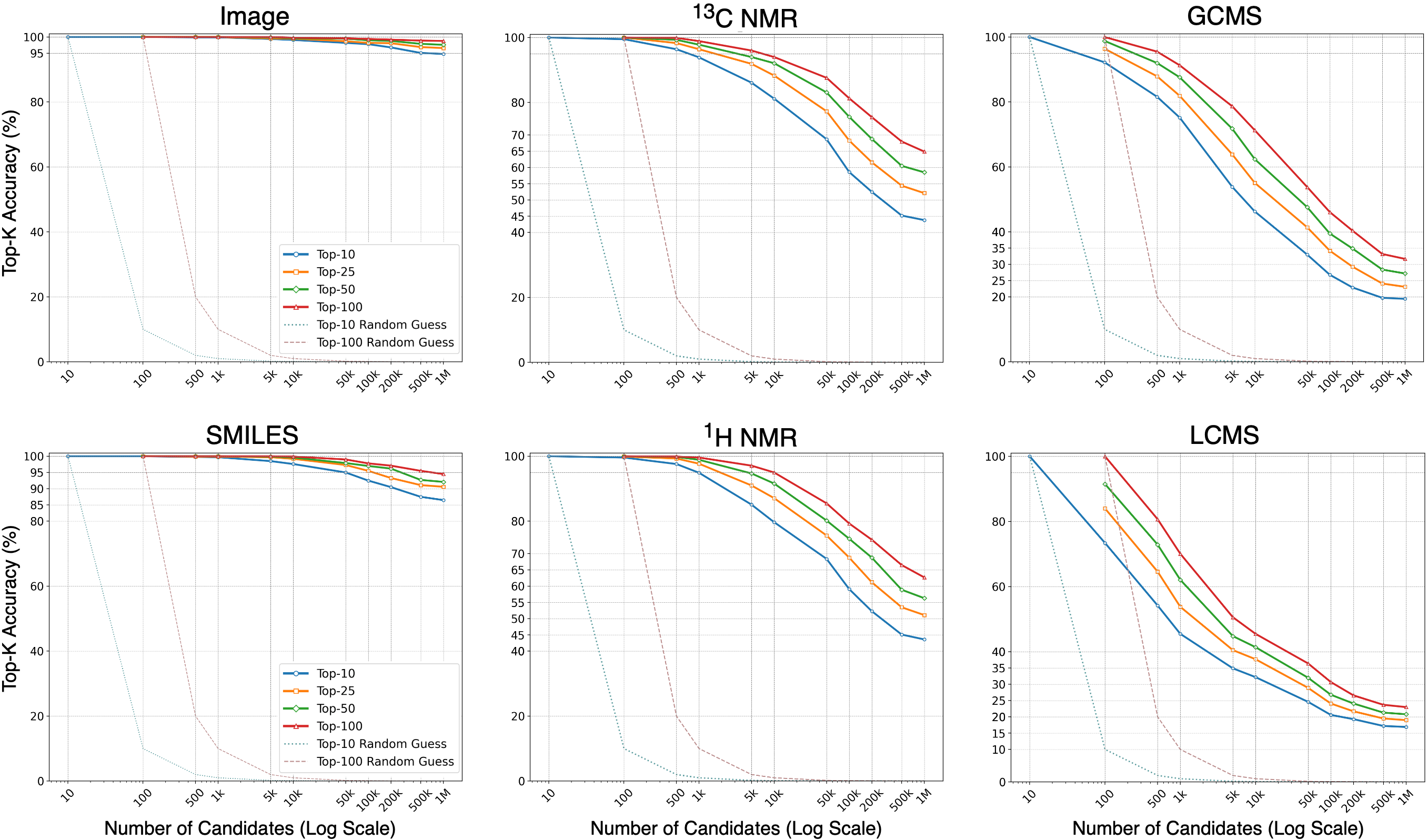}
    \caption{Cross-modality retrieval accuracy across all ACML models, including G-SMILES, G-Image, G-\textsuperscript{1}H NMR, G-\textsuperscript{13}C NMR, G-GCMS, and G-LCMS. The x-axis represents the molecular pool ranging from 1,000 (1K) to 1 million (1M) molecules on the log scale. Top-10 and top-100 accuracy of random guesses are incorporated. Each model retrieves the target molecule by ranking similarity scores within this pool. The y-axis is the top k accuracy, measuring whether the target molecule is within the k most similar molecules recommended by the model. All models consistently identify the target molecule from pools exceeding 5,000 candidates. Notably, Image- and SMILES-based models maintain over 90\% top-100 accuracy even with a pool of 1M molecules.}
    \label{fig:Topacc_all}
\end{figure}

Figure \ref{fig:Topacc_all} presents the retrieval accuracy results for all ACML models, including G-SMILES, G-Image, G-\textsuperscript{1}H NMR, G-\textsuperscript{13}C NMR, G-GCMS, and G-LCMS. These results are obtained by calculating the top k accuracy when matching one molecule's chemical modality to its corresponding graph within a graph database of varying scales. The task becomes more challenging as the scale increases.  When the scale transitions from 1,000 (1K) to 1 million (1M), the G-Image gives the best results: the top-10 accuracy drops from 100\% to 94.8\%, which only experiences a reduction of 5.2\%.  G-SMILES gives the second best results, as top 10 accuracy drops from 100\% to 86.5\%. When it comes to G-\textsuperscript{1}H NMR, the results are not as good as G-Image and G-SMILES, as expected. Specifically, when the scale transitions from 1K to 10K, the top-10 accuracy drops from 94.9\% to 79.7\%. And with 1M unseen molecules, the top-10 accuracy is 43.6\%. G-\textsuperscript{13}C NMR spectrum exhibits similar performance, as the top-10 accuracy from 93.9\% to 81.2\% until 43.8\% when the scale transitions from 1K to 10K and up to 1M. Notably, G-\textsuperscript{13}C NMR performs better than G-\textsuperscript{1}H NMR when the search space expands. Intuitively, this aligns with the expectation, since the \textsuperscript{13}C NMR peaks are typically sharp and barely overlap in small molecules. On the contrary, the \textsuperscript{1}H NMR spectrums are more prone to signal overlap, especially in complex molecular structures. Although the encoders of \textsuperscript{13}C NMR and \textsuperscript{1}H NMR have been designed to cope with such differences, the training dataset inevitably falls short of encompassing the entirety of structural variations within the zero-shot validation set. Additionally, \textsuperscript{1}H NMR might encounter sensitivity issues, especially for compounds with low hydrogen content. Nonetheless, both the G-\textsuperscript{1}H NMR and G-\textsuperscript{13}C NMR exhibit satisfactory discrimination power in identifying the most probable graphs given the NMR sequences. This demonstrates the flexibility of our ACML framework with features from different spaces and dimensions generated from different designs of encoders. 

When it comes to mass spectra modality, the results do not align as favorably as with Images, SMILES, and NMR spectra, as initially anticipated. For G-GCMS, when the scale transitions from 1K to 10K, the top-10 accuracy experiences a certain degree of reduction from 75.2\% to 46.3\%. When the validation database extends to 1M, the top-10 accuracy dropped to 19.4\%. For G-LCMS, when the scale transitions from 1K to 10K, the top 10 accuracy experiences a certain degree of reduction from 45.5\% to 32.2\%. When the validation database extends to 1M, the top-10 accuracy dropped to 16.9\%. One may observe that G-GCMS has better performance on molecular recognition than G-LCMS. The reason for that is GCMS is typically carried out under harsh conditions, resulting in a wide array of fragmentation that unveils the molecular composition. In contrast, LCMS is generally performed under milder conditions, resulting in reduced fragmentation. Particularly, LC-MS/MS primarily concentrates on localized information near the precise molecular weight. However, in terms of cross-modality retrieval, LCMS offers less comprehensive information.   

\subsection{Isomer Discrimination}
\paragraph{Task Description.} Besides calculating accuracy among a large group of molecules, in which many molecules could be intrinsically different, isomer recognition stands as one of the most challenging tasks in molecular identification due to the intricacy of isomeric structures. Distinguishing between stereoisomers, which share the same connectivity but differ in spatial arrangement, poses a particular challenge and demands advanced analytical methods and meticulous data interpretation.

To thoroughly evaluate the model's discriminative power, we selected 140 isomer pairs from the test dataset in a systematic way, including both structural isomers and stereoisomers. First, all SMILES representations were converted to molecular formulas using the \texttt{RDKit} \cite{landrum2013rdkit} package in Python. Molecules sharing the same molecular formula were grouped, and pairs within each group were identified as isomers. To ensure meaningful comparisons, we visualized the isomer pairs and excluded those with significantly different structures (e.g., ring vs.\ chain) while retaining only pairs with high structural similarity (differing by fewer than three bonds). From the remaining isomer pairs, we randomly selected 40 stereoisomers and 100 structural isomers to form a representative test set.
The selected pairs and human evaluation procedures are provided in \textit{\textbf{Supporting Information S3}}, with the isomer distribution shown in Table~\ref{tab:isomer_dist}.

\paragraph{Results and Analysis.} Both human experts and the ACML model were tasked with associating NMR spectra to the most likely molecular isomers. Besides the binary classifications, the ACML model produces confidence scores that indicate the alignment strength between the NMR spectra and potential molecular structures. Higher confidence scores represent stronger alignment using our proposed framework. When comparing the expert's performance across different NMR spectroscopy techniques, the expert demonstrated better accuracy using $^1\text{H}$ NMR compared to $^13\text{c}$ NMR spectra. This is consistent with the fact that protons have the highest natural abundance ($\sim$ 99\%), resulting in denser $^1\text{H}$ NMR signals that provide more information for distinguishing between isomers. In contrast, \textsuperscript{13}C's lower natural abundance ($\sim$ 1\%) means fewer signals are available for analysis.
The performance comparison between the model and the expert is presented in Table \ref{tab:isomer_perf}. Our model outperforms the human expert for both \textsuperscript{1}H NMR and \textsuperscript{13}C NMR tasks, with a much more significant lead for \textsuperscript{13}C NMR tasks. This demonstrates the effectiveness of representation learning using our proposed framework. 
\begin{table}[t!]
\centering
\footnotesize
\caption{Analysis of Isomer Distribution, Similarity, and Performance.}
\label{tab:merged_isomer_table}
\begin{subtable}[t]{1.0\textwidth}
\centering
\caption{Distribution and Similarity of Isomer Pairs. Tanimoto Similarity \cite{tanimoto1958elementary} is used to measure the similarity between chemical structures.}
\label{tab:isomer_dist}
\begin{tabular}{ccc}
\toprule
\textbf{} & \textbf{Structural Isomer Count \tiny{(Avg. Similarity)}} & \textbf{Stereoisomer Count \tiny{(Avg. Similarity)}} \\
\midrule
\textsuperscript{1}H NMR  & 47 (0.43) & 15 (1.00) \\
\textsuperscript{13}C NMR & 61 (0.47) & 17 (1.00) \\
\bottomrule
\end{tabular}
\end{subtable}

\vspace{0.5cm} % Adjust spacing between subtables

\begin{subtable}[t]{1.0\textwidth}
\centering
\caption{Performance Comparison between the Model and the Expert.}
\label{tab:isomer_perf}
\begin{tabular}{ccccc}
\toprule
\textbf{} & \multicolumn{2}{c}{{\textsuperscript{1}H NMR}} & \multicolumn{2}{c}{{\textsuperscript{13}C NMR}} \\
\cmidrule(lr){2-3} \cmidrule(lr){4-5}
\textbf{} & Structural Isomer & Stereoisomer & Structural Isomer & Stereoisomer \\
\midrule
\textbf{Model Accuracy} & 82.6\% & 51.6\% & 82.9\% & 60.0\% \\
\textbf{Human Accuracy} & 78.0\% & 51.6\% & 51.9\% & 57.1\% \\
\bottomrule
\end{tabular}
\end{subtable}
\end{table}
% \begin{table}[t!]
% \centering
% \footnotesize
% \caption{Distribution and Similarity of isomer pairs. Tanimoto Similarity \cite{tanimoto1958elementary} is used to measure the similarity between chemical structures.}
% \label{tab:isomer_dist}
% \begin{tabular}{l|c|c}
% \toprule
% \textbf{}          & {\textbf{Structural Isomer}} & {\textbf{Stereoisomer}} \\ 
% \hline
% \textbf{}     & Count (Avg. Similarity) & Count (Avg. Similarity) \\\hline
% \textsuperscript{1}H NMR    & 47   (0.43)      & 15  (1.00)\\ 
% \hline
% \textsuperscript{13}C NMR   & 61   (0.47)      & 17  (1.00)  \\ \bottomrule
% \end{tabular}
% \end{table}

% \begin{table}[t!]
% \centering
% \footnotesize
% \caption{Performance Comparison between the Model and the Expert.}
% \label{tab:isomer_perf}
% \begin{tabular}{l|cc|cc}
% \toprule
% \textbf{}          & \multicolumn{2}{c|}{\textbf{\textsuperscript{1}H NMR}} & \multicolumn{2}{c}{\textbf{\textsuperscript{13}C NMR}} \\ \hline
% \textbf{}          & \textbf{Structural Isomer} & \textbf{Stereoisomer} & \textbf{Structural Isomer} & \textbf{Stereoisomer} \\ \hline
% Model Accuracy     & 82.6\%  & 51.6\%  & 82.9\% & 60.0\%                 \\ \hline
% Human Accuracy    & 78.0\%  & 51.6\%  & 51.9\% & 57.1\%                 \\
% \bottomrule
% \end{tabular}
% \end{table}

To visualize the model's performance, four cases of \textsuperscript{13}C NMR and four cases \textsuperscript{1}H NMR examples are shown in \textbf{\textit{Supporting Information S2: Figure S1- S2}}. In each figure, the proposed model correctly identifies 3 pairs out of the 4 challenging isomer pairs. The cases in which the model failed to identify the candidate involved NMR spectra that were extremely similar. In the example of 2A–2B in Figure S1, the molecules differ only in a small region of their structure, leading to minimal differences in the \textsuperscript{13}C NMR spectrum.  Besides, 4A–4B in Figure S2 exhibit notable structural differences, yet their \textsuperscript{1}H NMR spectra remain nearly identical, as the variations do not significantly affect the proton environments captured by \textsuperscript{1}H NMR.

\subsection{Visualization of Molecular Representations}
\label{sec:embedding-visualizations}
% Describe the target of this experiment
\paragraph{Task Description.} In the ACML framework, the graph modality serves as an information receptor capable of bringing forth the information or chemical rules from another modality via coordination. Thus, it may reveal distinct dimensions of chemical significance among diverse chemical modalities. To illuminate variations of chemical semantics embedded in hidden graph representations acquired from various ACML frameworks, we computed graph embeddings of previously unseen 30,000 molecules using pre-trained graph encoders for analysis. Given 8 important chemical characteristics in drug discovery: molecular weight (MW), the logarithmic partition coefficient (LogP), hydrogen-bonding acceptor (HBA), hydrogen-bonding donor (HBD), topological polar surface area (PSA), the number of rotatable bonds (\#R-Bonds), and the number of stereogenic (chiral) centers (\#C-Centers), we decided to examine how these characteristics correlate with the graph embeddings. The scope of this inquiry extends to understanding the impact of these modalities on crucial parameters within the domain of drug discovery, highlighting the strengths of each molecular modality.

\begin{figure}[H]
    \centering\includegraphics[width=1\linewidth]{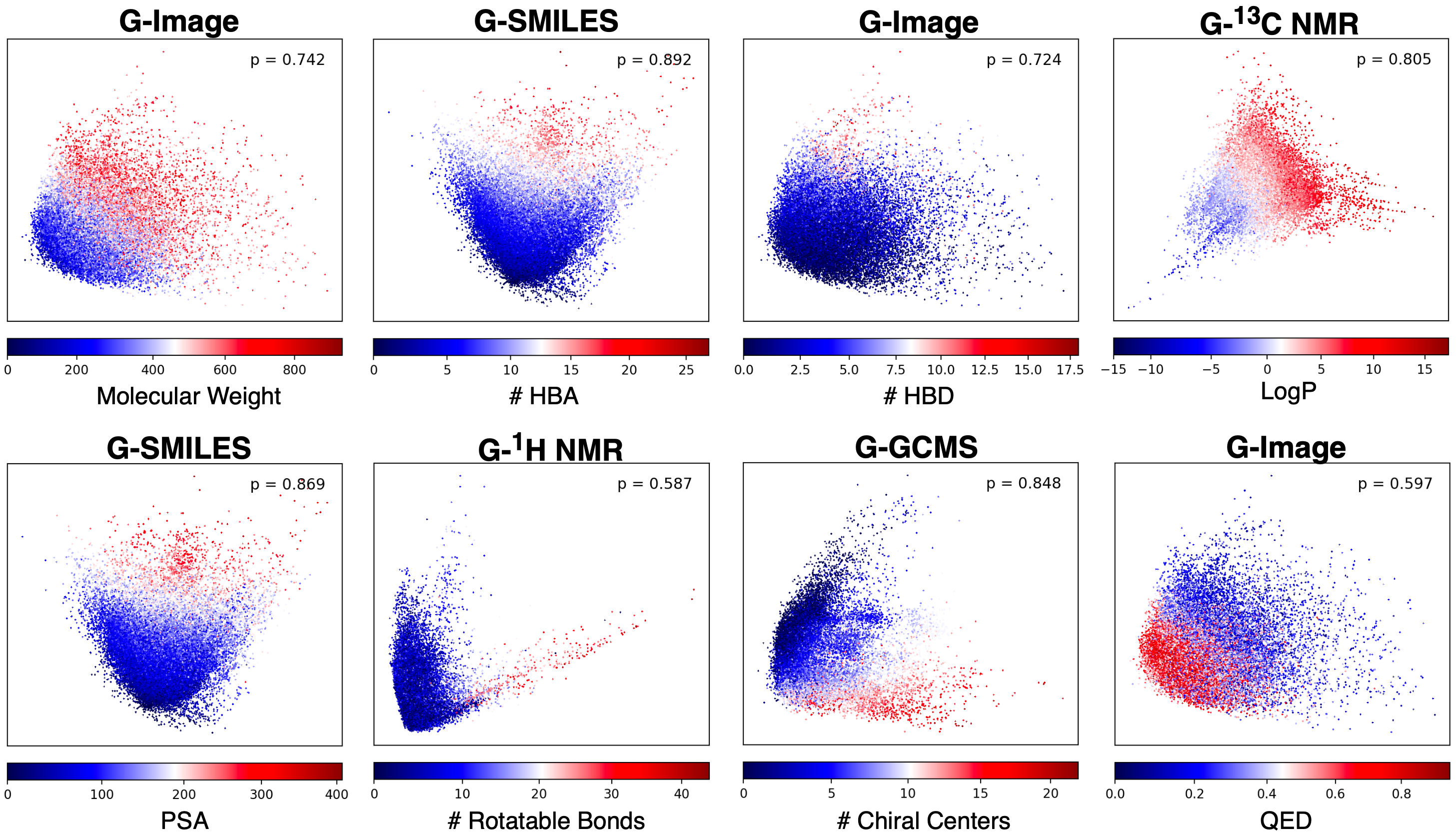}
    \caption{Graph embedding visualization via PCA. For each property, one representative figure is shown using the ACML instantiation that achieves the highest PCC score.}
    \label{fig:embed_visual}
\end{figure}

\paragraph{Evaluations and Baselines.} We decomposed the feature dimensions via PCA \cite{jackson2005user} and mapped the resulting embeddings onto a two-dimensional plane. The outcomes were visualized as 2D points with coloring based on their corresponding chemical property values. Figure~\ref{fig:embed_visual} presents the visualization results. Notably, a remarkable consistency in the distribution of points concerning the chemical property values was observed. This observation underscores the proficiency of the pre-trained graph encoder in capturing the inherent chemical semantics. We then conducted a quantitative evaluation of the relationship between graph embedding and the chemical properties using Pearson correlation coefficients (PCC) \cite{pearson1920notes}. In our approach, we first established a linear regression model to determine the optimal linear combination of two PCA components that maximizes the Pearson correlation score between the linear combination of PCA components and chemical properties of interest.  We also showed the graph embeddings from other pretraining graph encoders and performed PCA in the same fashion to compare how the latent representations differ with respect to ACML. Among the baselines, L2P-GNN \cite{lu2021learning} and MGSSL \cite{zhang2021motif} develop motif-level self-supervised tasks during pretraining, while MolCLR \cite{wang2022molecular} and MoMu \cite{su2022molecular} use contrastive learning in pretraining.

% 

% To highlight variations across a range of modalities, we present the translation of chemical principles into a Graph Neural Network (GNN) with our ACML model. We employ both molecular GNN unimodal embeddings and higher-level abstract projection embeddings to explore how chemical rules are represented within these embeddings. 

% Description of experiments. 
%In order to condense the feature dimensions, we employed PCA \cite{jackson2005user} and mapped these embeddings onto a two-dimensional plane. The outcomes were visualized as 2D points with coloring based on their corresponding chemical property values. Figure~\ref{fig:embed_visual} presents the visualization results. Notably, a remarkable consistency in the distribution of points concerning the chemical property values was observed. This observation underscores the proficiency of the pre-trained graph encoder in capturing the inherent chemical semantics.

\paragraph{Results and Analysis.} Table~\ref{tab:correlation-value} presents the PCC scores of all ACML frameworks and baselines. All graph embedding visualizations are provided in \textit{\textbf{Supporting Information S2: Figure S3-S13}}. The knowledge of the chemical modalities was captured and highlighted by the degree of correlation, particularly with respect to key properties relevant to drug discovery. Among the methods analyzed, the ACML frameworks generally outperform baseline models, achieving the highest PCCs in seven out of eight properties, with the exception of \#R-Bonds, where MoMu achieves the top PCC (0.617).
Regarding MW, G-Image and G-SMILES are leading performers. Molecular weights can effectively be expressed by the pixels of molecular depiction image, and structural motifs encoded by SMILES. These account for the superior performance of these two models in this context. Additionally, G-GCMS and G-LCMS also demonstrate competitive performances. This can be attributed to their direct measurement of molecular mass in spectrometry.
Regarding \#HBA, G-SMILES emerges as the top performer. It is expected because electronegative HBA atoms, such as oxygen (O) or nitrogen (N), are directly denoted in SMILES. 
Regarding \#HBD, G-Image is the top-performing model, as HBD groups, such as hydroxyl groups (-OH), amine groups (-NH2), and carboxylic acids (-COOH), are explicitly illustrated in the images. However, hydrogens are not always contained in SMILES, resulting in G-SMILES has a slightly lower performance. 
Regarding LogP, G-\textsuperscript{13}C NMR is the game winners, followed by the MGSSL (BFS) baseline. This outcome is attributable to the fact that LogP measures the solubility of a solute in both water and organic solvents, making it highly reliant on electronic information within the molecule. This electronic information is well captured by the \textsuperscript{13}C NMR techniques, and it remains concealed in other modalities such as Image, SMILES, or mass spectrometry.
Regarding PSA, G-SMILES achieves the highest, followed closely by G-Image. The strong performance of G-SMILES reflects its explicit representation of electronegative atoms, such as oxygen and nitrogen, which directly contribute to PSA. Besides, G-Image captures polar regions through pixel-based molecular depictions.
Regarding \#R-Bonds, MoMu demonstrates the highest PCC, while G-\textsuperscript{1}H NMR closely follows as the second best and significantly outperforms all other modalities. The reason is the rotation of chemical bonds can be derived from the magnetic equivalence of vicinal and geminal protons, while not revealed in other modalities.  
% \textcolor{blue}{Regarding \#C-Centers, G-GCMS achieves the best PCC of 0.848, closed followed by G-SMILES with a 0.830. This strong correlation can be elucidated by the inherent advantage of GCMS and SMILES: in GCMS, the subtle mass fragmentation differences in chiral compounds enable GCMS to capture chirality information effectively; in SMILES, atoms with chiral centers are directly annotated in SMILES representations, that is, ``@" represents chirality with inward direction and ``@@" represents chirality with outward direction.}

\textbf{Remark 1.} Note that PCC only measures the linear correlation and doesn't account for any nonlinear associations, a low PCC score doesn't necessarily imply a lack of correlation. For example, when we examined the embedding visualization of G-\textsuperscript{1}H NMR for the "\#R-Bonds" task shown in Figure~\ref{fig:embed_visual}, we observed clear continuous color distributions, despite the PCC value being relatively low at 0.587.

\textbf{Remark 2.} These 8 chemical properties play vital roles in drug discovery. If the graph neural network is able to reveal this information, it offers the potential for deep models to understand the basic philosophy of molecular design and would even make breakthroughs in drug discovery. Note that these properties are never seen in ACML learning and the molecules we selected are also never seen in training, the surprising findings demonstrate our ACML framework can make a transition of chemical meaning to graph encoder, enhancing its interpretability. 

\begin{table}[t]
\centering
\footnotesize
\caption{Pearson correlation test results between graph representations and the chemical property of interest. The highest scores are marked as bold while the second highest are underlined. A higher score indicates that the low-dimensional graph representations have a higher linear correlation with the chemical property of interest.}
\label{tab:correlation-value}
\begin{tabular}{lccccccccc}
\toprule
& \textbf{MW} & \textbf{\#HBA} & \textbf{\#HBD} & \textbf{LogP} & \textbf{PSA} & \textbf{\#R-Bonds} & \textbf{\#C-Centers} & \textbf{QED} \\ 
\hline
L2P-GNN & 0.053 & 0.026 & 0.027 & 0.029 & 0.026 & 0.024 & 0.237 & 0.056 \\
MGSSL (BFS) & 0.259 & 0.639 & 0.618 & \underline{0.700} & 0.670 & 0.331 & 0.789 & 0.371 \\
MGSSL (DFS) & 0.255 & 0.638 & 0.504 & 0.673 & 0.649 & 0.266 & 0.759 & 0.381 \\
MoMu & 0.263 & 0.305 & 0.225 & 0.386 & 0.280 & \textbf{0.617} & 0.609 & 0.213 \\
MolCLR & 0.228 & 0.294 & 0.207 & 0.164 & 0.244 & 0.378 & 0.558 & 0.182 \\
\hline
G-Image & \textbf{0.742}  & 0.671 & \textbf{0.724} & 0.355 & \underline{0.736} & 0.325 & 0.630 & \textbf{0.597} \\
G-SMILES & \underline{0.623}  & \textbf{0.892} & \underline{0.716} & 0.564 & \textbf{0.869} & 0.244 & \underline{0.830} & \underline{0.492} \\
G-\textsuperscript{1}H NMR & 0.202 & 0.361 & 0.227 & 0.588 & 0.303 & \underline{0.587} & 0.388 & 0.235 \\
G-\textsuperscript{13}C NMR & 0.195 & 0.673 & 0.569 & \textbf{0.805} & 0.659 & 0.277 & 0.641 & 0.188 \\
G-GCMS & 0.553 & \underline{0.676} & 0.623 & 0.444 & 0.670 & 0.076 & \textbf{0.848} & 0.435 \\
G-LCMS & 0.522 & 0.403 & 0.317 & 0.357 & 0.385 & 0.144 & 0.669 & 0.223 \\
\bottomrule

% proj \\
% \hline
% Image & \textbf{0.9176} & 0.7391 & 0.5266 & 0.1253 & 0.6731 & 0.3471 & 0.7643 & \textbf{0.6116} \\
% Smiles & 0.6449 & 0.7640 & 0.4383 & 0.3434 & 0.7166 & 0.1688 & 0.7991 & 0.4705 \\
% HNMR & 0.0842 & 0.5095 & 0.4007 & 0.5963 & 0.4440 & 0.0470 & 0.4217 & 0.0373 \\
% CNMR & 0.2226 & 0.6637 & 0.5541 & 0.7495 & 0.6439 & 0.1070 & 0.7006 & 0.2039 \\
% GCMS & 0.2232 & 0.5735 & 0.4733 & 0.5483 & 0.5475 & 0.2538 & 0.6341 & 0.3654 \\
% LCMS & 0.2866 & 0.4421 & 0.3074 & 0.2955 & 0.4242 & 0.1792 & 0.6091 & 0.0987 \\
% \hline
\end{tabular}
\end{table}

%To exhibit intuitively the learned graph representations from the ACML framework, we mapped them to a two-dimensional space using dimension reduction methods such as PCA and t-SNE. Note that we random selected 50,000 molecules from the PubChem database (\textcolor{blue}{To be confirmed}) for visualizations, which is zero shot. There are 5 molecular properties considered in this study, including. (\textcolor{blue}{How these properties being calculated and why these properties are important.}) Note that these property related information is never involved in the multimodal learning phase. 

%rule of 5: Adhering to the Rule of Five entails satisfying criteria like molecular weight under 500, maintaining hydrogen-bonding donors below 5, containing hydrogen-bonding acceptors fewer than 10, ensuring logP is less than 5, and constraining rotatable bonds to 10 or fewer.

\subsection{Molecular Property Prediction Tasks}
\label{sec:property-prediction}
Molecular property prediction is a critical graph-level prediction task used to assess the generalization capability of pre-trained graph encoders. In other words, it evaluates how well the encoder's learned representations can be adapted to downstream tasks. Additionally, finetuning on a small dataset is challenging because models primarily depend on their pre-training capabilities in this scenario. By using a limited dataset, we can better assess the model's representation learning capabilities and understand how well the pre-trained knowledge transfers to specific tasks. Thus, we intentionally choose to finetune the model using relatively small datasets to evaluate the pretrained graph encoder from the ACML framework across a range of downstream tasks. 
\subsubsection{Experimental Settings}
\paragraph{Datasets.} We employ two key molecular benchmark resources. The first, MoleculeNet \cite{wu2018moleculenet}, is a commonly used benchmark specially designed for testing machine learning methods of molecular properties. The second, Therapeutics Data Commons (TDC) \cite{huang2022artificial}, is a recent platform providing molecular datasets and learning tasks for therapeutics. The molecular data of TDC are in SMILES format, we convert them into molecular graphs using OGB package \cite{hu2020ogb}. Our focus is on tasks involving small molecule property prediction, with an emphasis on properties relevant to drug discovery. Therefore, in MoleculeNet, we consider 6 classification and 2 regression datasets. In TDC, we consider 12 binary classification datasets about (ADME) properties (absorption, distribution, metabolism, and excretion). A small-model drug needs to travel from the site of administration (e.g., oral) to the site of action (e.g., a tissue) and then decompose, and exit the body. To do that safely and efficaciously, the chemical is required to have numerous ideal ADME properties. For classification tasks, ROC-AUC is used as the evaluation metric, and for regression tasks, RMSE is used. 

\paragraph{Models and Baselines.} For each task, we employ the GIN \cite{xu2018how} with 5 layers as the backbone graph encoder, initialized with weights pre-trained by the ACML framework. We compare our models with other pretraining baselines on molecular graph learning. Among the baselines, L2P-GNN \cite{lu2021learning} and MGSSL \cite{zhang2021motif} with different generation orders (BFS and DFS), MotifConv \cite{wang2023motif} propose self-supervised tasks within graphs to capture substructural or motif-level information during pretraining, while MolCLR \cite{wang2022molecular} and MoMu \cite{su2022molecular} develop contrastive learning using augmented graphs or other modalities during pretraining. MotifConv was excluded from TDC experiments due to its high complexity during pretraining. For a fair comparison, all baselines also use the 5-layer GIN model as the backbone graph encoder. Experiments using GIN without pretraining are also conducted for comparison. More experimental details can be found in \textbf{\textit{Supporting Information S1: Experimental Details}}. For each dataset, the average and standard deviation of model performance are reported based on 5 independent runs.

\subsubsection{Results and Analysis}
\paragraph{MoleculeNet Datasets.}
Table~\ref{tab:finetune-results} shows the test performance on MoleculeNet. First, by comparing models without pretrainnig, we observed that the best ACMLs generally outperform by big margins. Specifically, G-Image makes improvements in all datasets, including an average increase in AUC-ROC of 10.0\% in BACE, 16.4\% in Clintox, 7.9\% in HIV, and reductions in RMSE with a 33.3\% average decrease in ESOL and a 47.9\% decrease in FreeSolv. These results suggest that contrastive learning with G-Image effectively captures useful information in molecular depictions, leading to a better understanding of the overall molecular structure. 
Second, when compared with other self-supervised pretraining strategies, the proposed ACML framework achieves the best performance on 7 of 8 datasets. These improvements indicate that our ACML framework is a powerful self-supervised learning strategy, which is easy to implement and requires minimal manual design of self-supervised tasks. 
Third, by comparing ACMLs using different chemical modalities as inputs, we observed that G-Image generally outperforms other modalities as it achieves the best performance in BACE, Sider, and ESOL, and ranks second in FreeSolv. Although G-Image does not lead in BBBP, Tox21 and HIV, its performance is close to the top. This is likely due to its ability to capture comprehensive molecular structure information, which is crucial for most molecular property prediction tasks. These observations on the superiority of G-Image are also consistent with the analysis in the Section of~\nameref{sec:embedding-visualizations}, where graph representations from G-Image exhibit strong correlations with various chemical properties relevant to drug discovery.
In contrast, other modalities provide specific insights that could be beneficial for certain tasks but not universally effective. For instance, G-\textsuperscript{1}H NMR achieves the highest average ROC-AUC of 83.9\% in Clintox, 81.6\% in Tox21, and 84.9\% in HIV, but performs worse compared to the no-pre-trained models in BBBP (-3.3\%) and Sider (-3.3\%). Similar cases are observed for other tasks and other modalities.

\begin{table}[H]
\centering
\tiny
\caption{Compare test performance (mean $\pm$ std) on molecular property prediction benchmarks. RMSE is used for evaluations on ESOL and FreeSolv datasets. ROC-AUC (\%) is used for evaluations on all remaining datasets. The best result for each dataset is in \textbf{bold}, with the second best \underline{underlined}.}

\begin{subtable}[t]{1.0\textwidth}
\setlength{\tabcolsep}{2pt}
\centering
\caption{Results in MoleculeNet.}
\label{tab:finetune-results}
\scalebox{1.2}{
\begin{tabular}{lcccccccc}
\toprule
\textbf{Model} & \textbf{BACE $\uparrow$} & \textbf{BBBP $\uparrow$} & \textbf{Clintox $\uparrow$} & \textbf{Sider $\uparrow$} & \textbf{Tox21 $\uparrow$} & \textbf{HIV $\uparrow$} & \textbf{ESOL $\downarrow$} & \textbf{FreeSolv $\downarrow$} \\
\midrule
GIN(no pretrain) & 79.7 $\pm$ 4.9 & 87.3 $\pm$ 3.6 & 53.0 $\pm$ 6.5 & 61.6 $\pm$ 2.5 & 78.3 $\pm$ 2.4 & 76.2 $\pm$ 5.8 & 1.26 $\pm$ 0.16 & 4.67 $\pm$ 1.55 \\
L2P-GNN & 81.9 $\pm$ 0.4 & 87.0 $\pm$ 0.8 & 54.0 $\pm$ 2.4 & 61.7 $\pm$ 1.8 & 77.4 $\pm$ 0.6 & 78.0 $\pm$ 2.6 & 1.77 $\pm$ 0.48 & 3.73 $\pm$ 1.32 \\
MGSSL (DFS) & 79.7 $\pm$ 0.8 & 70.5 $\pm$ 1.1 & 79.7 $\pm$ 2.2 & 60.5 $\pm$ 0.7 & 76.4 $\pm$ 0.4 & 79.5 $\pm$ 1.1 & 0.91 $\pm$ 0.10 & 3.40 $\pm$ 0.90 \\
MGSSL (BFS) & 79.1 $\pm$ 0.9 & 69.7 $\pm$ 0.1 & \underline{80.7} $\pm$ \underline{2.1} & 61.8 $\pm$ 0.8 & 76.5 $\pm$ 0.3 & 78.8 $\pm$ 1.2 & \underline{0.86} $\pm$ \underline{0.06} & 2.60 $\pm$ 1.00 \\
MotifConv & 82.0 $\pm$ 5.5 & \textbf{90.0} $\pm$ \textbf{3.1} & 65.5 $\pm$ 13.9 & \underline{62.7} $\pm$ \underline{2.8} & 80.2 $\pm$ 1.5 & 80.0 $\pm$ 4.3 & 1.40 $\pm$ 0.15 & 3.70 $\pm$ 0.40 \\
MoMu & 72.1 $\pm$ 0.3 & 70.2 $\pm$ 0.4 & 71.1 $\pm$ 3.0 & 56.1 $\pm$ 0.4 & 74.5 $\pm$ 0.5 & 77.2 $\pm$ 1.5 & 1.41 $\pm$ 0.10 & 3.37 $\pm$ 0.75 \\
MolCLR & 74.6 $\pm$ 1.1 & 68.0 $\pm$ 2.0 & 62.0 $\pm$ 2.7 & 56.3 $\pm$ 2.2 & 73.9 $\pm$ 0.5 & 72.1 $\pm$ 1.3 & 1.35 $\pm$ 0.21 & 2.82 $\pm$ 0.91 \\
\midrule
G-Image & \textbf{89.7} $\pm$ \textbf{0.8} & \underline{87.4} $\pm$ \underline{1.0} & 69.4 $\pm$ 2.3 & \textbf{62.9} $\pm$ \textbf{1.6} & 80.9 $\pm$ 0.8 & 84.1 $\pm$ 1.7 & \textbf{0.84} $\pm$ \textbf{0.00} & \underline{2.43} $\pm$ \underline{0.00} \\
G-SMILES & 85.6 $\pm$ 0.5 & 83.7 $\pm$ 0.5 & 68.5 $\pm$ 0.4 & 58.6 $\pm$ 0.4 & 81.0 $\pm$ 0.2 & 82.8 $\pm$ 0.7 & 0.94 $\pm$ 0.00 & \textbf{2.34} $\pm$ \textbf{0.00} \\
G-\textsuperscript{1}H NMR & 86.6 $\pm$ 1.3 & 84.0 $\pm$ 3.8 & \textbf{83.9} $\pm$ \textbf{7.1} & 58.3 $\pm$ 0.9 & \textbf{81.6} $\pm$ \textbf{0.1} & \textbf{84.9} $\pm$ \textbf{0.9} & 1.06 $\pm$ 0.08 & 2.66 $\pm$ 0.30 \\
G-\textsuperscript{13}C NMR & \underline{87.5} $\pm$ \underline{1.1} & 81.7 $\pm$ 3.2 & 75.6 $\pm$ 6.0 & 54.8 $\pm$ 0.8 & \underline{81.6} $\pm$ \underline{0.5} & \underline{84.4} $\pm$ \underline{0.7} & 1.05 $\pm$ 0.04 & 3.34 $\pm$ 0.31 \\
G-GCMS & 84.7 $\pm$ 2.1 & 79.1 $\pm$ 2.1 & 67.7 $\pm$ 8.9 & 60.3 $\pm$ 2.7 & 79.9 $\pm$ 1.1 & 83.7 $\pm$ 0.9 & 1.12 $\pm$ 0.04 & 3.81 $\pm$ 0.28 \\
G-LCMS & 79.7 $\pm$ 4.7 & 86.5 $\pm$ 3.4 & 62.3 $\pm$ 9.6 & 60.9 $\pm$ 0.6 & 81.4 $\pm$ 0.8 & 82.2 $\pm$ 0.5 & 1.11 $\pm$ 0.14 & 3.89 $\pm$ 0.20 \\
\bottomrule
\end{tabular}}
\end{subtable}

\vspace{1em}
\begin{subtable}[t]{1.0\textwidth}
\setlength{\tabcolsep}{5.6pt}
\caption{Results in Therapeutics Data Commons (TDC).}
\label{table:tdc}
\scalebox{1.2}{
\begin{tabular}{l>{\centering\arraybackslash}p{1.5cm}>{\centering\arraybackslash}p{1.5cm}>{\centering\arraybackslash}p{1.5cm}>{\centering\arraybackslash}p{1.5cm}>{\centering\arraybackslash}p{1.5cm}>{\centering\arraybackslash}p{1.5cm}}
\toprule
 & \textbf{HIA} $\uparrow$ & \begin{tabular}[t]{@{}c@{}} \textbf{PAMPA} \ $\uparrow$\\ \textbf{Permeability} \end{tabular}  & \begin{tabular}[t]{@{}c@{}} \textbf{Bio} \ $\uparrow$ \\ \textbf{Availability} \end{tabular}  & \begin{tabular}[t]{@{}c@{}} \textbf{PGP} \\ \textbf{Inhibition}  \end{tabular} $\uparrow$ & \begin{tabular}[t]{@{}c@{}} \textbf{CYP2C19} \\ \textbf{Inhibition} \end{tabular} $\uparrow$ & \begin{tabular}[t]{@{}c@{}} \textbf{CYP2D6} \\ \textbf{Inhibition} \end{tabular} $\uparrow$ \\
\midrule
GIN (no pretrain) & 93.5 $\pm$ 1.7 & 67.1 $\pm$ 1.1 & 63.4 $\pm$ 0.9 & 89.1 $\pm$ 0.6 & 87.4 $\pm$ 0.3 & 84.8 $\pm$ 0.4 \\
L2P-GNN & 82.8 $\pm$ 2.0 & 50.3 $\pm$ 2.0 & 61.7 $\pm$ 1.9 & 76.6 $\pm$ 1.9 & 83.1 $\pm$ 0.2 & 80.2 $\pm$ 0.3 \\
MGSSL (BFS) & \underline{96.6} $\pm$ \underline{0.5} & 70.9 $\pm$ 0.9 & 62.1 $\pm$ 0.3 & 90.3 $\pm$ 0.6 & 88.0 $\pm$ 0.1& 85.0 $\pm$ 0.1 \\
MGSSL (DFS) & 91.8 $\pm$ 1.0 & 65.4 $\pm$ 4.4 & 59.9 $\pm$ 3.6 & 87.9 $\pm$ 0.4 & 87.2 $\pm$ 0.2 & 83.6 $\pm$ 0.4 \\
MoMu & 91.1 $\pm$ 2.7 & 67.5 $\pm$ 2.0 & 61.6 $\pm$ 0.8 & 88.8 $\pm$ 0.4 & 86.4 $\pm$ 0.1 & 83.2 $\pm$ 0.4 \\
MolCLR & 94.0 $\pm$ 1.5 & 70.1 $\pm$ 2.9 & 65.0 $\pm$ 2.0 & 88.0 $\pm$ 0.6 & 87.1 $\pm$ 0.4 & 85.1 $\pm$ 0.5 \\
\midrule
G-Image & 93.2 $\pm$ 1.0 & 69.4 $\pm$ 1.5 & 67.8 $\pm$ 7.0 & 90.5 $\pm$ 0.9 & 88.1 $\pm$ 0.2 & 85.2 $\pm$ 0.3 \\
G-SMILES & 95.5 $\pm$ 0.9 & \underline{71.5} $\pm$ \underline{2.6} & 64.1 $\pm$ 8.9 & \textbf{92.3} $\pm$ \textbf{2.9} & \textbf{88.5} $\pm$ \textbf{0.2} & \textbf{86.8 $\pm$ 0.1} \\
G-\textsuperscript{1}H NMR & 94.9 $\pm$ 2.2 & \textbf{72.7} $\pm$ \textbf{2.6} & \textbf{72.3} $\pm$ \textbf{2.5} & 90.3 $\pm$ 2.4 & 88.3 $\pm$ 0.3 & \underline{85.7} $\pm$ \underline{0.6} \\
G-\textsuperscript{13}C NMR & 96.5 $\pm$ 4.7 & 70.9 $\pm$ 2.9 & 67.5 $\pm$ 3.1 & 90.7 $\pm$ 1.9 & \underline{88.4} $\pm$ \underline{0.5} & 85.6 $\pm$ 0.8 \\
G-GCMS & \textbf{96.9} $\pm$ \textbf{1.6} & 67.5 $\pm$ 3.1 & \underline{68.0} $\pm$ \underline{2.6} & \underline{91.3} $\pm$ \underline{0.8} & 87.9 $\pm$ 0.2 & 84.3 $\pm$ 0.8 \\
G-LCMS & 92.2 $\pm$ 0.2 & 64.7 $\pm$ 2.5 & 67.9 $\pm$ 1.8 & 90.9 $\pm$ 0.5 & 87.7 $\pm$ 0.3 & 83.7 $\pm$ 0.3 \\
\bottomrule
\end{tabular}}

\vspace{1em} % Add some space between the two sections
%\scalebox{1.03}{\begin{tabular}{lcccccc}
\scalebox{1.2}{\begin{tabular}{l>{\centering\arraybackslash}p{1.5cm}>{\centering\arraybackslash}p{1.5cm}>{\centering\arraybackslash}p{1.5cm}>{\centering\arraybackslash}p{1.5cm}>{\centering\arraybackslash}p{1.5cm}>{\centering\arraybackslash}p{1.5cm}}
\toprule
 & \begin{tabular}[t]{@{}c@{}} \textbf{CYP3A4} \\ \textbf{Inhibition} \end{tabular} $\uparrow$ &\begin{tabular}[t]{@{}c@{}} \textbf{CYP1A2} \\ \textbf{Inhibition} \end{tabular} $\uparrow$ & \begin{tabular}[t]{@{}c@{}} \textbf{CYP2C9} \\ \textbf{Inhibition} \end{tabular} $\uparrow$& \begin{tabular}[t]{@{}c@{}} \textbf{CYP2C9} \\ \textbf{Substrate} \end{tabular} $\uparrow$ & \begin{tabular}[t]{@{}c@{}} \textbf{CYP2D6} \\ \textbf{Substrate} \end{tabular} $\uparrow$ & \begin{tabular}[t]{@{}c@{}} \textbf{CYP3A4} \\ \textbf{Substrate} \end{tabular} $\uparrow$ \\
\midrule
GIN (no pretrain) & 86.5 $\pm$ 0.5 & 92.0 $\pm$ 0.3 & 86.5 $\pm$ 0.4 & 56.6 $\pm$ 2.6 & 77.9 $\pm$ 2.8 & 56.7 $\pm$ 2.9 \\
% nopretrainsum & 86.5 $\pm$ 0.5 & 91.3 $\pm$ 0.2 & 86.3 $\pm$ 0.2 & 61.2 $\pm$ 2.5 & 74.6 $\pm$ 1.4 & 56.7 $\pm$ 2.9 \\
L2P-GNN & 79.1 $\pm$ 0.3 & 88.1 $\pm$ 0.2 & 78.6 $\pm$ 0.3 & 59.9 $\pm$ 2.0 & 70.6 $\pm$ 2.2 & 51.6 $\pm$ 1.7 \\
MGSSL (BFS) & 86.8 $\pm$ 0.2 & 92.0 $\pm$ 0.1 & 86.4 $\pm$ 0.2 & 58.0 $\pm$ 1.4 & 78.6 $\pm$ 1.4 & 57.4 $\pm$ 2.0 \\
MGSSL (DFS) & 85.5 $\pm$ 0.2 & 91.5 $\pm$ 0.1 & 85.5 $\pm$ 0.1 & 59.5 $\pm$ 1.4 & 77.7 $\pm$ 2.5 & 60.9 $\pm$ 0.7 \\
MoMu & 85.6 $\pm$ 0.4 & 91.5 $\pm$ 0.1 & 86.4 $\pm$ 0.3 & 56.1 $\pm$ 2.1 & 79.8 $\pm$ 0.6 & 57.7 $\pm$ 2.6 \\
MolCLR & 86.6 $\pm$ 0.5 & 92.1 $\pm$ 0.3 & 86.5 $\pm$ 0.3 & 56.9 $\pm$ 2.3 & 79.9 $\pm$ 3.4 & 60.4 $\pm$ 1.4 \\
\midrule
G-Image & 87.7 $\pm$ 0.3 & \textbf{92.5} $\pm$ \textbf{0.3} & 87.4 $\pm$ 0.3 & 61.9 $\pm$ 1.8 & \textbf{82.8} $\pm$ \textbf{1.1} & \textbf{65.2} $\pm$ \textbf{3.8} \\
G-SMILES & \textbf{88.0} $\pm$ \textbf{0.3} & \underline{92.4} $\pm$ \underline{0.1} & \textbf{88.3} $\pm$ \textbf{0.5} & 61.4 $\pm$ 2.8 & 76.8 $\pm$ 2.4 & \underline{64.1} $\pm$ \underline{2.9} \\
G-\textsuperscript{1}H NMR & \underline{87.8} $\pm$ \underline{0.3} & 92.2 $\pm$ 0.2 & \underline{87.8} $\pm$ \underline{0.2} & \textbf{63.4} $\pm$ \textbf{1.3} & \underline{79.1} $\pm$ \underline{4.0} & 57.3 $\pm$ 2.3 \\
G-\textsuperscript{13}C NMR & 87.5 $\pm$ 0.4 & 92.1 $\pm$ 0.3 & 87.6 $\pm$ 0.4 & \underline{62.6} $\pm$ \underline{1.6} & 76.9 $\pm$ 2.0 & 61.9 $\pm$ 4.6 \\
G-GCMS & 86.6 $\pm$ 0.3 & 91.3 $\pm$ 0.2 & 87.1 $\pm$ 0.4 & 60.2 $\pm$ 1.6 & 76.9 $\pm$ 1.4 & 62.4 $\pm$ 1.3 \\
G-LCMS & 87.1 $\pm$ 0.2 & 91.6 $\pm$ 0.2 & 87.0 $\pm$ 0.4 & 57.4 $\pm$ 2.2 & 73.7 $\pm$ 3.1 & 63.3 $\pm$ 1.8 \\
\bottomrule
\end{tabular}}
\end{subtable}
\end{table}

\paragraph{TDC Datasets.} The results are shown in Table~\ref{table:tdc}. Similar to the analysis on MoleculeNet datasets, the proposed ACML framework achieves the best performance on all datasets, demonstrating its superiority over other self-supervised pretraining strategies. Specifically, we found that G-SMILES consistently performs well on inhibition-related property tasks, achieving top or near-top results across six tasks: PGP Inhibition, CYP2C19 Inhibition, CYP2D6 Inhibition, CYP3A4 Inhibition, CYP1A2 Inhibition, and CYP2C9 Inhibition. The model's prediction determines whether a drug inhibits specific enzymes, potentially reducing the enzyme's ability to metabolize. Therefore, the strength of using SMILES modality can be attributed to its capability to represent functional groups and structural motifs, which effectively indicate how a molecule interacts with enzymes through binding. G-Image performs the best on two substrate tasks (which involve drugs metabolized by specific enzymes), with an average AUC-ROC of 82.8\% in CYP2D6 Substrate and 65.2\% in CYP3A4 Substrate, while G-\textsuperscript{1}H NMR works the best in PAMPA Permeability (72.7\%), Bioavailability (72.3\%), and CYP2C9 Substrate (63.4\%). G-GCMS achieves the highest average ROC-AUC of 96.9\% in HIA, along with second best results in Bioavailability (68.0\%) and PGP Inhibition (91.3\%), however, it shows only trivial gains or slight declines compared to non-pretrained model on CYP2D6 Inhibition (-0.5\%), CYP1A2 Inhibition (-0.7\%), and CYP2D6 Substrate (-1.0\%). These results suggest that no single chemical modality emerges as universally optimal, rather, each modality contributes unique advantages tailored to certain tasks, which cannot be fully substituted by others.

\section{Conclusions}
\label{sec:conclusion}
In this work, we introduced the ACML framework, a novel approach tailored for molecular representational learning based on multi-modality contrastive learning. We demonstrated the inheritance of molecular graph-like structure allows graph representation to convey extensive information about molecules and treat the graph neural networks, i.e., the graph encoder, as a receptor to assimilate chemical semantics through multi-modal coordination, leading to effective and explainable molecular representational learning. Extensive experimental results on important chemical tasks, such as isomer discrimination, uncovering crucial chemical properties for drug discovery, and molecular properties prediction from MoleculeNet and TDC datasets, demonstrate the cross-modality transfer ability and the power of graph neural networks. In addition, our findings reveal a series of interesting questions to consider: of particular interest is how to effectively train a graph encoder to make it tailored to molecular learning. Instead of designing a deep and complex GNN framework and proposing various levels of self-supervised pretraining tasks, we demonstrate that the use of multi-modality contrastive learning can allow a shallow GNN (e.g. with no more than 5 layers) with light training resources to produce an expressive graph encoder with great interpretability. ACML demonstrates its potential to advance AI for drug discovery by providing deeper insights into the chemical semantics across various modalities in an efficient way.

\section{Supporting Information}
Three supporting information files are provided. S3 will be released after this work is officially published.

\paragraph{Supporting Information S1.} Details of ACML architecture, model settings, data selection and preprocessing, training and finetuning details. 

\paragraph{Supporting Information S2.} Additional experimental figures, including case studies of isomer discrimination (Figure S1-S2) and visualizations of molecular representations (Figure S3-S13).

\paragraph{Supporting Information S3.} Procedures of huamn evaluations on the isomer discrimination task. Selected isomer pairs for Graph-\textsuperscript{13}C NMR and Graph-\textsuperscript{1}H NMR discrimination task. 
\section{Competing Interests}
All authors have no competing interests to declare.

\section{Author Contribution}
Y.W., Y.L.: Conceptualization, Investigation, Methodology, Data Curation, Software, Resources, Formal Analysis, Visualization, Validation, Writing-Original Draft, Writing-Review \& Editing; L.L.: Formal Analysis, Validation, Writing-Review \& Editing; P.H.: Funding Acquisition, Resources. H.X.: Conceptualization, Investigation, Methodology, Data Curation, Software, Resources, Supervision, Project Administration, Writing-Original Draft, Writing-Review.

\section{Declaration of Generative AI and AI-assisted Technologies in The Writing Process}

During the preparation of this work, the authors used ChatGPT and Grammarly to grammar check and improve writing. After using this tool/service, the authors reviewed and edited the content as needed and took full responsibility for the content of the publication.

\section{Data and Software Availability}
\label{sec5}
The pre-training data from NP-MRD \cite{wishart2022np} and MoNA (\url{https://mona.fiehnlab.ucdavis.edu}), and the zero-shot PubChem \cite{kim2023pubchem} dataset are publicly available online, well-organized, and can be directly downloaded. RDKit \cite{RDKit} was used in parsing the dataset into individual samples. The access date for these data is July 15, 2023. We provide data preprocessing, model settings and training details in \textit{\textbf{Supporting Information S1}}.

%We provide our code as supplementary materials.
The code accompanying this work is available on Github: \url{https://github.com/yifeiwang15/ACML}. 

\section{Acknowledgement}
This work is supported by NSF OAC-1920147.

\clearpage
\appendix
\renewcommand{\thefigure}{S\arabic{figure}}
\renewcommand{\thetable}{S\arabic{table}}
\begin{center}
    \Large{\textbf{Supporting Information S1}}
\end{center}

\section{Model Architecture}
Here are the details of unimodal chemical encoders, graph encoders, and contrastive loss used in the ACML framework. 
\label{sec:acml-framework}
\subsection{Unimodal Chemcial Encoder}
For each chemical modality, we make use of effective pre-trained encoders from publicly available sources, which have been demonstrated to be effective in the corresponding downstream tasks (See Table~\ref{tab:pretrain-encoder}). 

\begin{table}[h]
\centering
\caption{Unimodal encoders of pervasive chemical modalities.}
\label{tab:pretrain-encoder}
\begin{tabular}{l|c|c|c}
\hline
Unimodal & Representation & Encoder & Pre-trained Source \\ 
\hline
Image & 2D image & CNN & Img2mol \cite{clevert2021img2mol} \\
SMILES & Sequence &Transformer  & CReSS \cite{yang2021cross}  \\
\textsuperscript{1}H NMR & Sequence & 1D CNN & N/A$^a$  \\
\textsuperscript{13}C NMR & Sequence & 1D CNN & AutoEncoder \cite{costanti2023deep}  \\
GCMS \& LCMS & Sequence & 1D CNN & AutoEncoder \cite{costanti2023deep} \\
\hline
\end{tabular}

\begin{flushleft}
\scriptsize{$^a$The \textsuperscript{1}H NMR encoder undergoes pre-training using a CNN-based network similar to what's described in \cite{costanti2023deep}. Unlike the \textsuperscript{13}C NMR and GCMS/LCMS encoders, the \textsuperscript{1}H NMR encoder focuses on reducing the embedding dimensions to compress information while preserving satisfactory reconstruction ability. The diversity in encoder choices highlights the versatility of our suggested framework. This means encoders with varying degrees of modality information can be effectively integrated into the GNN network.}
\end{flushleft}
\end{table}

\subsection{Graph Encoder}
\label{sec:gnns}
Molecules can be naturally represented as attributed relational graphs $G = (\mathcal{V}, \mathcal{E})$, where $\mathcal{V}$ and $\mathcal{E}$ are the node set and the edge set. Here a node $v \in \mathcal{V}$ and an edge $(v, u) \in \mathcal{E}$ represent the atom $v$ and the chemical bond connecting $u$ and $v$, respectively. The corresponding atom features (atomic number, chirality tag, hybridization, etc,.) and bond features (bond types, stereotypes, etc,.) are treated as node attributes and edge attributes in the graph. Graph neural networks (GNNs) \cite{wu2020comprehensive, zhou2020graph} which operate on graphs, have been combined with deep learning \cite{lecun2015deep} to learn representations on graph data \cite{baskin1997neural, sperduti1997supervised, gori2005new}. GNNs are also demonstrated as the effective frameworks for molecular representational learning \cite{kearnes2016molecular, gilmer2017neural}. 

\paragraph{Graph Convolution.} Many popular graph convolutional frameworks follow the message-passing scheme, including an iterative way of updating node representations based on the aggregations from neighboring nodes. Widely used approaches, such as GCN \cite{kipf2017semisupervised}, GIN \cite{xu2018how}, GAT \cite{velickovic2017graph} and GraphSage \cite{hamilton2017inductive}, all follow message passing schemes. For the rest of the paper, we refer ``message passing based GNNs'' as ``GNN''. For simplicity, we denote $\mathbf{x}_{u}^{(k)}$ as node feature vector of node $u$ in $k$-th layer, $\mathbf{e}_{v, u}$ as (optional) edge feature vector from node $v$ to node $u$, and let $\mathcal{N}_{u}$ represent the neighboring node set of the central node $u$. One message passing based convolution layer includes two steps: (1) create the message between the central node $u$ and its neighbor $v$, formalized by the function $\phi^{(k)}(\mathbf{x}_{u}^{(k)}, \mathbf{x}_{v}^{(k)}, \mathbf{e}_{v, u})$; (2) aggregate the messages of all neighbors of the central node $u$ using the operator $\bigoplus$, which denotes a differentiable and permutation invariant function, e.g., sum, mean or max operator. Mathematically, the latent node presentation is updated according to Equation~\ref{equ:gnn-conv} and will be fed into the next convolutional layer.

\begin{equation}
    \mathbf{x}_{u}^{(k+1)} = f^{(k)}(\mathbf{x}_{u}^{(k)}, \bigoplus_{v \in \mathcal{N}(u)}\phi^{(k)}(\mathbf{x}_{u}^{(k)}, \mathbf{x}_{v}^{(k)}, \mathbf{e}_{v, u}))
    \label{equ:gnn-conv}
\end{equation}
where $f^{(k)}$ and $\phi^{(k)}$ denote differentiable functions like MLPs (Multi Layer Perceptrons). $\mathbf{x}_{u}^{(k+1)}$ is able to captures the structural information within its $k$-hop network neighborhood after $k$ convolutional layers. 

\paragraph{Graph Pooling and Readout.} The node representations at the final iteration will be processed to generate the fixed-length graph-level representation $\mathbf{h}_{g}$, which involves graph pooling and readout operations. The pooling layer is usually applied to get a coarser graph which can be further reduced within a readout function, which is similar to conventional CNNs \cite{gu2018recent}. There is no strict distinction between pooling and readout operations in GNNs. Despite some studies \cite{cangea2018towards, ying2018hierarchical} that leveraged stacks of pooling layers for coarse graphs, most widely used works handling pooling and readout simultaneously by applying a permutation invariant function directly on all nodes in the graph to generate a fixed length graph representation \cite{vinyals2015order, zhang2018end, corso2020principal, buterez2022graph}, which is generally formalized as
\begin{equation}
    \mathbf{h}_{g} = \text{READOUT}(\mathbf{x}_{u}^{(K)}\ |\ u \in \mathcal{V})
\end{equation}

% \textbf{Graph pooling and readout}. The node representations at the final iteration will be proceeded to generate the fixed-length graph-level representation $\mathbf{h}_{g}$, which involves graph pooling and readout operations. The pooling layer is usually applied to get a coarser graph which can be further reduced within a readout function, which is similar to conventional CNNs \cite{gu2018recent}. Actually, there is no strict distinction between pooling and readout operations in GNNs. Despite some studies \cite{cangea2018towards, ying2018hierarchical} that leveraged stacks of pooling layers for coarse graphs, most widely used works handling pooling and readout simultaneously by applying a permutation invariant function directly on all nodes in the graph to generate a fixed length graph representation \cite{vinyals2015order, zhang2018end, corso2020principal, buterez2022graph}. 
% \begin{equation}
%     \mathbf{h}_{g} = \text{READOUT}(\mathbf{x}_{u}^{(K)}\ |\ u \in \mathcal{V})
% \end{equation}

\paragraph{The Choice of Aggregation Operator.} Many previous works \cite{hamilton2017inductive, xu2018how, corso2020principal, tailor2022egc} demonstrate that the choice of aggregation function contributes significantly to the expressive power and performance of the model. In this work, we used sum operator in both message aggregation and graph readout function, which enables simple and effective learning of structural graph properties implied in \cite{xu2018how}.
%%%%%%%%%%%%%%%%%%%%%%%%%%%
\subsection{Multimodal Contrastive Learning Loss}
\label{sec:contrastive-loss}
% Contrastive learning allows us to flexibly define powerful losses by contrasting positive pairs from sets of negative samples.
% This approach allows us to effectively differentiate and characterize instances by leveraging information from multiple modalities simultaneously.
Instance discrimination is a versatile technique that can be harnessed to contrast representations in various ways: within the same modality, across different modalities, or even through a combination of both modalities, referring to ``joint" instance discrimination \cite{zolfaghari2021crossclr, morgado2021audio}. The target of ACML is to learn a joint embedding between two different modalities while at the same time ensuring that similar features from the same modality stay close-by in the joint embedding. Therefore, in the implementation of the contrastive loss mechanism we adopt across-modality instance discrimination \cite{Shariatnia_Simple_CLIP_2021}. This intermodal contrastive strategy proves to be particularly beneficial in scenarios where understanding relationships and translating knowledge across modalities is a primary objective.

Mathematically, given a mini-batch of $N$ samples, $\{(g_1, c_1), (g_2, c_2),..., (g_N, c_N)\}$, where $g_i$ represents the graph modality and $c_i$ represents the chemical modality of the $i$-th molecule. According to the notations in the main text, we denote $\{(\mathbf{h}^{\text{proj}}_{g_{1}}, \mathbf{h}^{\text{proj}}_{c_{1}}), (\mathbf{h}^{\text{proj}}_{g_{2}}, \mathbf{h}^{\text{proj}}_{c_{2}}),..., (\mathbf{h}^{\text{proj}}_{g_{N}}, \mathbf{h}^{\text{proj}}_{c_{N}})\}$ as the hidden representation vectors after the projections. We denote the contrastive loss for the $i$-th graph representation as:
\begin{equation}
    \begin{aligned}
    L_{g_{i}} &=\displaystyle\sum_{j=1}^{N}  -\frac{e^{s(g_{i}, c_{j})/\tau}}{\displaystyle\sum_{k=1}^{N} e^{s(g_{i}, c_{k})/\tau}} \cdot \log \frac{e^{\delta(g_{i}, c_{j})/\tau}}{\displaystyle\sum_{k=1}^{N} e^{\delta(g_{i}, c_{k})/\tau}} \\
    &= \displaystyle\sum_{j=1}^{N} -\text{softmax}(s(g_{i}, c_{j})/\tau) \cdot \text{log}(\text{softmax}(\delta(g_{i}, c_{j})/\tau))
    \label{eq:loss_function}
    \end{aligned}
\end{equation}

where  
$s(g_{i}, c_{j}) = ((\mathbf{h}^{\text{proj}}_{g_{i}})^{T} \cdot \mathbf{h}^{\text{proj}}_{g_{j}} + (\mathbf{h}^{\text{proj}}_{c_{i}})^{T} \cdot \mathbf{h}^{\text{proj}}_{c_{j}})/2$, $\delta(g_{i}, c_{j}) = (\mathbf{h}^{\text{proj}}_{g_{i}})^{T} \cdot \mathbf{h}^{\text{proj}}_{c_{j}}$, and $\tau$ is the temperature parameter. Similarly, the contrastive loss for the $i$-th chemical modality representation is:
\begin{equation}
    L_{c_{i}} = \displaystyle\sum_{j=1}^{N} -\text{softmax}(s(g_{j}, c_{i})/\tau) \cdot \text{log}(\text{softmax}(\delta(g_{j}, c_{i})/\tau))
\end{equation}

Such loss design aims to guarantee that irrespective of the input source, whether it originates from two modalities, the optimization procedure is oriented towards attaining consistent representations.
%Three common contrastive loss mechanisms exist: end-to-end, memory bank, and momentum contrast (MoCo) \cite{he2020momentum}. Regarding gradient flow, the end-to-end mechanism channels gradients to both encoders through back-propagation. In contrast, the memory bank and MoCo route gradients to only one of the encoders. The memory bank computes and retains all learned representations as a static resource, whereas MoCo dynamically generates and encodes the necessary representations as they are needed. Given our goal of translating chemical rules between modalities, the choice to constrain one unimodal encoder aligns with the core principle of MoCo. This approach can be viewed as an expansion of the MoCo framework. 

% \begin{figure}[t]
%     \centering
%     \includegraphics[width=0.65\linewidth]{figures/multimodal_contrastive_loss.png}
%     \caption{Multimodal Contrastive Loss Mechanism \textcolor{blue}{It will be redrawn soon}}
%     \label{fig:loss}
% \end{figure}

%The modalities have already passed unimodal encoders that is able to place semantically related inputs nearby in the original feature embedding, we should preserve this proximity in the joint embedding space. Therefore, $f_{x}\left( \cdot  \right)$ and $f_{y}\left( \cdot  \right)$ should be optimized to map $x_{i}$  and $y_{i}$  of a pair $(x_{i}, y_{i})$ to a proximate location in the joint embedding (inter-modality). %The inter-modality negative sets for sample $x_{i}$ are defined as: $N_{i} = \left\{ y_{j}|\forall y_{j}\in M, j \neq i \right\}$. Therefore, the learning objective is computed as follows:

\section{Experimental Details}
\subsection{Datasets}
\subsubsection{Datasets for Training ACML framework}
For the molecular depiction image dataset, the molecules were sourced from the Natural Products Magnetic Resonance Database (NP-MRD) \cite{wishart2022np} and the images were produced by RDKit \cite{RDKit}. Similarly, the SMILES dataset, consisting of molecules and their corresponding SMILES representations, was also sourced from NP-MRD. In total, the dataset contains approximately 270,000 samples. For \textsuperscript{1}H NMR dataset and \textsuperscript{13}C NMR dataset, the molecules, \textsuperscript{1}H NMR spectra, and \textsuperscript{13}C NMR spectra were obtained from NP-MRD as well. There are about 18,000 samples in \textsuperscript{1}H NMR dataset and 18,000 samples in \textsuperscript{13}C NMR dataset. These NMR data are represented as 1D sequential data, consisting of peak locations and intensities. In detail, the raw NMR data contains the chemical shift locations and peak intensities. For \textsuperscript{1}H NMR, the chemical shift ranges from 0 - 10 ppm and for \textsuperscript{13}C NMR, the chemical shift ranges from 0 - 220 ppm. During 1D NMR processing, \textsuperscript{1}H NMR data is mapped onto a grid spanning 0 to 10 ppm with increments of 0.01 ppm, and \textsuperscript{13}C NMR data is mapped onto a grid spanning 0 to 220 ppm with increments of 0.1 ppm. The intensity corresponding to each chemical shift is recorded on the grid, with all other grid points set to 0. For the GCMS dataset and LCMS dataset, molecules, LCMS spectra, and GCMS spectra were sourced from MassBank of North America (MoNA) \cite{MoNA}. There are about 10,000 samples in the GCMS dataset and 13,000 samples in the LCMS dataset. We removed molecules with a graph size of 100 or more atoms, excluding hydrogen atoms from the graph size count. All data was accessed on July 15th, 2023, with no evidence of potential bias. To ensure robust evaluations and avoid data leakage, we explicitly removed all molecules associated with downstream evaluation tasks from the pretraining dataset. Duplicated spectra and molecules were also removed.

\subsubsection{Zero-Shot Datasets}
For the molecular candidate pool, 1M molecules were randomly chosen from PubChem \cite{kim2023pubchem} with access on July 15th, 2023, with no intersection with the training dataset.  For the spectra, 1000 spectra were randomly chosen from the validation dataset. The data was accessed on July 15th, 2023, with no potential bias identified. To prevent data leakage, duplicated spectra and molecules were removed, and all data cleaning steps were fully documented.
% \subsubsection{Molecular property prediction datasets}

\subsection{Training}
\subsubsection{Pretraining in ACML}
For graph modality, each atom is embedded by its atomic number and chirality type, and each bond is embedded by its bond type and stereotype. We used GNNs with ReLU activation function and a sum pooling is applied on each graph as the final readout operation to extract the hidden graph representation. The best model settings are provided in Table~\ref{tab:gnn-training-setting}. Besides GIN \cite{xu2018how}, we tried multiple convolution techniques and found that GCN \cite{kipf2017semisupervised} achieved similar results to GIN, whereas GAT \cite{velickovic2017graph} and GraphSage \cite{hamilton2017inductive} degraded performance during validation. We noticed that the number of layers did not have a strong effect as long as it exceeded a value of 5, implying that a shallow graph encoder is expressive enough. For chemical modalities, we used pre-trained unimodal encoders according to Table \ref{tab:pretrain-encoder}, where each encoder keeps frozen in the training phase.  Outputs from the graph encoder and the chemical unimodal encoder are projected into the same dimensional joint hidden space using a 2-layer or 3-layer MLP with a specific projection dimension as provided in Table~\ref{tab:gnn-training-setting}. All models were optimized with AdamW \cite{loshchilov2017decoupled} in Pytorch implementation with $\beta_1=0.9, \beta_2=0.999$, and weight decay of 0.001. The model is trained with a batch size of 128 for a total of 100 epochs. Each model is trained on one NVIDIA Tesla V100 GPU using float32 precision. The entire framework is training efficiently since only one shallow graph encoder and two MLPs from the projection module need to be updated through training. 

\begin{table}[h]
\centering
\caption{Best model settings in ACML.}
\label{tab:gnn-training-setting}
\begin{tabular}{l|c|c|c|c}
\hline
Model & GNN type & \#\ Layers & GNN hidden dim & Projection dim \\ 
\hline
G-Image & GIN & 5 & 64 & 512 \\
G-SMILES & GIN & 5 & 64 & 512 \\
G-\textsuperscript{1}H NMR & GIN & 5 & 512 & 256 \\
G-\textsuperscript{13}C NMR & GIN & 3 & 512 & 256 \\
G-GCMS & GIN & 5 & 128 & [128, 128]$^*$ \\
G-LCMS & GIN & 5 & 128 & [128, 128]$^*$ \\
\hline
\end{tabular}
% \begin{flushleft}
\\ \scriptsize{$^*$ G-GCMS and G-LCMS use the 3-layer MLP in the projection module while others use the 2-layer MLP in projection.}
% \end{flushleft}
\end{table}

\subsubsection{Finetuning on Molecular Property Prediction Tasks}
\label{sec:finetuning}
The following configurations were applied to all finetuning tasks from MoleculeNet \cite{wu2018moleculenet} and TDC \cite{huang2022artificial} accessed on July 15, 2023. For each task, we utilized a 5-layer GIN \cite{xu2018how} as the backbone graph encoder, exploring hidden dimensions of [64, 128, 300]. The GIN encoder was initialized with weights pre-trained using the ACML framework. We used a batch size of 32 and a maximal training epoch of 100. We use scaffold splitting \cite{ramsundar2019deep} to divide the dataset into the training set, validation set, and test set with a ratio of 8:1:1. Scaffold splitting helps avoid data leakage in molecular datasets because it ensures that structurally similar molecules do not appear in both the training and test sets. All experiments were conducted on one Tesla V100 GPU. We used the Adam optimizer. Different learning rates from [1e-2, 1e-3, 1e-4, 1e-5] are explored and the best results are reported. Before training, we performed data cleaning to remove certain molecules that failed to pass the sanitizing process in the RDKit or contained abnormal valence of a certain atom, as suggested in \cite{ijcai2021, lim2021predicting, wang2023motif}. The detailed dataset statistics are summarized in Table~\ref{tab:2d-data}. 

\begin{table}[t!]
\caption{Summary of molecular property prediction datasets.}
\label{tab:2d-data}
\centering
\footnotesize
\begin{tabular}{llccc}
\hline
\textbf{Dataset} & \qquad \qquad \qquad \textbf{Description} & \textbf{\# Graphs} & \textbf{\# Valid graphs} & \textbf{\# Tasks} \\
\hline
\multirow{3}{*}{BACE} & \multirow{3}{0.4\textwidth}{\footnotesize{Quantitative (IC50) and qualitative (binary label) binding results for a set of inhibitors of human $\beta$-secretase 1 (BACE-1).}} & \multirow{3}{*}{1513} & \multirow{3}{*}{1513} & \multirow{3}{*}{1} \\
 & & & & \\
 & & & & \\
 \hline
\multirow{3}{*}{BBBP} & \multirow{3}{0.4\textwidth}{\footnotesize{Binary labels of blood-brain barrier penetration (permeability).}} & \multirow{3}{*}{2039} & \multirow{3}{*}{1953} & \multirow{3}{*}{1} \\
 & & & & \\
 & & & & \\
 \hline
\multirow{3}{*}{Clintox} & \multirow{3}{0.4\textwidth}{\footnotesize{Qualitative data of drugs approved by the FDA and those that have failed clinical trials for toxicity reasons.}} & \multirow{3}{*}{1478} & \multirow{3}{*}{1469} & \multirow{3}{*}{2} \\
 & & & & \\
 & & & & \\
 \hline
\multirow{3}{*}{Sider} & \multirow{3}{0.4\textwidth}{\footnotesize{Database of marketed drugs and adverse drug reactions (ADR), grouped into 27 system organ classes.}} & \multirow{3}{*}{1427} & \multirow{3}{*}{1295} & \multirow{3}{*}{27} \\
 & & & & \\
 & & & & \\
 \hline
\multirow{3}{*}{Tox21} & \multirow{3}{0.4\textwidth}{\footnotesize{Qualitative toxicity measurements on 12 biological targets, including nuclear receptors and stress response pathways.}} & \multirow{3}{*}{7831} & \multirow{3}{*}{7774} & \multirow{3}{*}{12} \\
 & & & & \\
 & & & & \\
 \hline
\multirow{3}{*}{HIV} & \multirow{3}{0.4\textwidth}{\footnotesize{Experimentally measured abilities to inhibit HIV replication.}} & \multirow{3}{*}{41127} & \multirow{3}{*}{41125} & \multirow{3}{*}{1} \\
 & & & & \\
 & & & & \\
 \hline
\multirow{3}{*}{Esol} & \multirow{3}{0.4\textwidth}{\footnotesize{Water solubility data (log solubility in mols per litre) for common organic small molecules.}} & \multirow{3}{*}{1128} & \multirow{3}{*}{1127} & \multirow{3}{*}{1} \\
 & & & & \\
 & & & & \\
 \hline
\multirow{3}{*}{FreeSolv} & \multirow{3}{0.4\textwidth}{\footnotesize{Experimental and calculated hydration free energy of small molecules in water.}} & \multirow{3}{*}{642} & \multirow{3}{*}{639} & \multirow{3}{*}{1} \\
 & & & & \\
 & & & & \\
\hline
\end{tabular}
\end{table}

\subsubsection{Discussion of Training Efficiency}
To validate our claim regarding training efficiency, we focused on the two phases: pretraining and fine-tuning. 

\paragraph{Pretraining.} We found the challenge of fairly comparing efficiency solely by time, given the different types of resource utilization involved, such as human efforts, dataset size, and CPU/GPU resources. Overall, the proposed approach uses a relatively small number of learnable parameters (only one graph encoder and two MLPs for projection), which is lightweight with minimal preprocessing and limited human involvement. For instance, SMILES-to-graph conversion takes just five minutes per million molecules on a standard laptop. In contrast, some baseline methods demand significant CPU or human resources for preprocessing (e.g., constructing motif vocabulary) or for multimodal pretraining, which requires generating augmented graphs and optimizing augmentation settings. Besides preprocessing efficiency, we provide a clear resource-efficiency comparison of pretraining. For MoMu and MotifConv, we report resource utilization as stated in their original papers. For L2P-GNN, MGSSL, MolCLR and the proposed ACML instantiations, we reproduced pretraining on a single NVIDIA Tesla V100 32GB GPU with 100 epochs. We reported both the dataset sizes and the training time per epoch in Table~\ref{tab:pretraining-cost-comparison}. These benchmarks indicate that our approach requires fewer resources, which is advantageous in scenarios where computational and financial costs are critical factors.

\begin{table}[t!]
    \centering
    \caption{Resource Utilization and Time per Epoch Comparison}
    \label{tab:pretraining-cost-comparison}
    \begin{tabular}{llll}
        \toprule
        \textbf{Model} & \textbf{GPU Requirement} & \textbf{Dataset Size} & \textbf{Time per Epoch} \\
        \midrule
        MoMu           & 8 V100, 32G & $\sim$ 16K              & –         \\
        MotifConv      & 8 RTX 2080, 11G   & –                & Several hours per dataset \\
        L2P-GNN        & 1 V100, 32G  & $\sim$ 250K               &  $\sim$ 7 min          \\
        MGSSL          & 1 V100, 32G  & $\sim$ 250K     &  $\sim$ 40 min         \\
        MolCLR         & 1 V100, 32G  & $\sim$ 1 Million &  $\sim$ 80 min         \\
        \midrule
        G-Image        & 1 V100, 32G  & $\sim$ 250K    &  $\sim$ 3 min          \\
        G-SMILES       & 1 V100, 32G  & $\sim$ 250K     &  $\sim$ 7 min          \\
        G-\textsuperscript{13}C NMR         & 1 V100, 32G  & $\sim$ 18K   &   $\sim$ 12 sec         \\
        G-\textsuperscript{1}H NMR         & 1 V100, 32G  & $\sim$ 18K &  $\sim$ 12 sec         \\
        G-GCMS         & 1 V100, 32G  & $\sim$ 13K   &  $\sim$ 8 sec          \\
        G-LCMS         & 1 V100, 32G  & $\sim$ 10K    &  $\sim$ 7 sec          \\
        \bottomrule
    \end{tabular}
\end{table}

\paragraph{Finetuning.} This stage is standardized across models, so the efficiency gains result solely from the graph architecture. We observed that in some datasets, a pretrained graph encoder with a smaller hidden dimension is already enough to achieve a good performance, as implied in Table~\ref{tab:embedding_dimensions}. We believe this is because the proposed approach can learn good chemical knowledge through ACML pretraining.

\begin{table}[h!]
\centering
\caption{Embedding Dimensions of the Best Models for Different Datasets.}
\label{tab:embedding_dimensions}
\scalebox{0.95}{
\begin{tabular}{cl}
\toprule
\textbf{Embedding Dimension} & \textbf{Datasets} \\ \midrule
\begin{tabular}[t]{@{}l@{}} \\ 64 \\  \end{tabular}& \begin{tabular}[t]{@{}l@{}} BACE, BBBP, SIDER, ESOL, FreeSolv, HIA, PGP Inhibition, \\  CYP2C19 Inhibition, CYP2D6 Inhibition, CYP3A4 Inhibition \\ CYP1A2 Inhibition, CYP2C9 Inhibition, CYP2C9 Substrate \end{tabular}   \\ \midrule
128 & \begin{tabular}[t]{@{}l@{}} ClinTox, Tox21, HIV, CYP2D6 Substrate, CYP3A4 Substrate  \end{tabular} \\ \midrule
300 & PAMPA Permeability, Bioavailability \\ 
\bottomrule
\end{tabular}}
\end{table}
\clearpage
\begin{center}
    \Large{\textbf{Supporting Information S2}}
\end{center}

\section{Additional Case Studies of Isomer Discrimination Task}
\begin{figure}[h]
    \centering
    \includegraphics[width=1.0\linewidth]{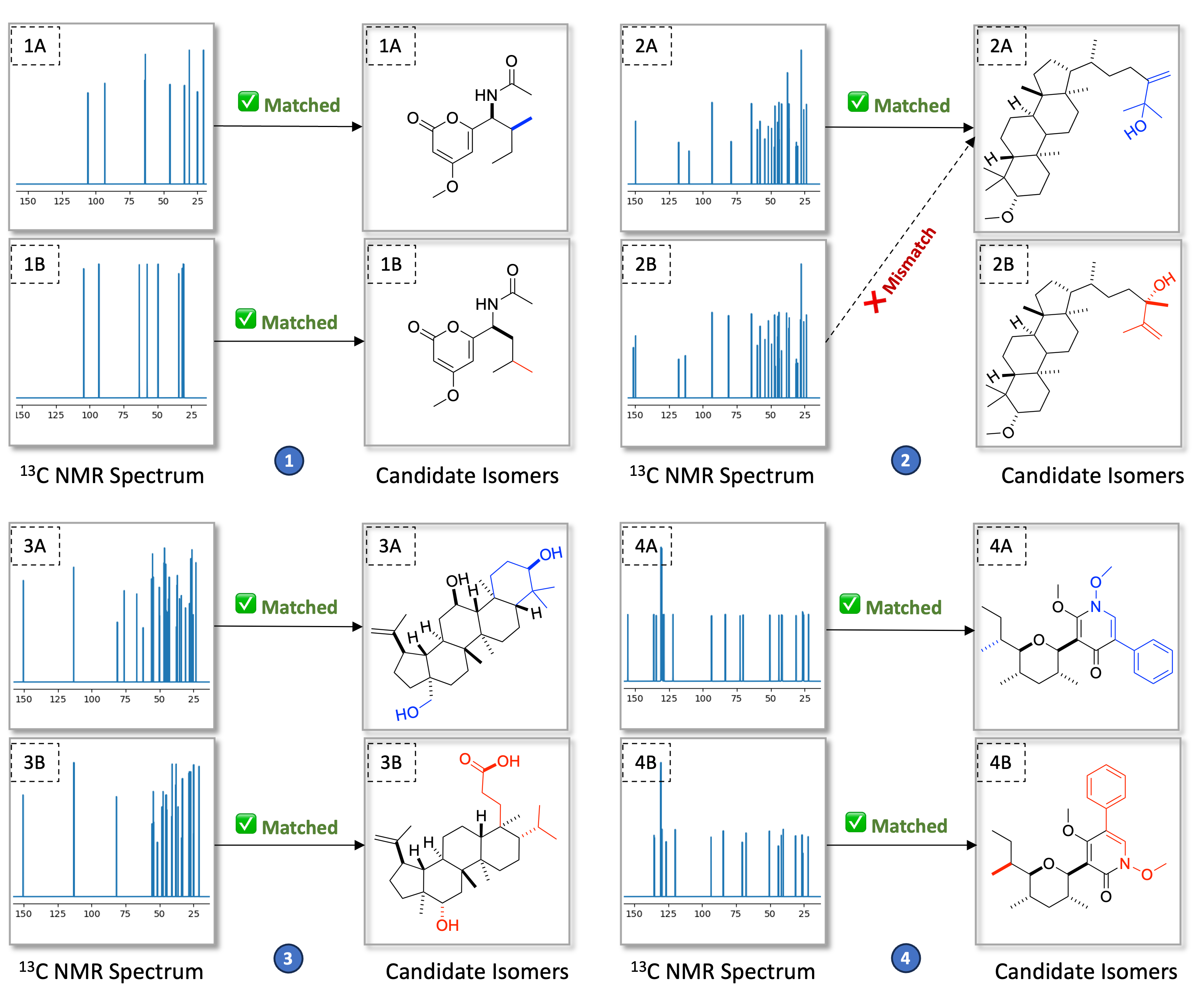}
    \caption{\textsuperscript{13}C NMR Isomer discrimination task demonstrations. The correct pairs of \textsuperscript{13}C NMR and molecule are presented horizontally. If model correctly identifies the correct molecule from spectrum, they are linked by solid arrow and labeled as ``Matched''. Out of the 4 challenging isomer pairs, our proposed model correctly identifies 3 isomers.}
    \label{fig:cnmr_case}
\end{figure}

\begin{figure}[h]
    \centering
    \includegraphics[width=1.0\linewidth]{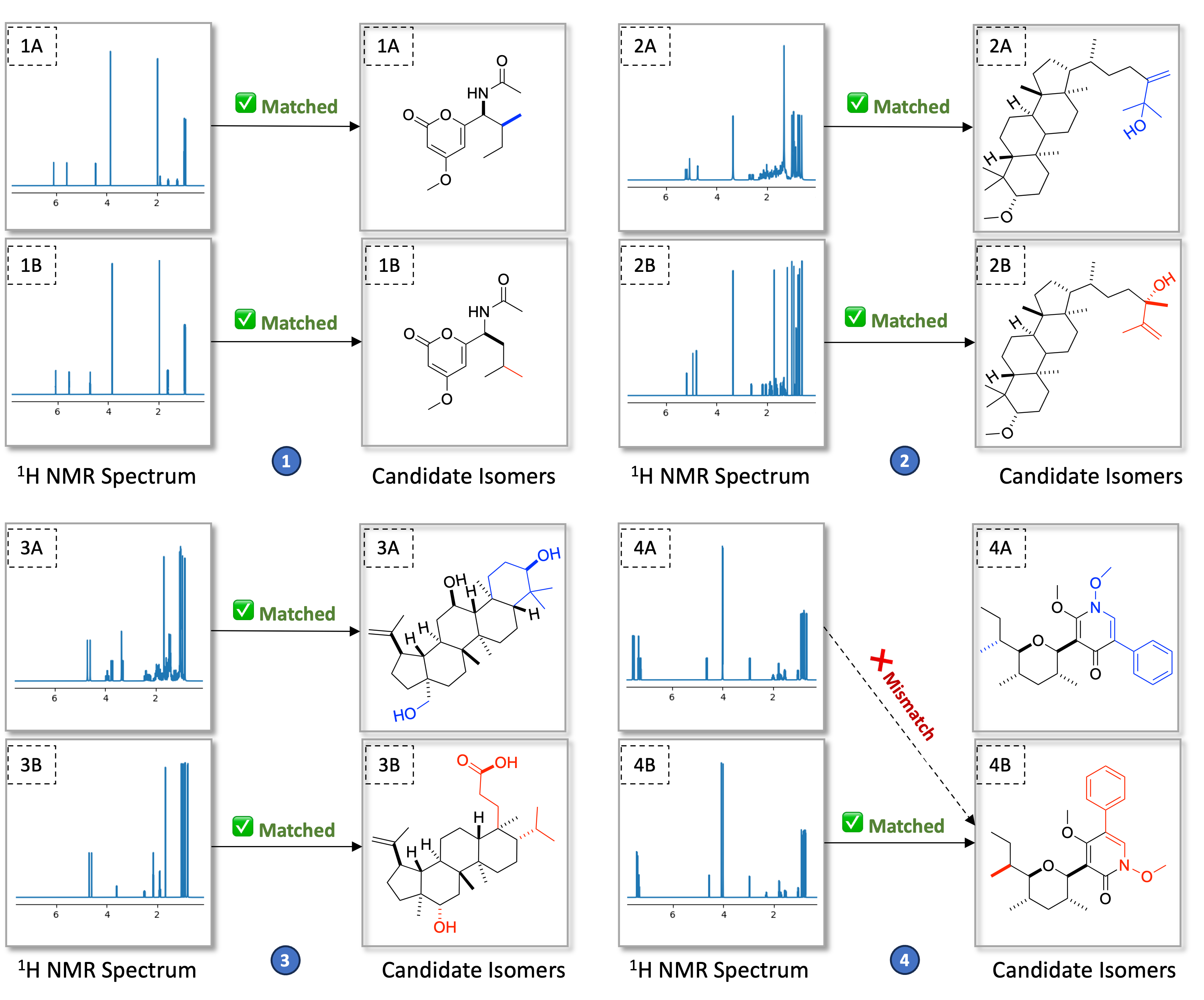}
    \caption{\textsuperscript{1}H NMR Isomer discrimination demonstrations. The correct pairs of \textsuperscript{1}H NMR and molecule are presented horizontally. If model correctly identifies the correct molecule from spectrum, they are linked by solid arrow and labeled as ``Matched''. Out of the 4 challenging isomer pairs, our proposed model correctly identifies 3 isomers.}
    \label{fig:hnmr_case}
\end{figure}

\clearpage

\section{Additional Visualizations of Molecular Representations}
We provide all graph embedding visualization results from different pretrained graph encoders.
\begin{figure}[h]
    \centering\includegraphics[width=1\linewidth]{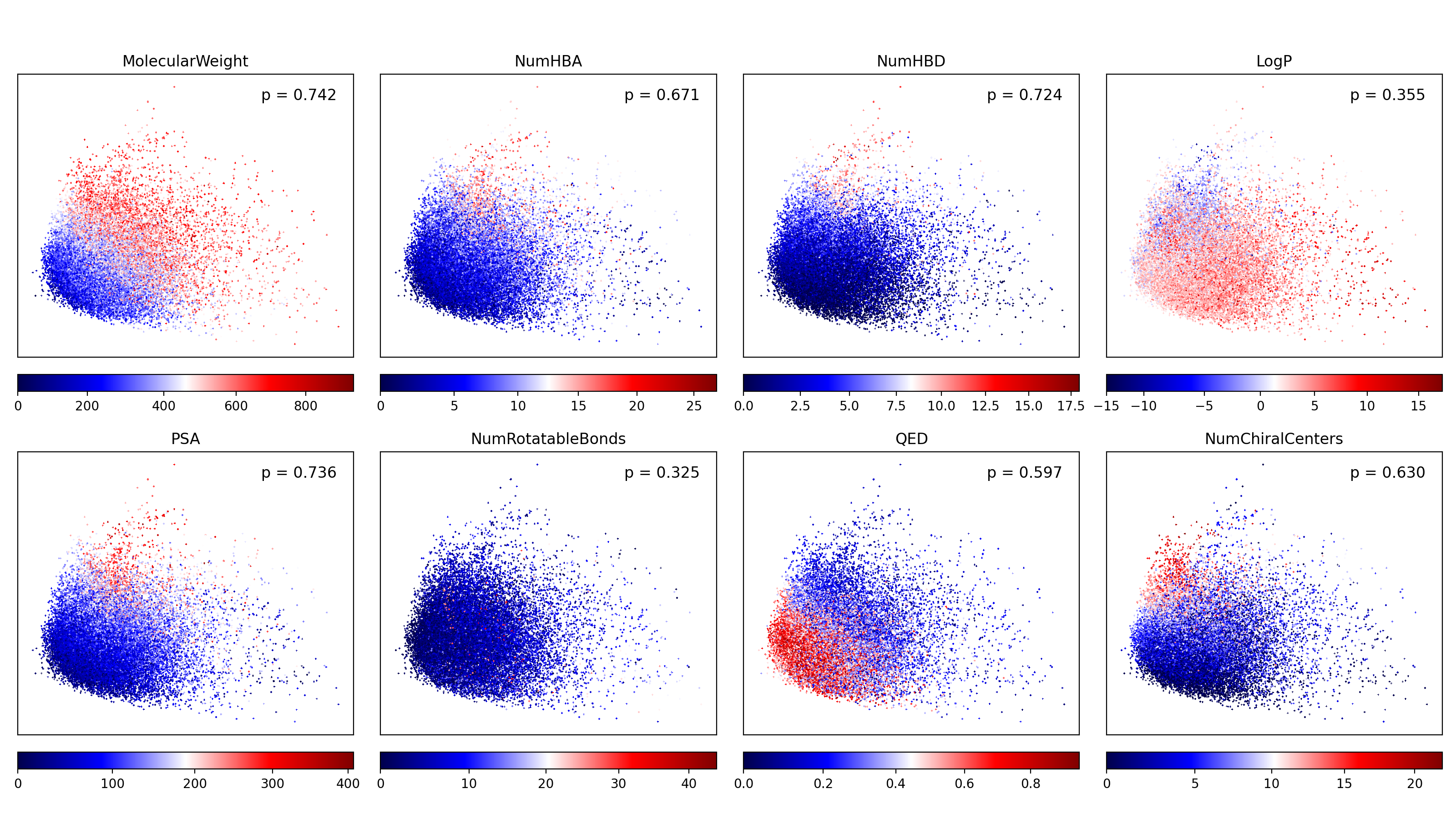}
    \caption{G-Image}
\end{figure}

\begin{figure}[h]
    \centering\includegraphics[width=1\linewidth]{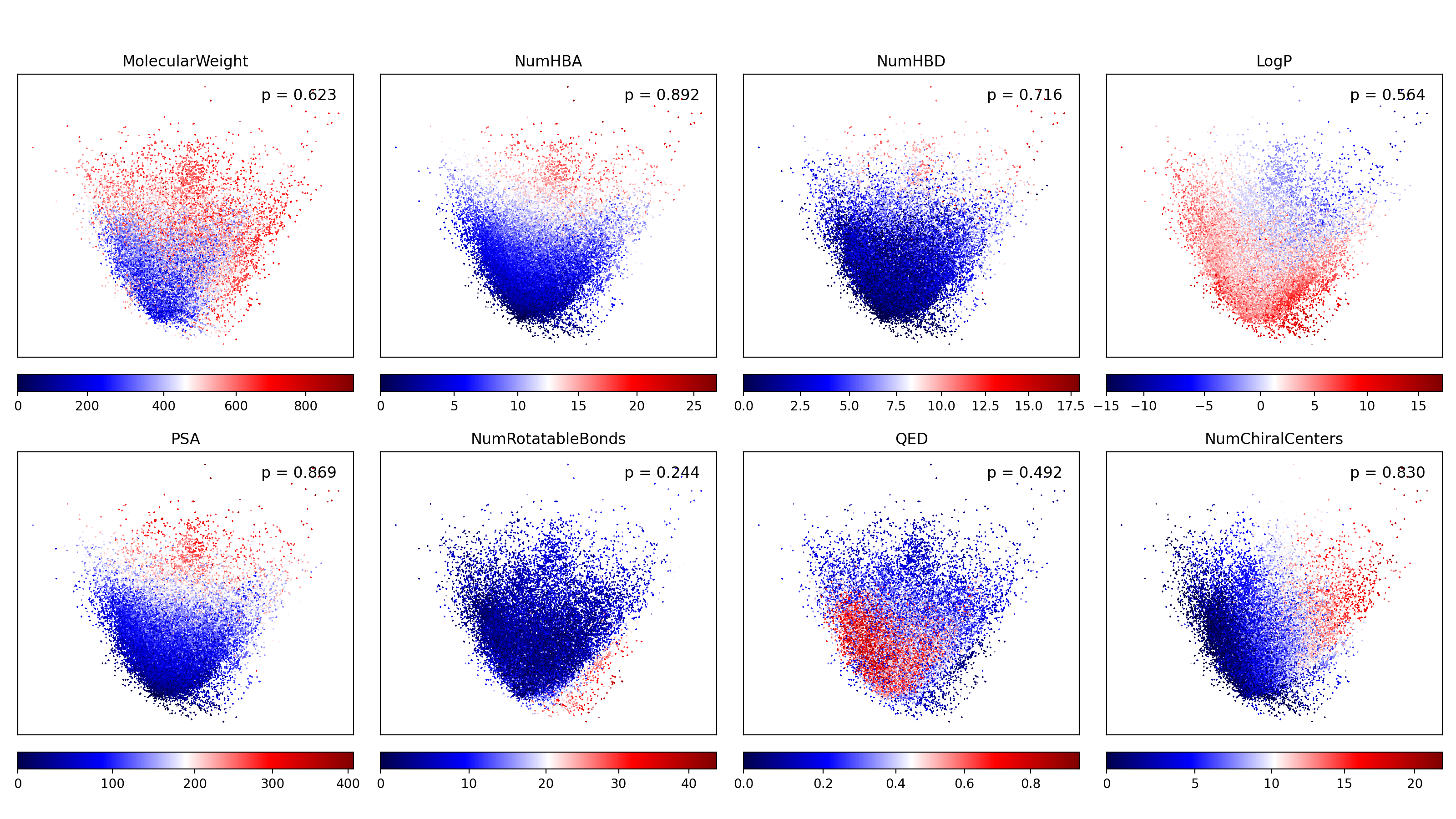}
    \caption{G-SMILES}
\end{figure}

\begin{figure}[h]
    \centering\includegraphics[width=1\linewidth]{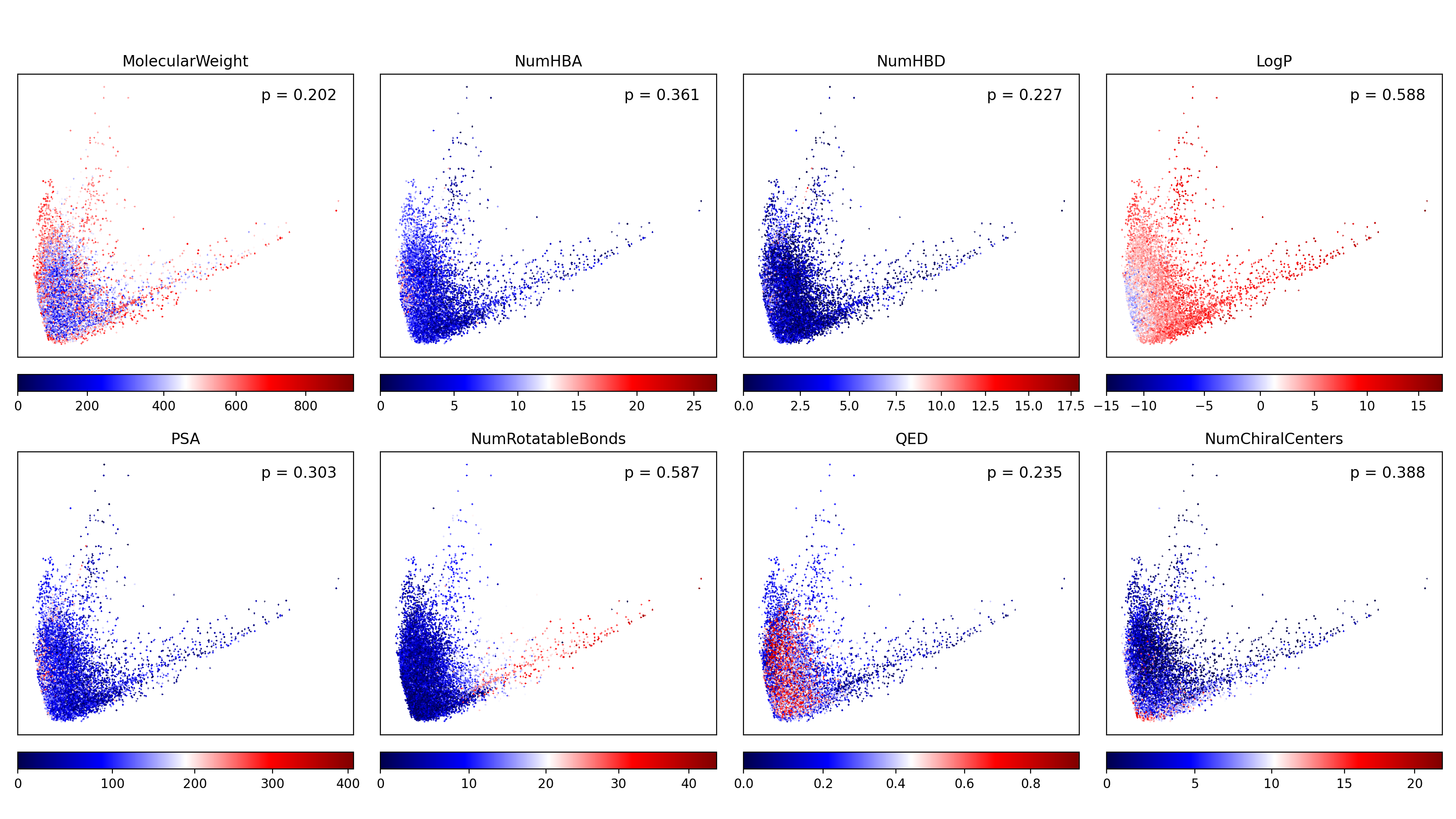}
    \caption{G-\textsuperscript{1}H NMR}
\end{figure}

\begin{figure}[h]
    \centering\includegraphics[width=1\linewidth]{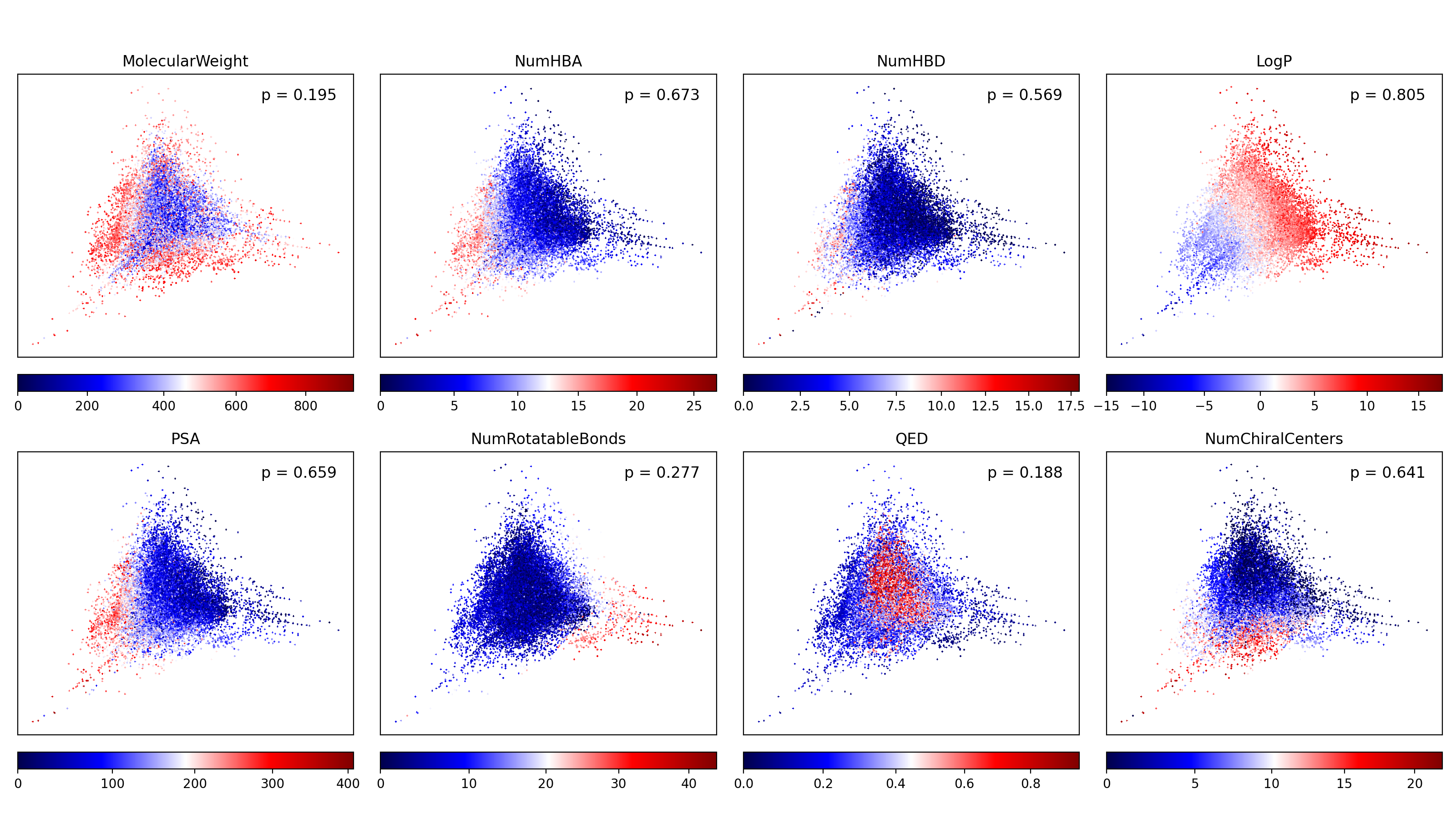}
    \caption{G-\textsuperscript{13}C NMR}
\end{figure}

\begin{figure}[h]
    \centering\includegraphics[width=1\linewidth]{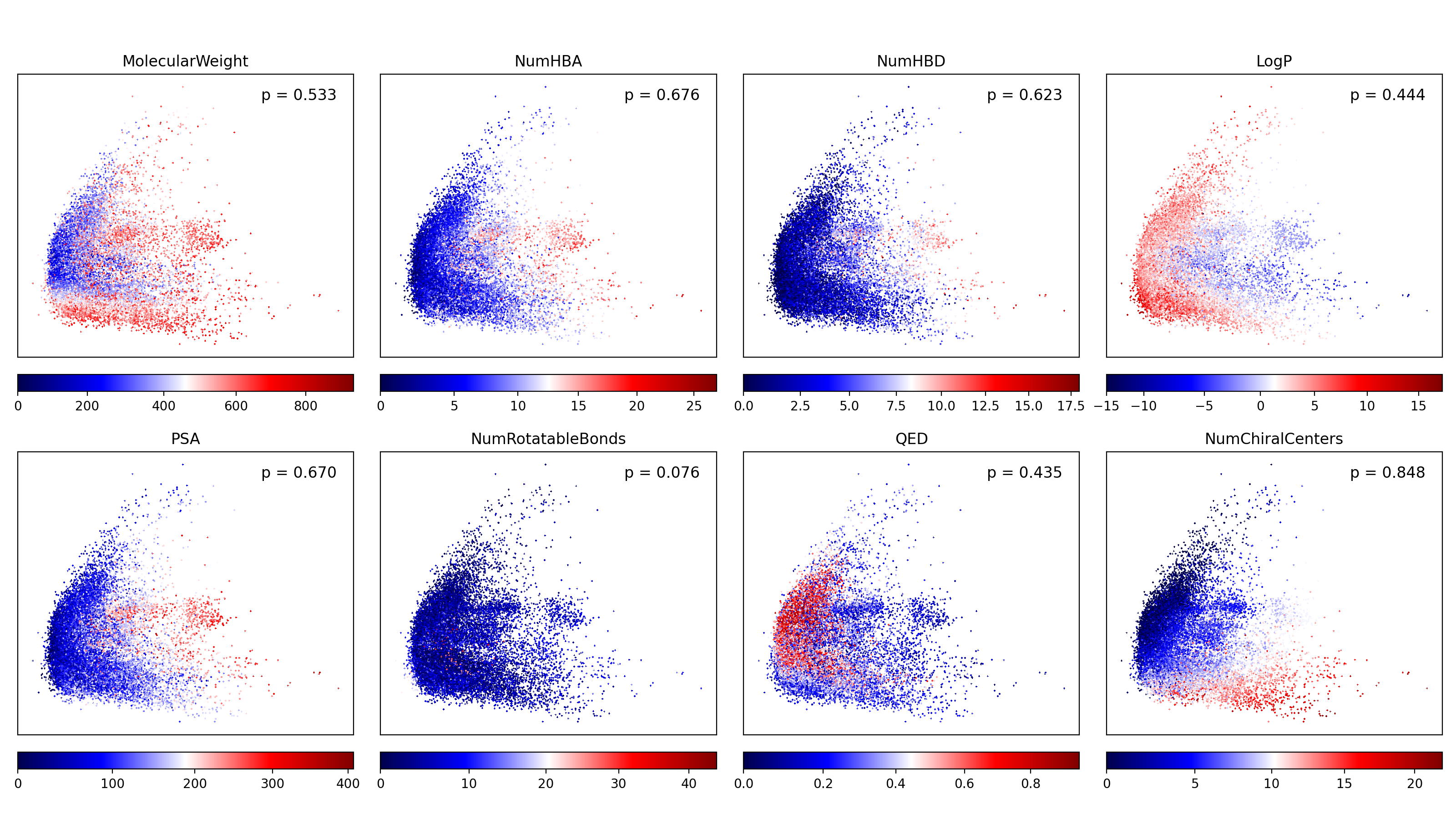}
    \caption{G-GCMS}
\end{figure}

\begin{figure}[h]
    \centering\includegraphics[width=1\linewidth]{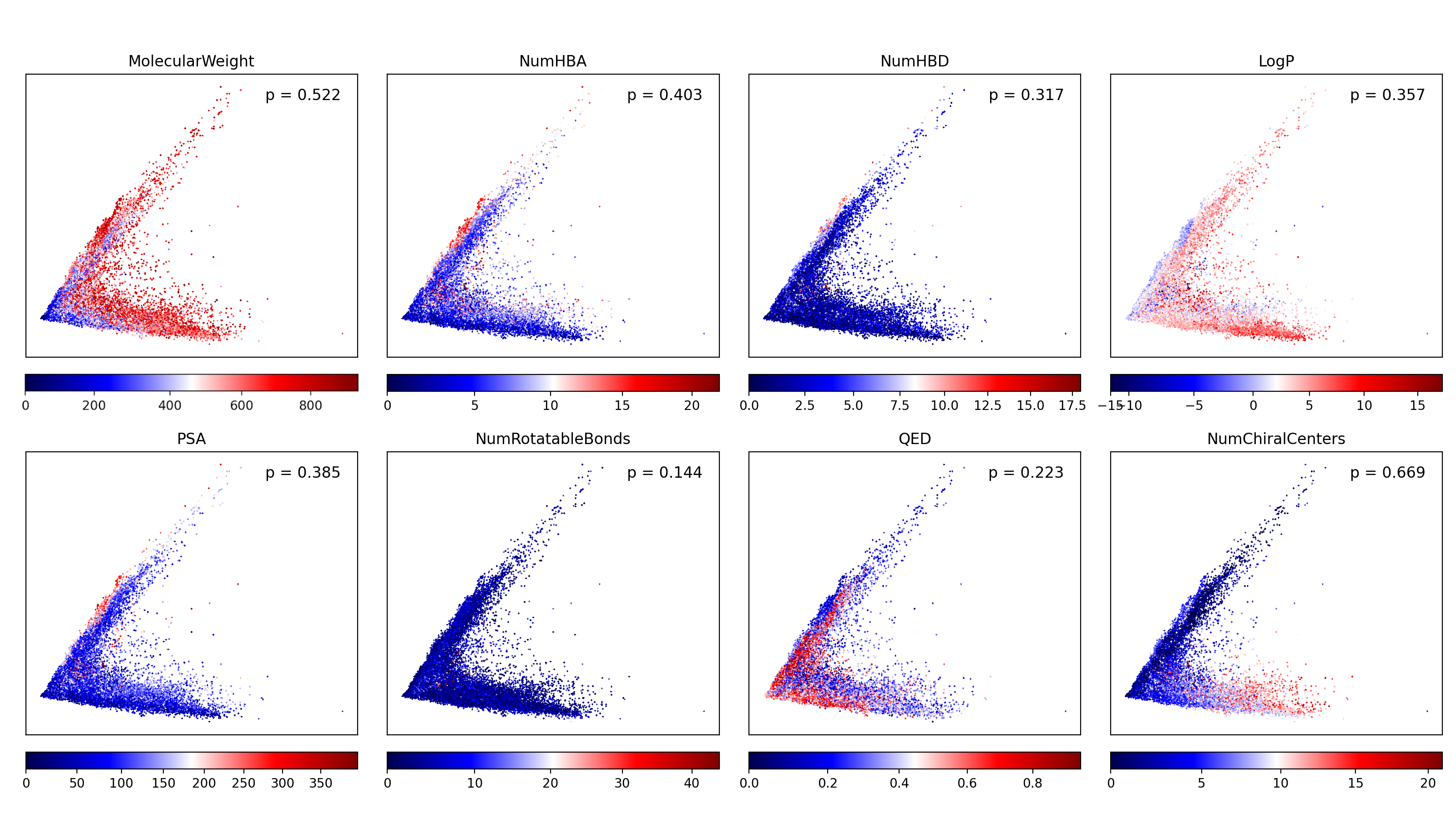}
    \caption{G-LCMS}
\end{figure}

\begin{figure}[h]
    \centering\includegraphics[width=1\linewidth]{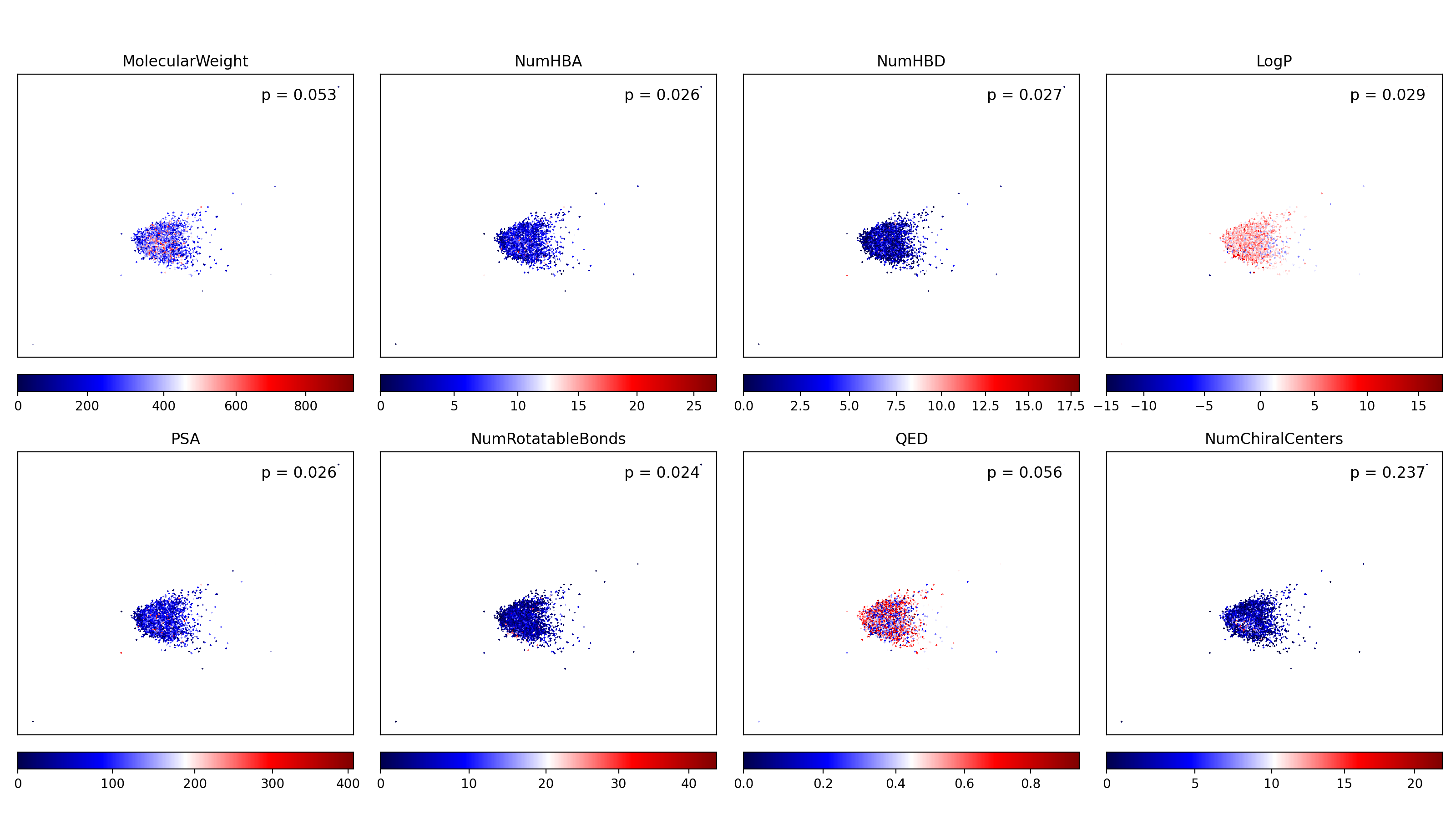}
    \caption{L2P-GNN}
\end{figure}

\begin{figure}[h]
    \centering\includegraphics[width=1\linewidth]{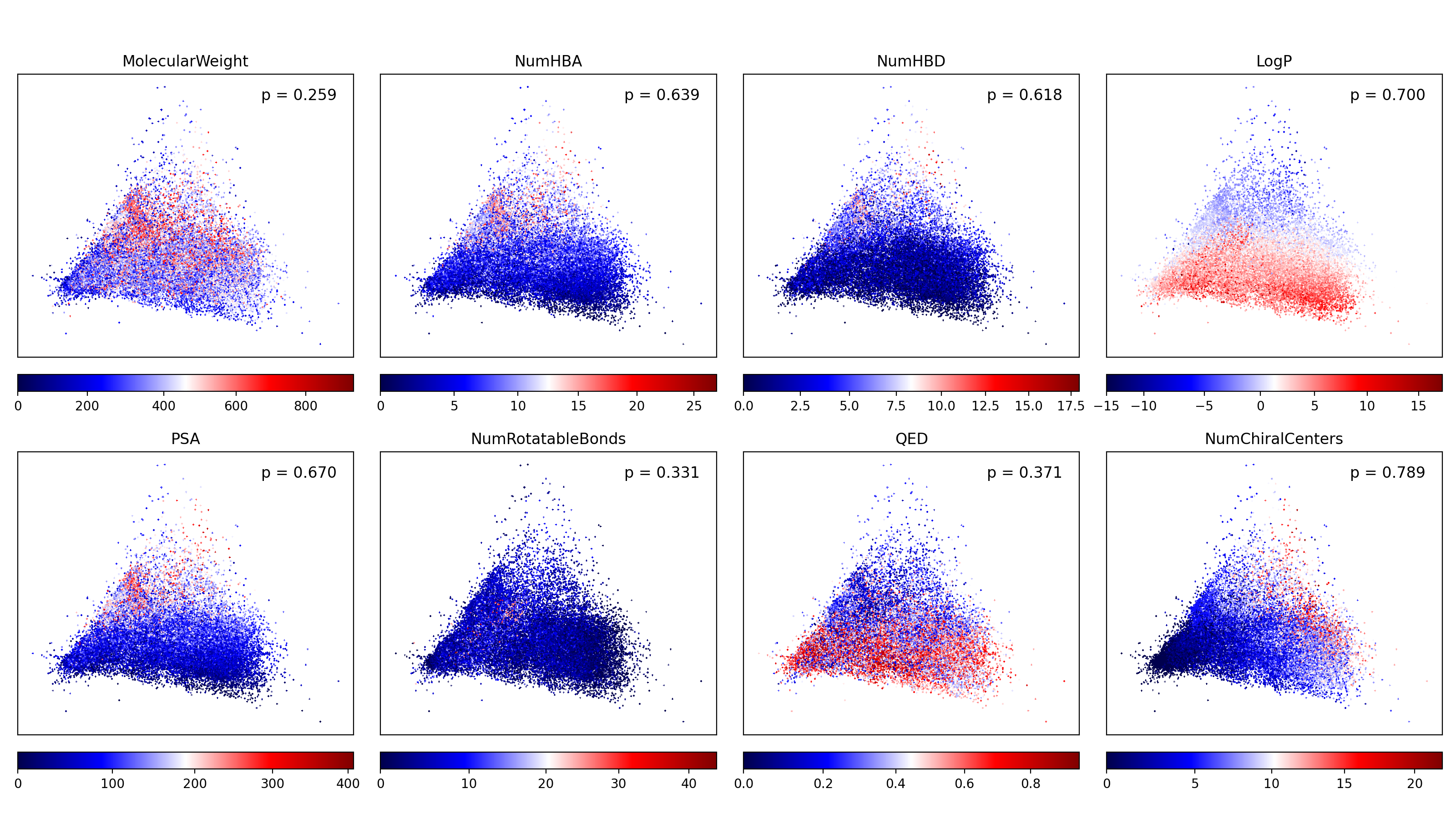}
    \caption{MGSSL (bfs)}
\end{figure}

\begin{figure}[h]
    \centering\includegraphics[width=1\linewidth]{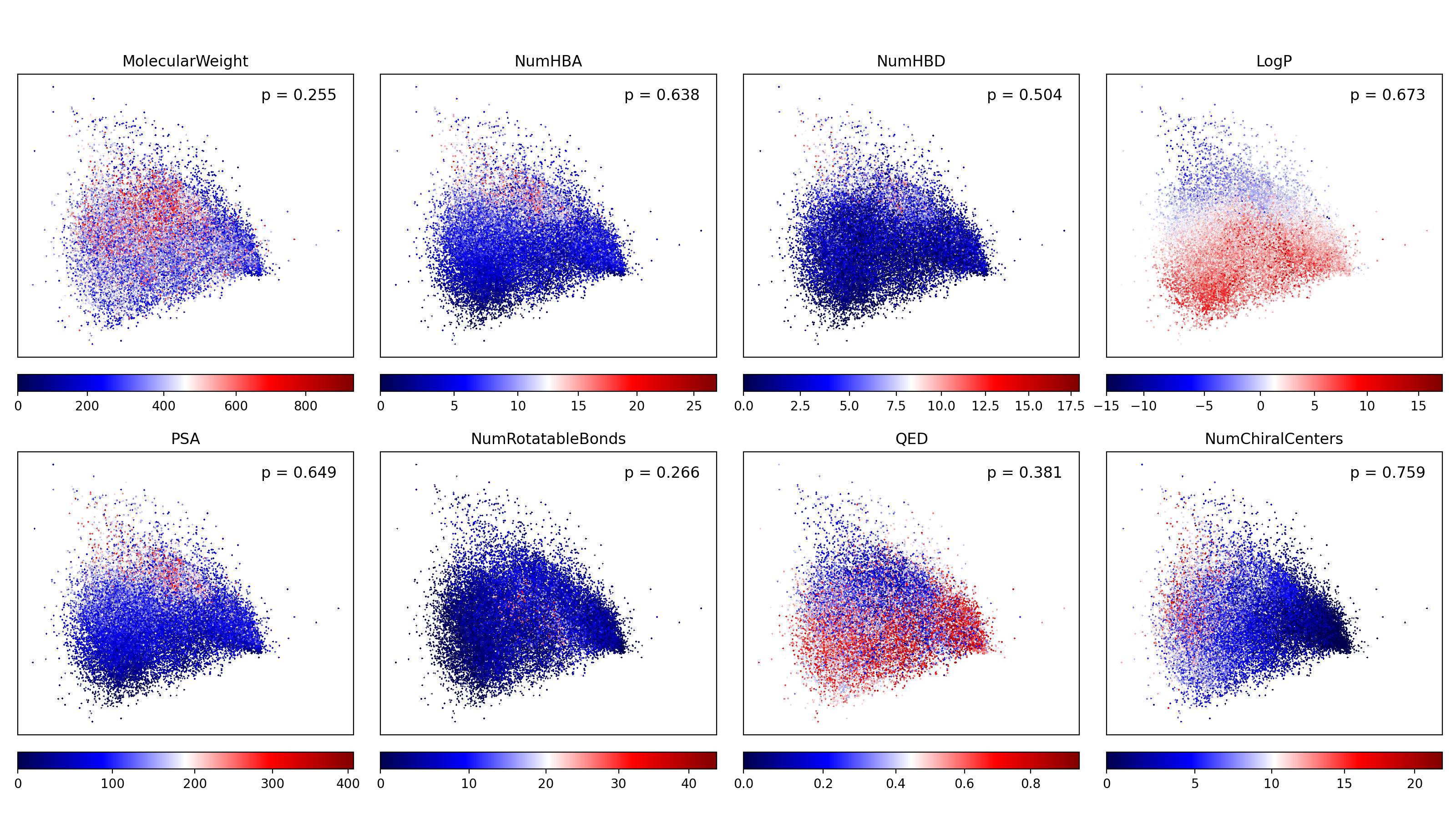}
    \caption{MGSSL (dfs)}
\end{figure}

\begin{figure}[h]
    \centering\includegraphics[width=1\linewidth]{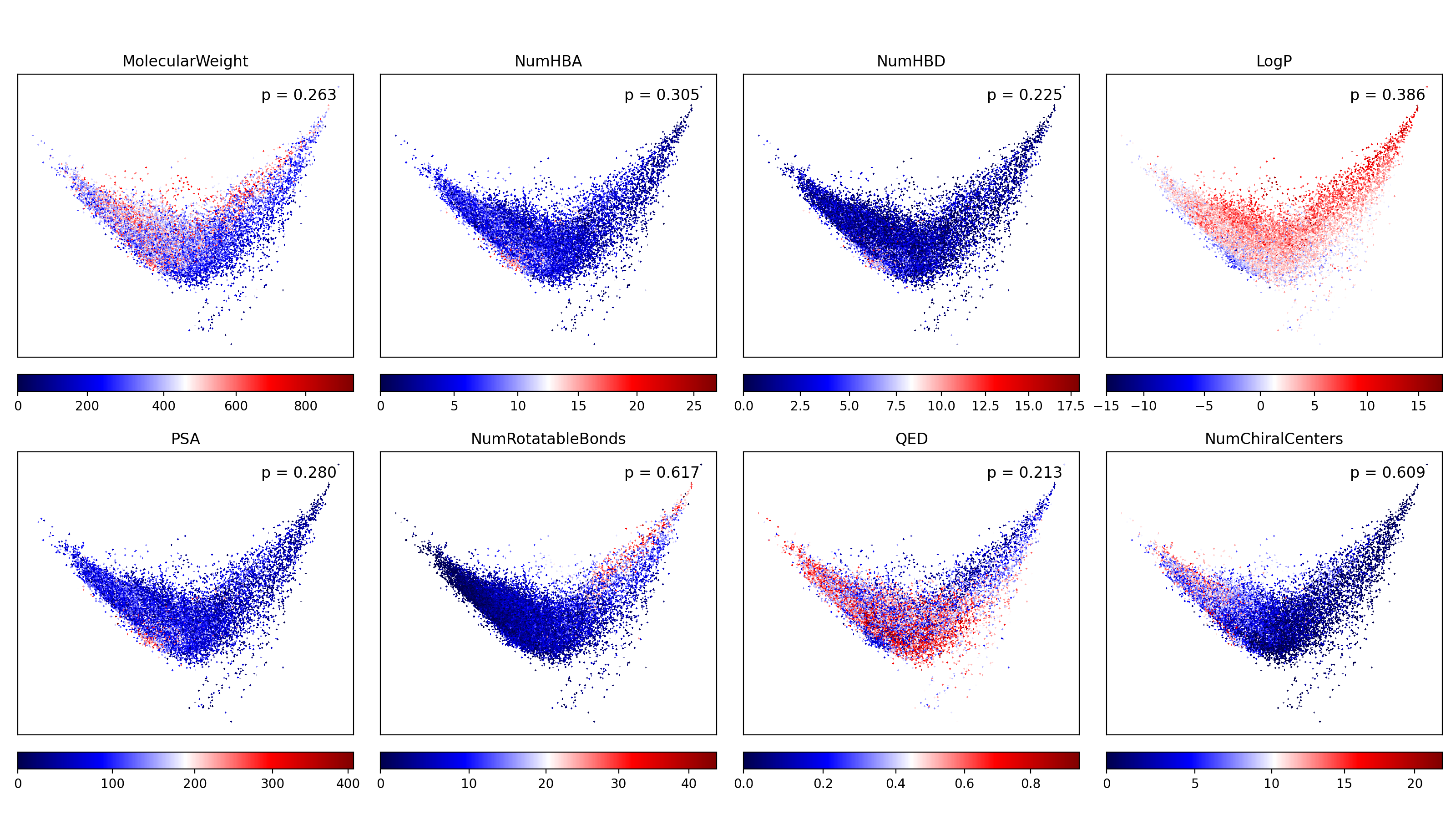}
    \caption{MoMu}
\end{figure}

\begin{figure}[h]
    \centering\includegraphics[width=1\linewidth]{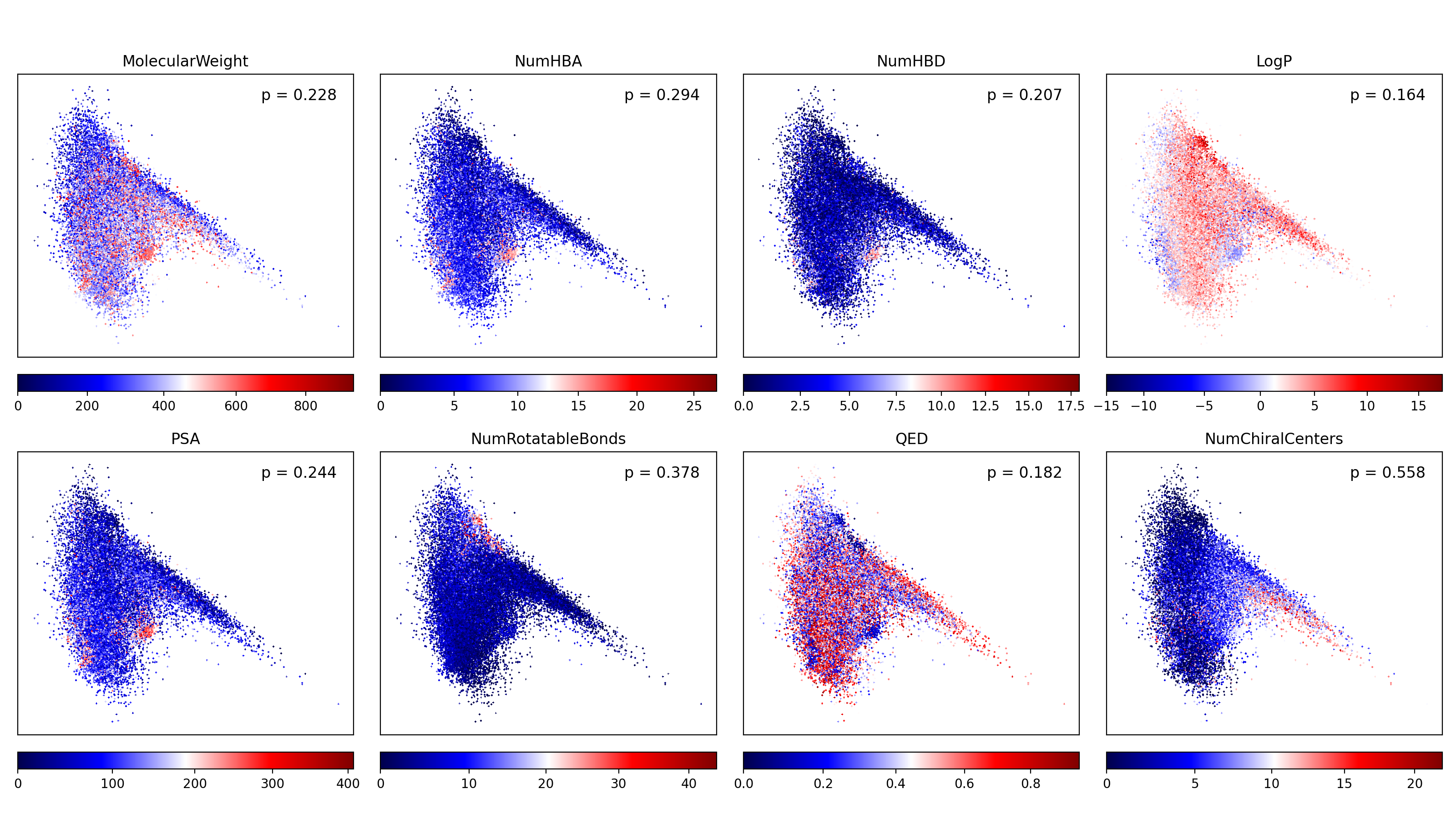}
    \caption{MolCLR}
\end{figure}

\clearpage
{\small
\bibliography{sn.bib}}

\providecommand{\latin}[1]{#1}
\makeatletter
\providecommand{\doi}
  {\begingroup\let\do\@makeother\dospecials
  \catcode`\{=1 \catcode`\}=2 \doi@aux}
\providecommand{\doi@aux}[1]{\endgroup\texttt{#1}}
\makeatother
\providecommand*\mcitethebibliography{\thebibliography}
\csname @ifundefined\endcsname{endmcitethebibliography}  {\let\endmcitethebibliography\endthebibliography}{}
\begin{mcitethebibliography}{93}
\providecommand*\natexlab[1]{#1}
\providecommand*\mciteSetBstSublistMode[1]{}
\providecommand*\mciteSetBstMaxWidthForm[2]{}
\providecommand*\mciteBstWouldAddEndPuncttrue
  {\def\EndOfBibitem{\unskip.}}
\providecommand*\mciteBstWouldAddEndPunctfalse
  {\let\EndOfBibitem\relax}
\providecommand*\mciteSetBstMidEndSepPunct[3]{}
\providecommand*\mciteSetBstSublistLabelBeginEnd[3]{}
\providecommand*\EndOfBibitem{}
\mciteSetBstSublistMode{f}
\mciteSetBstMaxWidthForm{subitem}{(\alph{mcitesubitemcount})}
\mciteSetBstSublistLabelBeginEnd
  {\mcitemaxwidthsubitemform\space}
  {\relax}
  {\relax}

\bibitem[Liang \latin{et~al.}(2023)Liang, Zadeh, and Morency]{bib1}
Liang,~P.~P.; Zadeh,~A.; Morency,~L.-P. Foundations and Trends in Multimodal Machine Learning: Principles, Challenges, and Open Questions. \textbf{2023}, \relax
\mciteBstWouldAddEndPunctfalse
\mciteSetBstMidEndSepPunct{\mcitedefaultmidpunct}
{}{\mcitedefaultseppunct}\relax
\EndOfBibitem
\bibitem[Ramesh \latin{et~al.}(2022)Ramesh, Dhariwal, Nichol, Chu, and Chen]{bib4}
Ramesh,~A.; Dhariwal,~P.; Nichol,~A.; Chu,~C.; Chen,~M. Hierarchical Text-Conditional Image Generation with CLIP Latents. 2022\relax
\mciteBstWouldAddEndPuncttrue
\mciteSetBstMidEndSepPunct{\mcitedefaultmidpunct}
{\mcitedefaultendpunct}{\mcitedefaultseppunct}\relax
\EndOfBibitem
\bibitem[Alayrac \latin{et~al.}(2022)Alayrac, Donahue, Luc, Miech, Barr, Hasson, Lenc, Mensch, Millican, Reynolds, Ring, Rutherford, Cabi, Han, Gong, Samangooei, Monteiro, Menick, Borgeaud, Brock, Nematzadeh, Sharifzadeh, Bi\'{n}kowski, Barreira, Vinyals, Zisserman, and Simonyan]{bib2}
Alayrac,~J.-B. \latin{et~al.}  Flamingo: a Visual Language Model for Few-Shot Learning. Advances in Neural Information Processing Systems. New York, 2022; pp 23716--23736\relax
\mciteBstWouldAddEndPuncttrue
\mciteSetBstMidEndSepPunct{\mcitedefaultmidpunct}
{\mcitedefaultendpunct}{\mcitedefaultseppunct}\relax
\EndOfBibitem
\bibitem[Sun \latin{et~al.}(2019)Sun, Myers, Vondrick, Murphy, and Schmid]{bib3}
Sun,~C.; Myers,~A.; Vondrick,~C.; Murphy,~K.; Schmid,~C. Videobert: A Joint Model for Video and Language Representation Learning. Proceedings of the IEEE International Conference on Computer Vision. 2019; pp 7464--7473\relax
\mciteBstWouldAddEndPuncttrue
\mciteSetBstMidEndSepPunct{\mcitedefaultmidpunct}
{\mcitedefaultendpunct}{\mcitedefaultseppunct}\relax
\EndOfBibitem
\bibitem[Rombach \latin{et~al.}(2022)Rombach, Blattmann, Lorenz, Esser, and Ommer]{bib5}
Rombach,~R.; Blattmann,~A.; Lorenz,~D.; Esser,~P.; Ommer,~B. High-Resolution Image Synthesis With Latent Diffusion Models. Proceedings of the IEEE/CVF Conference on Computer Vision and Pattern Recognition (CVPR). 2022; pp 10684--10695\relax
\mciteBstWouldAddEndPuncttrue
\mciteSetBstMidEndSepPunct{\mcitedefaultmidpunct}
{\mcitedefaultendpunct}{\mcitedefaultseppunct}\relax
\EndOfBibitem
\bibitem[Brodeur \latin{et~al.}(2017)Brodeur, Perez, Anand, Golemo, Celotti, \latin{et~al.} others]{bib6}
Brodeur,~S.; Perez,~E.; Anand,~A.; Golemo,~F.; Celotti,~L.; others HoME: a Household Multimodal Environment. NIPS 2017's Visually-Grounded Interaction and Language Workshop. 2017\relax
\mciteBstWouldAddEndPuncttrue
\mciteSetBstMidEndSepPunct{\mcitedefaultmidpunct}
{\mcitedefaultendpunct}{\mcitedefaultseppunct}\relax
\EndOfBibitem
\bibitem[Savva \latin{et~al.}(2019)Savva, Kadian, Maksymets, Zhao, Wijmans, Jain, Straub, Liu, Koltun, Malik, Parikh, and Batra]{bib7}
Savva,~M.; Kadian,~A.; Maksymets,~O.; Zhao,~Y.; Wijmans,~E.; Jain,~B.; Straub,~J.; Liu,~J.; Koltun,~V.; Malik,~J.; Parikh,~D.; Batra,~D. Habitat: A Platform for Embodied AI Research. 2019 IEEE/CVF International Conference on Computer Vision (ICCV). 2019; pp 9338--9346\relax
\mciteBstWouldAddEndPuncttrue
\mciteSetBstMidEndSepPunct{\mcitedefaultmidpunct}
{\mcitedefaultendpunct}{\mcitedefaultseppunct}\relax
\EndOfBibitem
\bibitem[Zubatiuk and Isayev(2021)Zubatiuk, and Isayev]{bib8}
Zubatiuk,~T.; Isayev,~O. Development of Multimodal Machine Learning Potentials: Toward a Physics-Aware Artificial Intelligence. \emph{Accounts of Chemical Research} \textbf{2021}, \emph{54}, 1575--1585\relax
\mciteBstWouldAddEndPuncttrue
\mciteSetBstMidEndSepPunct{\mcitedefaultmidpunct}
{\mcitedefaultendpunct}{\mcitedefaultseppunct}\relax
\EndOfBibitem
\bibitem[Stahlschmidt \latin{et~al.}(2022)Stahlschmidt, Ulfenborg, and Synnergren]{bib9}
Stahlschmidt,~S.~R.; Ulfenborg,~B.; Synnergren,~J. {Multimodal deep learning for biomedical data fusion: a review}. \emph{Briefings in Bioinformatics} \textbf{2022}, \emph{23}, bbab569\relax
\mciteBstWouldAddEndPuncttrue
\mciteSetBstMidEndSepPunct{\mcitedefaultmidpunct}
{\mcitedefaultendpunct}{\mcitedefaultseppunct}\relax
\EndOfBibitem
\bibitem[Kline \latin{et~al.}(2022)Kline, Wang, Li, Dennis, Hutch, Xu, Wang, Cheng, and Luo]{bib10}
Kline,~A.; Wang,~H.; Li,~Y.; Dennis,~S.; Hutch,~M.; Xu,~Z.; Wang,~F.; Cheng,~F.; Luo,~Y. Multimodal Machine Learning in Precision Health: A Scoping Review. \emph{npj Digital Medicine} \textbf{2022}, \emph{5}, 171\relax
\mciteBstWouldAddEndPuncttrue
\mciteSetBstMidEndSepPunct{\mcitedefaultmidpunct}
{\mcitedefaultendpunct}{\mcitedefaultseppunct}\relax
\EndOfBibitem
\bibitem[Ektefaie \latin{et~al.}(2023)Ektefaie, Dasoulas, Noori, Farhat, and Zitnik]{ektefaie2023multimodal}
Ektefaie,~Y.; Dasoulas,~G.; Noori,~A.; Farhat,~M.; Zitnik,~M. Multimodal learning with graphs. \emph{Nature Machine Intelligence} \textbf{2023}, \emph{5}, 340--350\relax
\mciteBstWouldAddEndPuncttrue
\mciteSetBstMidEndSepPunct{\mcitedefaultmidpunct}
{\mcitedefaultendpunct}{\mcitedefaultseppunct}\relax
\EndOfBibitem
\bibitem[Wen \latin{et~al.}(2023)Wen, Zhang, Rush, Panickan, Li, Cai, Zhou, Ho, Costa, Begoli, \latin{et~al.} others]{wen2023multimodal}
Wen,~J.; Zhang,~X.; Rush,~E.; Panickan,~V.~A.; Li,~X.; Cai,~T.; Zhou,~D.; Ho,~Y.-L.; Costa,~L.; Begoli,~E.; others Multimodal representation learning for predicting molecule--disease relations. \emph{Bioinformatics} \textbf{2023}, \emph{39}, btad085\relax
\mciteBstWouldAddEndPuncttrue
\mciteSetBstMidEndSepPunct{\mcitedefaultmidpunct}
{\mcitedefaultendpunct}{\mcitedefaultseppunct}\relax
\EndOfBibitem
\bibitem[Radford \latin{et~al.}(2021)Radford, Kim, Hallacy, Ramesh, Goh, Agarwal, Sastry, Askell, Mishkin, Clark, \latin{et~al.} others]{radford2021learning}
Radford,~A.; Kim,~J.~W.; Hallacy,~C.; Ramesh,~A.; Goh,~G.; Agarwal,~S.; Sastry,~G.; Askell,~A.; Mishkin,~P.; Clark,~J.; others Learning transferable visual models from natural language supervision. International conference on machine learning. 2021; pp 8748--8763\relax
\mciteBstWouldAddEndPuncttrue
\mciteSetBstMidEndSepPunct{\mcitedefaultmidpunct}
{\mcitedefaultendpunct}{\mcitedefaultseppunct}\relax
\EndOfBibitem
\bibitem[Kim \latin{et~al.}(2021)Kim, Lee, Ahn, and Lee]{bib11}
Kim,~H.; Lee,~J.; Ahn,~S.; Lee,~J.~R. A Merged Molecular Representation Learning for Molecular Properties Prediction with a Web-based Service. \emph{Scientific Reports} \textbf{2021}, \emph{11}, 11028\relax
\mciteBstWouldAddEndPuncttrue
\mciteSetBstMidEndSepPunct{\mcitedefaultmidpunct}
{\mcitedefaultendpunct}{\mcitedefaultseppunct}\relax
\EndOfBibitem
\bibitem[Edwards \latin{et~al.}(2021)Edwards, Zhai, and Ji]{bib12}
Edwards,~C.; Zhai,~C.; Ji,~H. {T}ext2{M}ol: Cross-Modal Molecule Retrieval with Natural Language Queries. Proceedings of the 2021 Conference on Empirical Methods in Natural Language Processing. Online and Punta Cana, Dominican Republic, 2021; pp 595--607\relax
\mciteBstWouldAddEndPuncttrue
\mciteSetBstMidEndSepPunct{\mcitedefaultmidpunct}
{\mcitedefaultendpunct}{\mcitedefaultseppunct}\relax
\EndOfBibitem
\bibitem[Pinheiro \latin{et~al.}(2022)Pinheiro, Da~Silva, and Quiles]{bib13}
Pinheiro,~G.~A.; Da~Silva,~J. L.~F.; Quiles,~M.~G. SMICLR: Contrastive Learning on Multiple Molecular Representations for Semisupervised and Unsupervised Representation Learning. \emph{Journal of Chemical Information and Modeling} \textbf{2022}, \emph{62}, 3948--3960\relax
\mciteBstWouldAddEndPuncttrue
\mciteSetBstMidEndSepPunct{\mcitedefaultmidpunct}
{\mcitedefaultendpunct}{\mcitedefaultseppunct}\relax
\EndOfBibitem
\bibitem[Wang \latin{et~al.}(2022)Wang, Wang, Cao, and Barati~Farimani]{bib14}
Wang,~Y.; Wang,~J.; Cao,~Z.; Barati~Farimani,~A. Molecular Contrastive Learning of Representations via Graph Neural Networks. \emph{Nature Machine Intelligence} \textbf{2022}, \emph{4}, 279--287\relax
\mciteBstWouldAddEndPuncttrue
\mciteSetBstMidEndSepPunct{\mcitedefaultmidpunct}
{\mcitedefaultendpunct}{\mcitedefaultseppunct}\relax
\EndOfBibitem
\bibitem[Su \latin{et~al.}(2022)Su, Du, Yang, Zhou, Li, Rao, Sun, Lu, and Wen]{bib15}
Su,~B.; Du,~D.; Yang,~Z.; Zhou,~Y.; Li,~J.; Rao,~A.; Sun,~H.; Lu,~Z.; Wen,~J.-R. A Molecular Multimodal Foundation Model Associating Molecule Graphs with Natural Language. \textbf{2022}, \relax
\mciteBstWouldAddEndPunctfalse
\mciteSetBstMidEndSepPunct{\mcitedefaultmidpunct}
{}{\mcitedefaultseppunct}\relax
\EndOfBibitem
\bibitem[Ektefaie \latin{et~al.}(2023)Ektefaie, Dasoulas, Noori, Farhat, and Zitnik]{bib16}
Ektefaie,~Y.; Dasoulas,~G.; Noori,~A.; Farhat,~M.; Zitnik,~M. Multimodal Learning with Graphs. \emph{Nature Machine Intelligence} \textbf{2023}, \emph{5}, 340--350\relax
\mciteBstWouldAddEndPuncttrue
\mciteSetBstMidEndSepPunct{\mcitedefaultmidpunct}
{\mcitedefaultendpunct}{\mcitedefaultseppunct}\relax
\EndOfBibitem
\bibitem[Fang \latin{et~al.}(2023)Fang, Zhang, Zhang, Chen, Zhuang, Shao, Fan, and Chen]{bib17}
Fang,~Y.; Zhang,~Q.; Zhang,~N.; Chen,~Z.; Zhuang,~X.; Shao,~X.; Fan,~X.; Chen,~H. Knowledge Graph-Enhanced Molecular Contrastive Learning with Functional Prompt. \emph{Nature Machine Intelligence} \textbf{2023}, \emph{5}, 542--553\relax
\mciteBstWouldAddEndPuncttrue
\mciteSetBstMidEndSepPunct{\mcitedefaultmidpunct}
{\mcitedefaultendpunct}{\mcitedefaultseppunct}\relax
\EndOfBibitem
\bibitem[Yang \latin{et~al.}(2021)Yang, Song, Yang, Yao, Zhang, Shi, Ji, Deng, and Wang]{bib18}
Yang,~Z.; Song,~J.; Yang,~M.; Yao,~L.; Zhang,~J.; Shi,~H.; Ji,~X.; Deng,~Y.; Wang,~X. Cross-Modal Retrieval between 13C NMR Spectra and Structures for Compound Identification Using Deep Contrastive Learning. \emph{Analytical Chemistry} \textbf{2021}, \emph{93}, 16947--16955\relax
\mciteBstWouldAddEndPuncttrue
\mciteSetBstMidEndSepPunct{\mcitedefaultmidpunct}
{\mcitedefaultendpunct}{\mcitedefaultseppunct}\relax
\EndOfBibitem
\bibitem[Zhang \latin{et~al.}(2020)Zhang, Xie, Cai, Hu, Liu, Hong, and Tian]{zhang2020transfer}
Zhang,~R.; Xie,~H.; Cai,~S.; Hu,~Y.; Liu,~G.-k.; Hong,~W.; Tian,~Z.-q. Transfer-learning-based Raman spectra identification. \emph{Journal of Raman Spectroscopy} \textbf{2020}, \emph{51}, 176--186\relax
\mciteBstWouldAddEndPuncttrue
\mciteSetBstMidEndSepPunct{\mcitedefaultmidpunct}
{\mcitedefaultendpunct}{\mcitedefaultseppunct}\relax
\EndOfBibitem
\bibitem[Gao \latin{et~al.}(2020)Gao, Nguyen, Sresht, Mathiowetz, Tu, and Wei]{gao20202d}
Gao,~K.; Nguyen,~D.~D.; Sresht,~V.; Mathiowetz,~A.~M.; Tu,~M.; Wei,~G.-W. Are 2D fingerprints still valuable for drug discovery? \emph{Physical chemistry chemical physics} \textbf{2020}, \emph{22}, 8373--8390\relax
\mciteBstWouldAddEndPuncttrue
\mciteSetBstMidEndSepPunct{\mcitedefaultmidpunct}
{\mcitedefaultendpunct}{\mcitedefaultseppunct}\relax
\EndOfBibitem
\bibitem[Li \latin{et~al.}(2023)Li, Wang, and Ma]{li2023deep}
Li,~W.; Wang,~G.; Ma,~J. {Deep learning for complex chemical systems}. \emph{National Science Review} \textbf{2023}, \emph{10}, nwad335\relax
\mciteBstWouldAddEndPuncttrue
\mciteSetBstMidEndSepPunct{\mcitedefaultmidpunct}
{\mcitedefaultendpunct}{\mcitedefaultseppunct}\relax
\EndOfBibitem
\bibitem[Zou \latin{et~al.}(2023)Zou, Zhang, Liang, Wei, Leng, Jiang, Luo, and Hu]{zou2023deep}
Zou,~Z.; Zhang,~Y.; Liang,~L.; Wei,~M.; Leng,~J.; Jiang,~J.; Luo,~Y.; Hu,~W. A deep learning model for predicting selected organic molecular spectra. \emph{Nature Computational Science} \textbf{2023}, 1--8\relax
\mciteBstWouldAddEndPuncttrue
\mciteSetBstMidEndSepPunct{\mcitedefaultmidpunct}
{\mcitedefaultendpunct}{\mcitedefaultseppunct}\relax
\EndOfBibitem
\bibitem[Hirst \latin{et~al.}(2023)Hirst, Boobier, Coughlan, Streets, Jacob, Pugh, {\"O}zcan, and Woodward]{hirst2023ml}
Hirst,~J.~D.; Boobier,~S.; Coughlan,~J.; Streets,~J.; Jacob,~P.~L.; Pugh,~O.; {\"O}zcan,~E.; Woodward,~S. ML meets MLn: machine learning in ligand promoted homogeneous catalysis. \emph{Artificial Intelligence Chemistry} \textbf{2023}, 100006\relax
\mciteBstWouldAddEndPuncttrue
\mciteSetBstMidEndSepPunct{\mcitedefaultmidpunct}
{\mcitedefaultendpunct}{\mcitedefaultseppunct}\relax
\EndOfBibitem
\bibitem[Lai \latin{et~al.}(2023)Lai, Tew, Zhong, Yin, Li, Yan, and Wang]{lai2023artificial}
Lai,~N.~S.; Tew,~Y.~S.; Zhong,~X.; Yin,~J.; Li,~J.; Yan,~B.; Wang,~X. Artificial Intelligence (AI) Workflow for Catalyst Design and Optimization. \emph{Industrial \& Engineering Chemistry Research} \textbf{2023}, \emph{62}, 17835--17848\relax
\mciteBstWouldAddEndPuncttrue
\mciteSetBstMidEndSepPunct{\mcitedefaultmidpunct}
{\mcitedefaultendpunct}{\mcitedefaultseppunct}\relax
\EndOfBibitem
\bibitem[Tao \latin{et~al.}(2024)Tao, Feng, Wang, Han, Smith, and Jiang]{tao2024machine}
Tao,~S.; Feng,~Y.; Wang,~W.; Han,~T.; Smith,~P.~E.; Jiang,~J. A machine learning protocol for geometric information retrieval from molecular spectra. \emph{Artificial Intelligence Chemistry} \textbf{2024}, \emph{2}, 100031\relax
\mciteBstWouldAddEndPuncttrue
\mciteSetBstMidEndSepPunct{\mcitedefaultmidpunct}
{\mcitedefaultendpunct}{\mcitedefaultseppunct}\relax
\EndOfBibitem
\bibitem[Clevert \latin{et~al.}(2021)Clevert, Le, Winter, and Montanari]{clevert2021img2mol}
Clevert,~D.-A.; Le,~T.; Winter,~R.; Montanari,~F. Img2Mol--accurate SMILES recognition from molecular graphical depictions. \emph{Chemical science} \textbf{2021}, \emph{12}, 14174--14181\relax
\mciteBstWouldAddEndPuncttrue
\mciteSetBstMidEndSepPunct{\mcitedefaultmidpunct}
{\mcitedefaultendpunct}{\mcitedefaultseppunct}\relax
\EndOfBibitem
\bibitem[Wang \latin{et~al.}(2022)Wang, Wang, Cao, and Barati~Farimani]{wang2022molecular}
Wang,~Y.; Wang,~J.; Cao,~Z.; Barati~Farimani,~A. Molecular contrastive learning of representations via graph neural networks. \emph{Nature Machine Intelligence} \textbf{2022}, \emph{4}, 279--287\relax
\mciteBstWouldAddEndPuncttrue
\mciteSetBstMidEndSepPunct{\mcitedefaultmidpunct}
{\mcitedefaultendpunct}{\mcitedefaultseppunct}\relax
\EndOfBibitem
\bibitem[Weininger(1988)]{weininger1988smiles}
Weininger,~D. SMILES, a chemical language and information system. 1. Introduction to methodology and encoding rules. \emph{Journal of chemical information and computer sciences} \textbf{1988}, \emph{28}, 31--36\relax
\mciteBstWouldAddEndPuncttrue
\mciteSetBstMidEndSepPunct{\mcitedefaultmidpunct}
{\mcitedefaultendpunct}{\mcitedefaultseppunct}\relax
\EndOfBibitem
\bibitem[Zong \latin{et~al.}(2023)Zong, Aodha, and Hospedales]{zong2023self}
Zong,~Y.; Aodha,~O.~M.; Hospedales,~T. Self-Supervised Multimodal Learning: A Survey. \textbf{2023}, \relax
\mciteBstWouldAddEndPunctfalse
\mciteSetBstMidEndSepPunct{\mcitedefaultmidpunct}
{}{\mcitedefaultseppunct}\relax
\EndOfBibitem
\bibitem[Guo \latin{et~al.}(2023)Guo, Xue, Sun, Jiang, and Pu]{bib21}
Guo,~H.; Xue,~K.; Sun,~H.; Jiang,~W.; Pu,~S. Contrastive Learning-Based Embedder for the Representation of Tandem Mass Spectra. \emph{Analytical Chemistry} \textbf{2023}, \emph{95}, 7888--7896\relax
\mciteBstWouldAddEndPuncttrue
\mciteSetBstMidEndSepPunct{\mcitedefaultmidpunct}
{\mcitedefaultendpunct}{\mcitedefaultseppunct}\relax
\EndOfBibitem
\bibitem[Zeng \latin{et~al.}(2022)Zeng, Xiang, Yu, Wang, Li, Nussinov, and Cheng]{zeng2022accurate}
Zeng,~X.; Xiang,~H.; Yu,~L.; Wang,~J.; Li,~K.; Nussinov,~R.; Cheng,~F. Accurate prediction of molecular properties and drug targets using a self-supervised image representation learning framework. \emph{Nature Machine Intelligence} \textbf{2022}, \emph{4}, 1004--1016\relax
\mciteBstWouldAddEndPuncttrue
\mciteSetBstMidEndSepPunct{\mcitedefaultmidpunct}
{\mcitedefaultendpunct}{\mcitedefaultseppunct}\relax
\EndOfBibitem
\bibitem[Li \latin{et~al.}(2022)Li, Feng, Liu, Yao, \latin{et~al.} others]{li2022novel}
Li,~C.; Feng,~J.; Liu,~S.; Yao,~J.; others A novel molecular representation learning for molecular property prediction with a multiple SMILES-based augmentation. \emph{Computational Intelligence and Neuroscience} \textbf{2022}, \emph{2022}\relax
\mciteBstWouldAddEndPuncttrue
\mciteSetBstMidEndSepPunct{\mcitedefaultmidpunct}
{\mcitedefaultendpunct}{\mcitedefaultseppunct}\relax
\EndOfBibitem
\bibitem[Guo \latin{et~al.}(2024)Guo, Fan, Yu, Lu, and Zhang]{guo2024gcmsformer}
Guo,~Z.; Fan,~Y.; Yu,~C.; Lu,~H.; Zhang,~Z. GCMSFormer: A Fully Automatic Method for the Resolution of Overlapping Peaks in Gas Chromatography--Mass Spectrometry. \emph{Analytical Chemistry} \textbf{2024}, \emph{96}, 5878--5886\relax
\mciteBstWouldAddEndPuncttrue
\mciteSetBstMidEndSepPunct{\mcitedefaultmidpunct}
{\mcitedefaultendpunct}{\mcitedefaultseppunct}\relax
\EndOfBibitem
\bibitem[Jiang \latin{et~al.}(2021)Jiang, Wu, Hsieh, Chen, Liao, Wang, Shen, Cao, Wu, and Hou]{jiang2021could}
Jiang,~D.; Wu,~Z.; Hsieh,~C.-Y.; Chen,~G.; Liao,~B.; Wang,~Z.; Shen,~C.; Cao,~D.; Wu,~J.; Hou,~T. Could graph neural networks learn better molecular representation for drug discovery? A comparison study of descriptor-based and graph-based models. \emph{Journal of cheminformatics} \textbf{2021}, \emph{13}, 1--23\relax
\mciteBstWouldAddEndPuncttrue
\mciteSetBstMidEndSepPunct{\mcitedefaultmidpunct}
{\mcitedefaultendpunct}{\mcitedefaultseppunct}\relax
\EndOfBibitem
\bibitem[David \latin{et~al.}(2020)David, Thakkar, Mercado, and Engkvist]{bib22}
David,~L.; Thakkar,~A.; Mercado,~R.; Engkvist,~O. Molecular Representations in AI-driven Drug Discovery: A Review and Practical Guide. \emph{Journal of Cheminformatics} \textbf{2020}, \emph{12}, 56\relax
\mciteBstWouldAddEndPuncttrue
\mciteSetBstMidEndSepPunct{\mcitedefaultmidpunct}
{\mcitedefaultendpunct}{\mcitedefaultseppunct}\relax
\EndOfBibitem
\bibitem[Gilmer \latin{et~al.}(2017)Gilmer, Schoenholz, Riley, Vinyals, and Dahl]{gilmer2017neural}
Gilmer,~J.; Schoenholz,~S.~S.; Riley,~P.~F.; Vinyals,~O.; Dahl,~G.~E. Neural message passing for quantum chemistry. International conference on machine learning. 2017; pp 1263--1272\relax
\mciteBstWouldAddEndPuncttrue
\mciteSetBstMidEndSepPunct{\mcitedefaultmidpunct}
{\mcitedefaultendpunct}{\mcitedefaultseppunct}\relax
\EndOfBibitem
\bibitem[Alsentzer \latin{et~al.}(2020)Alsentzer, Finlayson, Li, and Zitnik]{alsentzer2020subgraph}
Alsentzer,~E.; Finlayson,~S.; Li,~M.; Zitnik,~M. Subgraph neural networks. \emph{Advances in Neural Information Processing Systems} \textbf{2020}, \emph{33}, 8017--8029\relax
\mciteBstWouldAddEndPuncttrue
\mciteSetBstMidEndSepPunct{\mcitedefaultmidpunct}
{\mcitedefaultendpunct}{\mcitedefaultseppunct}\relax
\EndOfBibitem
\bibitem[Sun \latin{et~al.}(2021)Sun, Li, Peng, Wu, Ning, Yu, and He]{sun2021sugar}
Sun,~Q.; Li,~J.; Peng,~H.; Wu,~J.; Ning,~Y.; Yu,~P.~S.; He,~L. Sugar: Subgraph neural network with reinforcement pooling and self-supervised mutual information mechanism. Proceedings of the Web Conference 2021. 2021; pp 2081--2091\relax
\mciteBstWouldAddEndPuncttrue
\mciteSetBstMidEndSepPunct{\mcitedefaultmidpunct}
{\mcitedefaultendpunct}{\mcitedefaultseppunct}\relax
\EndOfBibitem
\bibitem[Guo \latin{et~al.}(2022)Guo, Nan, Tian, Wiest, Zhang, and Chawla]{guo2022graph}
Guo,~Z.; Nan,~B.; Tian,~Y.; Wiest,~O.; Zhang,~C.; Chawla,~N.~V. Graph-based molecular representation learning. \emph{arXiv preprint arXiv:2207.04869} \textbf{2022}, \relax
\mciteBstWouldAddEndPunctfalse
\mciteSetBstMidEndSepPunct{\mcitedefaultmidpunct}
{}{\mcitedefaultseppunct}\relax
\EndOfBibitem
\bibitem[Sun \latin{et~al.}(2022)Sun, Wen, Wang, Ruan, Yang, Kang, Zhang, Zhang, and Lu]{sun2022prediction}
Sun,~J.; Wen,~M.; Wang,~H.; Ruan,~Y.; Yang,~Q.; Kang,~X.; Zhang,~H.; Zhang,~Z.; Lu,~H. Prediction of drug-likeness using graph convolutional attention network. \emph{Bioinformatics} \textbf{2022}, \emph{38}, 5262--5269\relax
\mciteBstWouldAddEndPuncttrue
\mciteSetBstMidEndSepPunct{\mcitedefaultmidpunct}
{\mcitedefaultendpunct}{\mcitedefaultseppunct}\relax
\EndOfBibitem
\bibitem[Wang \latin{et~al.}(2023)Wang, Chen, Chen, Shurberg, Liu, and Hong]{wang2023motif}
Wang,~Y.; Chen,~S.; Chen,~G.; Shurberg,~E.; Liu,~H.; Hong,~P. Motif-Based Graph Representation Learning with Application to Chemical Molecules. Informatics. 2023; p~8\relax
\mciteBstWouldAddEndPuncttrue
\mciteSetBstMidEndSepPunct{\mcitedefaultmidpunct}
{\mcitedefaultendpunct}{\mcitedefaultseppunct}\relax
\EndOfBibitem
\bibitem[Mucllari \latin{et~al.}(2023)Mucllari, Zadorozhnyy, Ye, and Nguyen]{mucllari2023novel}
Mucllari,~E.; Zadorozhnyy,~V.; Ye,~Q.; Nguyen,~D.~D. Novel Molecular Representations Using Neumann-Cayley Orthogonal Gated Recurrent Unit. \emph{Journal of Chemical Information and Modeling} \textbf{2023}, \emph{63}, 2656--2666\relax
\mciteBstWouldAddEndPuncttrue
\mciteSetBstMidEndSepPunct{\mcitedefaultmidpunct}
{\mcitedefaultendpunct}{\mcitedefaultseppunct}\relax
\EndOfBibitem
\bibitem[Wu \latin{et~al.}(2018)Wu, Zhao, Wang, and Wei]{wu2018topp}
Wu,~K.; Zhao,~Z.; Wang,~R.; Wei,~G.-W. TopP--S: Persistent homology-based multi-task deep neural networks for simultaneous predictions of partition coefficient and aqueous solubility. \emph{Journal of computational chemistry} \textbf{2018}, \emph{39}, 1444--1454\relax
\mciteBstWouldAddEndPuncttrue
\mciteSetBstMidEndSepPunct{\mcitedefaultmidpunct}
{\mcitedefaultendpunct}{\mcitedefaultseppunct}\relax
\EndOfBibitem
\bibitem[Wu and Wei(2018)Wu, and Wei]{wu2018quantitative}
Wu,~K.; Wei,~G.-W. Quantitative toxicity prediction using topology based multitask deep neural networks. \emph{Journal of chemical information and modeling} \textbf{2018}, \emph{58}, 520--531\relax
\mciteBstWouldAddEndPuncttrue
\mciteSetBstMidEndSepPunct{\mcitedefaultmidpunct}
{\mcitedefaultendpunct}{\mcitedefaultseppunct}\relax
\EndOfBibitem
\bibitem[Schroff \latin{et~al.}(2015)Schroff, Kalenichenko, and Philbin]{schroff2015facenet}
Schroff,~F.; Kalenichenko,~D.; Philbin,~J. Facenet: A unified embedding for face recognition and clustering. Proceedings of the IEEE conference on computer vision and pattern recognition. 2015; pp 815--823\relax
\mciteBstWouldAddEndPuncttrue
\mciteSetBstMidEndSepPunct{\mcitedefaultmidpunct}
{\mcitedefaultendpunct}{\mcitedefaultseppunct}\relax
\EndOfBibitem
\bibitem[Yu \latin{et~al.}(2023)Yu, Cui, Li, Luo, Jiang, and Zhao]{yu2023enzyme}
Yu,~T.; Cui,~H.; Li,~J.~C.; Luo,~Y.; Jiang,~G.; Zhao,~H. Enzyme function prediction using contrastive learning. \emph{Science} \textbf{2023}, \emph{379}, 1358--1363\relax
\mciteBstWouldAddEndPuncttrue
\mciteSetBstMidEndSepPunct{\mcitedefaultmidpunct}
{\mcitedefaultendpunct}{\mcitedefaultseppunct}\relax
\EndOfBibitem
\bibitem[Jiang \latin{et~al.}(2024)Jiang, Nguyen, Ishwar, and Aeron]{jiang2024supervised}
Jiang,~R.; Nguyen,~T.; Ishwar,~P.; Aeron,~S. Supervised contrastive learning with hard negative samples. 2024 International Joint Conference on Neural Networks (IJCNN). 2024; pp 1--8\relax
\mciteBstWouldAddEndPuncttrue
\mciteSetBstMidEndSepPunct{\mcitedefaultmidpunct}
{\mcitedefaultendpunct}{\mcitedefaultseppunct}\relax
\EndOfBibitem
\bibitem[Landrum(2013)]{landrum2013rdkit}
Landrum,~G. Rdkit documentation. \emph{Release} \textbf{2013}, \emph{1}, 4\relax
\mciteBstWouldAddEndPuncttrue
\mciteSetBstMidEndSepPunct{\mcitedefaultmidpunct}
{\mcitedefaultendpunct}{\mcitedefaultseppunct}\relax
\EndOfBibitem
\bibitem[Tanimoto(1958)]{tanimoto1958elementary}
Tanimoto,~T.~T. Elementary mathematical theory of classification and prediction. \textbf{1958}, \relax
\mciteBstWouldAddEndPunctfalse
\mciteSetBstMidEndSepPunct{\mcitedefaultmidpunct}
{}{\mcitedefaultseppunct}\relax
\EndOfBibitem
\bibitem[Jackson(2005)]{jackson2005user}
Jackson,~J.~E. \emph{A user's guide to principal components}; John Wiley \& Sons, 2005\relax
\mciteBstWouldAddEndPuncttrue
\mciteSetBstMidEndSepPunct{\mcitedefaultmidpunct}
{\mcitedefaultendpunct}{\mcitedefaultseppunct}\relax
\EndOfBibitem
\bibitem[Pearson(1920)]{pearson1920notes}
Pearson,~K. Notes on the history of correlation. \emph{Biometrika} \textbf{1920}, \emph{13}, 25--45\relax
\mciteBstWouldAddEndPuncttrue
\mciteSetBstMidEndSepPunct{\mcitedefaultmidpunct}
{\mcitedefaultendpunct}{\mcitedefaultseppunct}\relax
\EndOfBibitem
\bibitem[Lu \latin{et~al.}(2021)Lu, Jiang, Fang, and Shi]{lu2021learning}
Lu,~Y.; Jiang,~X.; Fang,~Y.; Shi,~C. Learning to pre-train graph neural networks. Proceedings of the AAAI conference on artificial intelligence. 2021; pp 4276--4284\relax
\mciteBstWouldAddEndPuncttrue
\mciteSetBstMidEndSepPunct{\mcitedefaultmidpunct}
{\mcitedefaultendpunct}{\mcitedefaultseppunct}\relax
\EndOfBibitem
\bibitem[Zhang \latin{et~al.}(2021)Zhang, Liu, Wang, Lu, and Lee]{zhang2021motif}
Zhang,~Z.; Liu,~Q.; Wang,~H.; Lu,~C.; Lee,~C.-K. Motif-based Graph Self-Supervised Learning for Molecular Property Prediction. \emph{Advances in Neural Information Processing Systems} \textbf{2021}, \emph{34}\relax
\mciteBstWouldAddEndPuncttrue
\mciteSetBstMidEndSepPunct{\mcitedefaultmidpunct}
{\mcitedefaultendpunct}{\mcitedefaultseppunct}\relax
\EndOfBibitem
\bibitem[Su \latin{et~al.}(2022)Su, Du, Yang, Zhou, Li, Rao, Sun, Lu, and Wen]{su2022molecular}
Su,~B.; Du,~D.; Yang,~Z.; Zhou,~Y.; Li,~J.; Rao,~A.; Sun,~H.; Lu,~Z.; Wen,~J.-R. A molecular multimodal foundation model associating molecule graphs with natural language. \emph{arXiv preprint arXiv:2209.05481} \textbf{2022}, \relax
\mciteBstWouldAddEndPunctfalse
\mciteSetBstMidEndSepPunct{\mcitedefaultmidpunct}
{}{\mcitedefaultseppunct}\relax
\EndOfBibitem
\bibitem[Wu \latin{et~al.}(2018)Wu, Ramsundar, Feinberg, Gomes, Geniesse, Pappu, Leswing, and Pande]{wu2018moleculenet}
Wu,~Z.; Ramsundar,~B.; Feinberg,~E.~N.; Gomes,~J.; Geniesse,~C.; Pappu,~A.~S.; Leswing,~K.; Pande,~V. MoleculeNet: a benchmark for molecular machine learning. \emph{Chemical science} \textbf{2018}, \emph{9}, 513--530\relax
\mciteBstWouldAddEndPuncttrue
\mciteSetBstMidEndSepPunct{\mcitedefaultmidpunct}
{\mcitedefaultendpunct}{\mcitedefaultseppunct}\relax
\EndOfBibitem
\bibitem[Huang \latin{et~al.}(2022)Huang, Fu, Gao, Zhao, Roohani, Leskovec, Coley, Xiao, Sun, and Zitnik]{huang2022artificial}
Huang,~K.; Fu,~T.; Gao,~W.; Zhao,~Y.; Roohani,~Y.; Leskovec,~J.; Coley,~C.~W.; Xiao,~C.; Sun,~J.; Zitnik,~M. Artificial intelligence foundation for therapeutic science. \emph{Nature chemical biology} \textbf{2022}, \emph{18}, 1033--1036\relax
\mciteBstWouldAddEndPuncttrue
\mciteSetBstMidEndSepPunct{\mcitedefaultmidpunct}
{\mcitedefaultendpunct}{\mcitedefaultseppunct}\relax
\EndOfBibitem
\bibitem[Hu \latin{et~al.}(2020)Hu, Fey, Zitnik, Dong, Ren, Liu, Catasta, and Leskovec]{hu2020ogb}
Hu,~W.; Fey,~M.; Zitnik,~M.; Dong,~Y.; Ren,~H.; Liu,~B.; Catasta,~M.; Leskovec,~J. Open Graph Benchmark: Datasets for Machine Learning on Graphs. \emph{arXiv preprint arXiv:2005.00687} \textbf{2020}, \relax
\mciteBstWouldAddEndPunctfalse
\mciteSetBstMidEndSepPunct{\mcitedefaultmidpunct}
{}{\mcitedefaultseppunct}\relax
\EndOfBibitem
\bibitem[Xu \latin{et~al.}(2019)Xu, Hu, Leskovec, and Jegelka]{xu2018how}
Xu,~K.; Hu,~W.; Leskovec,~J.; Jegelka,~S. How Powerful are Graph Neural Networks? International Conference on Learning Representations. 2019\relax
\mciteBstWouldAddEndPuncttrue
\mciteSetBstMidEndSepPunct{\mcitedefaultmidpunct}
{\mcitedefaultendpunct}{\mcitedefaultseppunct}\relax
\EndOfBibitem
\bibitem[Wishart \latin{et~al.}(2022)Wishart, Sayeeda, Budinski, Guo, Lee, Berjanskii, Rout, Peters, Dizon, Mah, \latin{et~al.} others]{wishart2022np}
Wishart,~D.~S.; Sayeeda,~Z.; Budinski,~Z.; Guo,~A.; Lee,~B.~L.; Berjanskii,~M.; Rout,~M.; Peters,~H.; Dizon,~R.; Mah,~R.; others NP-MRD: the natural products magnetic resonance database. \emph{Nucleic Acids Research} \textbf{2022}, \emph{50}, D665--D677\relax
\mciteBstWouldAddEndPuncttrue
\mciteSetBstMidEndSepPunct{\mcitedefaultmidpunct}
{\mcitedefaultendpunct}{\mcitedefaultseppunct}\relax
\EndOfBibitem
\bibitem[Kim \latin{et~al.}(2023)Kim, Chen, Cheng, Gindulyte, He, He, Li, Shoemaker, Thiessen, Yu, \latin{et~al.} others]{kim2023pubchem}
Kim,~S.; Chen,~J.; Cheng,~T.; Gindulyte,~A.; He,~J.; He,~S.; Li,~Q.; Shoemaker,~B.~A.; Thiessen,~P.~A.; Yu,~B.; others PubChem 2023 update. \emph{Nucleic acids research} \textbf{2023}, \emph{51}, D1373--D1380\relax
\mciteBstWouldAddEndPuncttrue
\mciteSetBstMidEndSepPunct{\mcitedefaultmidpunct}
{\mcitedefaultendpunct}{\mcitedefaultseppunct}\relax
\EndOfBibitem
\bibitem[RDK()]{RDKit}
{RDKit}: Open-source cheminformatics. \url{http://www.rdkit.org}\relax
\mciteBstWouldAddEndPuncttrue
\mciteSetBstMidEndSepPunct{\mcitedefaultmidpunct}
{\mcitedefaultendpunct}{\mcitedefaultseppunct}\relax
\EndOfBibitem
\bibitem[Yang \latin{et~al.}(2021)Yang, Song, Yang, Yao, Zhang, Shi, Ji, Deng, and Wang]{yang2021cross}
Yang,~Z.; Song,~J.; Yang,~M.; Yao,~L.; Zhang,~J.; Shi,~H.; Ji,~X.; Deng,~Y.; Wang,~X. Cross-modal retrieval between 13C NMR spectra and structures for compound identification using deep contrastive learning. \emph{Analytical Chemistry} \textbf{2021}, \emph{93}, 16947--16955\relax
\mciteBstWouldAddEndPuncttrue
\mciteSetBstMidEndSepPunct{\mcitedefaultmidpunct}
{\mcitedefaultendpunct}{\mcitedefaultseppunct}\relax
\EndOfBibitem
\bibitem[Costanti \latin{et~al.}(2023)Costanti, Kola, Scarselli, Valensin, and Bianchini]{costanti2023deep}
Costanti,~F.; Kola,~A.; Scarselli,~F.; Valensin,~D.; Bianchini,~M. A Deep Learning Approach to Analyze NMR Spectra of SH-SY5Y Cells for Alzheimer’s Disease Diagnosis. \emph{Mathematics} \textbf{2023}, \emph{11}, 2664\relax
\mciteBstWouldAddEndPuncttrue
\mciteSetBstMidEndSepPunct{\mcitedefaultmidpunct}
{\mcitedefaultendpunct}{\mcitedefaultseppunct}\relax
\EndOfBibitem
\bibitem[Wu \latin{et~al.}(2020)Wu, Pan, Chen, Long, Zhang, and Philip]{wu2020comprehensive}
Wu,~Z.; Pan,~S.; Chen,~F.; Long,~G.; Zhang,~C.; Philip,~S.~Y. A comprehensive survey on graph neural networks. \emph{IEEE transactions on neural networks and learning systems} \textbf{2020}, \emph{32}, 4--24\relax
\mciteBstWouldAddEndPuncttrue
\mciteSetBstMidEndSepPunct{\mcitedefaultmidpunct}
{\mcitedefaultendpunct}{\mcitedefaultseppunct}\relax
\EndOfBibitem
\bibitem[Zhou \latin{et~al.}(2020)Zhou, Cui, Hu, Zhang, Yang, Liu, Wang, Li, and Sun]{zhou2020graph}
Zhou,~J.; Cui,~G.; Hu,~S.; Zhang,~Z.; Yang,~C.; Liu,~Z.; Wang,~L.; Li,~C.; Sun,~M. Graph neural networks: A review of methods and applications. \emph{AI open} \textbf{2020}, \emph{1}, 57--81\relax
\mciteBstWouldAddEndPuncttrue
\mciteSetBstMidEndSepPunct{\mcitedefaultmidpunct}
{\mcitedefaultendpunct}{\mcitedefaultseppunct}\relax
\EndOfBibitem
\bibitem[LeCun \latin{et~al.}(2015)LeCun, Bengio, and Hinton]{lecun2015deep}
LeCun,~Y.; Bengio,~Y.; Hinton,~G. Deep learning. \emph{nature} \textbf{2015}, \emph{521}, 436--444\relax
\mciteBstWouldAddEndPuncttrue
\mciteSetBstMidEndSepPunct{\mcitedefaultmidpunct}
{\mcitedefaultendpunct}{\mcitedefaultseppunct}\relax
\EndOfBibitem
\bibitem[Baskin \latin{et~al.}(1997)Baskin, Palyulin, and Zefirov]{baskin1997neural}
Baskin,~I.~I.; Palyulin,~V.~A.; Zefirov,~N.~S. A neural device for searching direct correlations between structures and properties of chemical compounds. \emph{Journal of chemical information and computer sciences} \textbf{1997}, \emph{37}, 715--721\relax
\mciteBstWouldAddEndPuncttrue
\mciteSetBstMidEndSepPunct{\mcitedefaultmidpunct}
{\mcitedefaultendpunct}{\mcitedefaultseppunct}\relax
\EndOfBibitem
\bibitem[Sperduti and Starita(1997)Sperduti, and Starita]{sperduti1997supervised}
Sperduti,~A.; Starita,~A. Supervised neural networks for the classification of structures. \emph{IEEE Transactions on Neural Networks} \textbf{1997}, \emph{8}, 714--735\relax
\mciteBstWouldAddEndPuncttrue
\mciteSetBstMidEndSepPunct{\mcitedefaultmidpunct}
{\mcitedefaultendpunct}{\mcitedefaultseppunct}\relax
\EndOfBibitem
\bibitem[Gori \latin{et~al.}(2005)Gori, Monfardini, and Scarselli]{gori2005new}
Gori,~M.; Monfardini,~G.; Scarselli,~F. A new model for learning in graph domains. Proceedings. 2005 IEEE International Joint Conference on Neural Networks, 2005. 2005; pp 729--734\relax
\mciteBstWouldAddEndPuncttrue
\mciteSetBstMidEndSepPunct{\mcitedefaultmidpunct}
{\mcitedefaultendpunct}{\mcitedefaultseppunct}\relax
\EndOfBibitem
\bibitem[Kearnes \latin{et~al.}(2016)Kearnes, McCloskey, Berndl, Pande, and Riley]{kearnes2016molecular}
Kearnes,~S.; McCloskey,~K.; Berndl,~M.; Pande,~V.; Riley,~P. Molecular graph convolutions: moving beyond fingerprints. \emph{Journal of computer-aided molecular design} \textbf{2016}, \emph{30}, 595--608\relax
\mciteBstWouldAddEndPuncttrue
\mciteSetBstMidEndSepPunct{\mcitedefaultmidpunct}
{\mcitedefaultendpunct}{\mcitedefaultseppunct}\relax
\EndOfBibitem
\bibitem[Kipf and Welling(2017)Kipf, and Welling]{kipf2017semisupervised}
Kipf,~T.~N.; Welling,~M. Semi-Supervised Classification with Graph Convolutional Networks. International Conference on Learning Representations. 2017\relax
\mciteBstWouldAddEndPuncttrue
\mciteSetBstMidEndSepPunct{\mcitedefaultmidpunct}
{\mcitedefaultendpunct}{\mcitedefaultseppunct}\relax
\EndOfBibitem
\bibitem[Velickovic \latin{et~al.}(2017)Velickovic, Cucurull, Casanova, Romero, Lio, Bengio, \latin{et~al.} others]{velickovic2017graph}
Velickovic,~P.; Cucurull,~G.; Casanova,~A.; Romero,~A.; Lio,~P.; Bengio,~Y.; others Graph attention networks. \emph{stat} \textbf{2017}, \emph{1050}, 10--48550\relax
\mciteBstWouldAddEndPuncttrue
\mciteSetBstMidEndSepPunct{\mcitedefaultmidpunct}
{\mcitedefaultendpunct}{\mcitedefaultseppunct}\relax
\EndOfBibitem
\bibitem[Hamilton \latin{et~al.}(2017)Hamilton, Ying, and Leskovec]{hamilton2017inductive}
Hamilton,~W.; Ying,~Z.; Leskovec,~J. Inductive representation learning on large graphs. \emph{Advances in neural information processing systems} \textbf{2017}, \emph{30}\relax
\mciteBstWouldAddEndPuncttrue
\mciteSetBstMidEndSepPunct{\mcitedefaultmidpunct}
{\mcitedefaultendpunct}{\mcitedefaultseppunct}\relax
\EndOfBibitem
\bibitem[Gu \latin{et~al.}(2018)Gu, Wang, Kuen, Ma, Shahroudy, Shuai, Liu, Wang, Wang, Cai, \latin{et~al.} others]{gu2018recent}
Gu,~J.; Wang,~Z.; Kuen,~J.; Ma,~L.; Shahroudy,~A.; Shuai,~B.; Liu,~T.; Wang,~X.; Wang,~G.; Cai,~J.; others Recent advances in convolutional neural networks. \emph{Pattern recognition} \textbf{2018}, \emph{77}, 354--377\relax
\mciteBstWouldAddEndPuncttrue
\mciteSetBstMidEndSepPunct{\mcitedefaultmidpunct}
{\mcitedefaultendpunct}{\mcitedefaultseppunct}\relax
\EndOfBibitem
\bibitem[Cangea \latin{et~al.}(2018)Cangea, Veli{\v{c}}kovi{\'c}, Jovanovi{\'c}, Kipf, and Li{\`o}]{cangea2018towards}
Cangea,~C.; Veli{\v{c}}kovi{\'c},~P.; Jovanovi{\'c},~N.; Kipf,~T.; Li{\`o},~P. Towards sparse hierarchical graph classifiers. \emph{arXiv preprint arXiv:1811.01287} \textbf{2018}, \relax
\mciteBstWouldAddEndPunctfalse
\mciteSetBstMidEndSepPunct{\mcitedefaultmidpunct}
{}{\mcitedefaultseppunct}\relax
\EndOfBibitem
\bibitem[Ying \latin{et~al.}(2018)Ying, You, Morris, Ren, Hamilton, and Leskovec]{ying2018hierarchical}
Ying,~Z.; You,~J.; Morris,~C.; Ren,~X.; Hamilton,~W.; Leskovec,~J. Hierarchical graph representation learning with differentiable pooling. \emph{Advances in neural information processing systems} \textbf{2018}, \emph{31}\relax
\mciteBstWouldAddEndPuncttrue
\mciteSetBstMidEndSepPunct{\mcitedefaultmidpunct}
{\mcitedefaultendpunct}{\mcitedefaultseppunct}\relax
\EndOfBibitem
\bibitem[Vinyals \latin{et~al.}(2015)Vinyals, Bengio, and Kudlur]{vinyals2015order}
Vinyals,~O.; Bengio,~S.; Kudlur,~M. Order matters: Sequence to sequence for sets. \emph{arXiv preprint arXiv:1511.06391} \textbf{2015}, \relax
\mciteBstWouldAddEndPunctfalse
\mciteSetBstMidEndSepPunct{\mcitedefaultmidpunct}
{}{\mcitedefaultseppunct}\relax
\EndOfBibitem
\bibitem[Zhang \latin{et~al.}(2018)Zhang, Cui, Neumann, and Chen]{zhang2018end}
Zhang,~M.; Cui,~Z.; Neumann,~M.; Chen,~Y. An end-to-end deep learning architecture for graph classification. Proceedings of the AAAI conference on artificial intelligence. 2018\relax
\mciteBstWouldAddEndPuncttrue
\mciteSetBstMidEndSepPunct{\mcitedefaultmidpunct}
{\mcitedefaultendpunct}{\mcitedefaultseppunct}\relax
\EndOfBibitem
\bibitem[Corso \latin{et~al.}(2020)Corso, Cavalleri, Beaini, Li{\`o}, and Veli{\v{c}}kovi{\'c}]{corso2020principal}
Corso,~G.; Cavalleri,~L.; Beaini,~D.; Li{\`o},~P.; Veli{\v{c}}kovi{\'c},~P. Principal neighbourhood aggregation for graph nets. \emph{Advances in Neural Information Processing Systems} \textbf{2020}, \emph{33}, 13260--13271\relax
\mciteBstWouldAddEndPuncttrue
\mciteSetBstMidEndSepPunct{\mcitedefaultmidpunct}
{\mcitedefaultendpunct}{\mcitedefaultseppunct}\relax
\EndOfBibitem
\bibitem[Buterez \latin{et~al.}(2022)Buterez, Janet, Kiddle, Oglic, and Li{\`o}]{buterez2022graph}
Buterez,~D.; Janet,~J.~P.; Kiddle,~S.~J.; Oglic,~D.; Li{\`o},~P. Graph neural networks with adaptive readouts. \emph{Advances in Neural Information Processing Systems} \textbf{2022}, \emph{35}, 19746--19758\relax
\mciteBstWouldAddEndPuncttrue
\mciteSetBstMidEndSepPunct{\mcitedefaultmidpunct}
{\mcitedefaultendpunct}{\mcitedefaultseppunct}\relax
\EndOfBibitem
\bibitem[Tailor \latin{et~al.}(2022)Tailor, Opolka, Lio, and Lane]{tailor2022egc}
Tailor,~S.~A.; Opolka,~F.; Lio,~P.; Lane,~N.~D. Do We Need Anistropic Graph Neural Networks? International Conference on Learning Representations. 2022\relax
\mciteBstWouldAddEndPuncttrue
\mciteSetBstMidEndSepPunct{\mcitedefaultmidpunct}
{\mcitedefaultendpunct}{\mcitedefaultseppunct}\relax
\EndOfBibitem
\bibitem[Zolfaghari \latin{et~al.}(2021)Zolfaghari, Zhu, Gehler, and Brox]{zolfaghari2021crossclr}
Zolfaghari,~M.; Zhu,~Y.; Gehler,~P.; Brox,~T. Crossclr: Cross-modal contrastive learning for multi-modal video representations. Proceedings of the IEEE/CVF International Conference on Computer Vision. 2021; pp 1450--1459\relax
\mciteBstWouldAddEndPuncttrue
\mciteSetBstMidEndSepPunct{\mcitedefaultmidpunct}
{\mcitedefaultendpunct}{\mcitedefaultseppunct}\relax
\EndOfBibitem
\bibitem[Morgado \latin{et~al.}(2021)Morgado, Vasconcelos, and Misra]{morgado2021audio}
Morgado,~P.; Vasconcelos,~N.; Misra,~I. Audio-visual instance discrimination with cross-modal agreement. Proceedings of the IEEE/CVF Conference on Computer Vision and Pattern Recognition. 2021; pp 12475--12486\relax
\mciteBstWouldAddEndPuncttrue
\mciteSetBstMidEndSepPunct{\mcitedefaultmidpunct}
{\mcitedefaultendpunct}{\mcitedefaultseppunct}\relax
\EndOfBibitem
\bibitem[Shariatnia(2021)]{Shariatnia_Simple_CLIP_2021}
Shariatnia,~M.~M. {Simple CLIP}. 2021\relax
\mciteBstWouldAddEndPuncttrue
\mciteSetBstMidEndSepPunct{\mcitedefaultmidpunct}
{\mcitedefaultendpunct}{\mcitedefaultseppunct}\relax
\EndOfBibitem
\bibitem[MoN()]{MoNA}
MassBank of North America (MoNA). \url{http://massbank.us}\relax
\mciteBstWouldAddEndPuncttrue
\mciteSetBstMidEndSepPunct{\mcitedefaultmidpunct}
{\mcitedefaultendpunct}{\mcitedefaultseppunct}\relax
\EndOfBibitem
\bibitem[Loshchilov and Hutter(2017)Loshchilov, and Hutter]{loshchilov2017decoupled}
Loshchilov,~I.; Hutter,~F. Decoupled weight decay regularization. \emph{arXiv preprint arXiv:1711.05101} \textbf{2017}, \relax
\mciteBstWouldAddEndPunctfalse
\mciteSetBstMidEndSepPunct{\mcitedefaultmidpunct}
{}{\mcitedefaultseppunct}\relax
\EndOfBibitem
\bibitem[Ramsundar \latin{et~al.}(2019)Ramsundar, Eastman, Walters, and Pande]{ramsundar2019deep}
Ramsundar,~B.; Eastman,~P.; Walters,~P.; Pande,~V. \emph{Deep learning for the life sciences: applying deep learning to genomics, microscopy, drug discovery, and more}; O'Reilly Media, 2019\relax
\mciteBstWouldAddEndPuncttrue
\mciteSetBstMidEndSepPunct{\mcitedefaultmidpunct}
{\mcitedefaultendpunct}{\mcitedefaultseppunct}\relax
\EndOfBibitem
\bibitem[Chen \latin{et~al.}(2021)Chen, Zheng, Song, Rao, and Yang]{ijcai2021}
Chen,~J.; Zheng,~S.; Song,~Y.; Rao,~J.; Yang,~Y. Learning Attributed Graph Representation with Communicative Message Passing Transformer. Proceedings of the Thirtieth International Joint Conference on Artificial Intelligence, {IJCAI-21}. 2021; pp 2242--2248, Main Track\relax
\mciteBstWouldAddEndPuncttrue
\mciteSetBstMidEndSepPunct{\mcitedefaultmidpunct}
{\mcitedefaultendpunct}{\mcitedefaultseppunct}\relax
\EndOfBibitem
\bibitem[Lim and Lee(2021)Lim, and Lee]{lim2021predicting}
Lim,~S.; Lee,~Y.~O. Predicting chemical properties using self-attention multi-task learning based on SMILES representation. 2020 25th International Conference on Pattern Recognition (ICPR). 2021; pp 3146--3153\relax
\mciteBstWouldAddEndPuncttrue
\mciteSetBstMidEndSepPunct{\mcitedefaultmidpunct}
{\mcitedefaultendpunct}{\mcitedefaultseppunct}\relax
\EndOfBibitem
\end{mcitethebibliography}

\end{document}